%% file: Short_Arxiv_version.tex
\documentclass[11pt,twoside]{article}

\usepackage{xcolor}

\usepackage{fullpage}

\usepackage{epsf}
\usepackage{fancyhdr}
\usepackage{graphics}
\usepackage{graphicx} 
\usepackage{float} 
\usepackage{subfigure} 
\usepackage{psfrag}
\usepackage{comment}

\usepackage[linesnumbered,ruled]{algorithm2e}
\DontPrintSemicolon	

\usepackage{color}
\usepackage{amsthm}
\usepackage{amsfonts}
\usepackage{amsmath}
\usepackage{bm}
\usepackage{amssymb,bbm}
\usepackage[numbers]{natbib}
\usepackage{algorithmic}
\usepackage[usestackEOL]{stackengine}

\usepackage{url}
\usepackage[colorlinks=True,linkcolor=magenta,citecolor=blue,urlcolor=blue,pagebackref=true,backref=true]
{hyperref}
\renewcommand*{\backref}[1]{\ifx#1\relax \else Page #1 \fi}
\renewcommand*{\backrefalt}[4]{%
    \ifcase #1 \footnotesize{(Not cited.)}%
    \or        \footnotesize{(Cited on page~#2.)}%
    \else      \footnotesize{(Cited on pages~#2.)}%
    \fi}

\usepackage{nicefrac}

\usepackage{chngpage}

\usepackage{tabularx}%

\usepackage{enumitem}
\usepackage{booktabs}
\usepackage{pbox}

\usepackage{caption}

\usepackage{mathtools}

\usepackage{fullpage}
\allowdisplaybreaks
\input{final_macros.tex}

\SetKwInput{KwInput}{Input}                
\SetKwInput{KwOutput}{Output}              

\begin{document}
\begin{center}

{\bf{\LARGE{An Exponentially Increasing Step-size for Parameter Estimation in Statistical Models}}}
  
\vspace*{.2in}
{{
\begin{tabular}{ccccc}
Nhat Ho$^{\flat}$ & Tongzheng Ren{$^{\diamond}$} & Sujay Sanghavi{$^\dagger$} & Purnamrita Sarkar{$^\flat$} & Rachel Ward{$^\circ$}\\
\end{tabular}
}}

\vspace*{.1in}

\begin{tabular}{c}
Department of Computer Science, University of Texas at Austin$^\diamond$, \\
Department of Statistics and Data Sciences, University of Texas at Austin$^\flat$, \\
Department of Electrical and Computer Engineering, University of Texas at Austin$^\dagger$, \\
Department of Mathematics, University of Texas at Austin$^{\circ}$,\\
\end{tabular}




\today

\vspace*{.2in}

\begin{abstract}
Using gradient descent (GD) with fixed or decaying step-size is a standard practice in unconstrained optimization problems. However, when the loss function is only locally convex, such a step-size schedule artificially slows GD down as it cannot explore the flat curvature of the loss function. To overcome that issue, we propose to exponentially increase the step-size of the GD algorithm. Under homogeneous assumptions on the loss function, we demonstrate that the iterates of the proposed \emph{exponential step size gradient descent} (EGD) algorithm converge linearly to the optimal solution. Leveraging that optimization insight, we then consider using the EGD algorithm for solving parameter estimation under both regular and non-regular statistical models whose loss function becomes locally convex when the sample size goes to infinity. We demonstrate that the EGD iterates reach the final statistical radius within the true parameter after a logarithmic number of iterations, which is in stark contrast to a \emph{polynomial} number of iterations of the GD algorithm in non-regular statistical models. Therefore, the total computational complexity of the EGD algorithm is \emph{optimal} and exponentially cheaper than that of the GD for solving parameter estimation in non-regular statistical models while being comparable to that of the GD in regular statistical settings. To the best of our knowledge, it resolves a long-standing gap between statistical and algorithmic computational complexities of parameter estimation in non-regular statistical models. Finally, we provide targeted applications of the general theory to several classes of statistical models, including generalized linear models with polynomial link functions and location Gaussian mixture models.

\end{abstract}

\end{center}

\section{Introduction}
\label{sec:introduction}
Gradient descent (GD)~\cite{Polyak_Introduction, Nesterov_Introduction, bubeck2015convex} has been a popular algorithm for solving the unconstrained optimization problem of the following form:
\begin{align}
    \min_{\theta \in \mathbb{R}^{d}} f(\theta), \label{eq:population_loss}
\end{align}
where $f: \mathbb{R}^{d} \to \mathbb{R}$ is referred to as \textit{population loss function} when applied to the statistical inference problem. We denote $\theta^{*}$ as the optimal solution of the optimization problem~\eqref{eq:population_loss}. When the population loss function $f$ is locally strongly convex and smooth around $\theta^{*}$, it is well-known that with proper step size and local initialization, the GD iterates converge geometrically fast to the optimal solution $\theta^{*}$ with the contraction rate depends on the condition number, namely, the ratio between the maximum and the minimum eigenvalue of the Hessian matrix of the function $f$. However, when the population loss function $f$ is locally convex, namely, the minimum eigenvalue of the Hessian matrix of $f$ is 0 at the true parameter $\theta^{*}$, the GD iterates converge sub-linearly to the optimal solution. The local convexity of the function $f$ emerges frequently from non-regular/singular statistical models when the signal-to-noise ratio (SNR) is weak or when these models are over-specified. Examples of the singular models include mixture models and hierarchical models~\cite{Lindsay-1995, Mclachlan-1988, Blei-et-al, Rousseau-2011} when the true number of clusters is unknown in practice and we over-specify the number of clusters to avoid under-fitting these models, leading to singularities. The singular models also appear in matrix-type problems, such as low-rank matrix factorization problems~\cite{candes2011tight, chen2015fast, li2018algorithmic} (e.g., matrix sensing and matrix completion) and dimension reduction methods (e.g, factor analysis and principal component analysis), where the true rank of the matrix is unknown and we over-specify the rank of the matrix. The singularity of statistical models also happens to non-linear regression problems, such as generalized linear models with polynomial link functions~\cite{Carroll-1997, Fienup_82,Shechtman_Yoav_etal_2015, candes_2011,Netrapalli_Prateek_Sanghavi_2015} when the signal-to-noise ratio is low. Finally, the singular models are also popular in econometrics literature, including stochastic frontier models~\cite{Lee_1986,Lee_1993} with singular information matrix, or missing data problems, including informative non-response model~\cite{Heckman_1976, Diggle_1994, shaiko1991pre} where the non-regularity problem arises from estimating non-responsive behavior in sample surveys.

\noindent\textbf{Sub-linear behavior of GD.} To shed light on the sub-linear behavior of the GD algorithm when $f$ is not locally strongly convex, we specifically consider $f(\theta) = \frac{\|\theta\|^{2p}}{2p}$ when $p > 1$. This function is convex but not strongly convex around the optimal solution $\theta^{*} = 0$. When $p = 2$, this function arises from the least-square loss function of the phase retrieval problem with infinite sample size~\cite{Fienup_82,Shechtman_Yoav_etal_2015, candes_2011,Netrapalli_Prateek_Sanghavi_2015}. For general $p$, this function corresponds to the least-square loss function of the generalized linear models with polynomial link function~\cite{Carroll-1997}. The GD iterates for solving that loss function are: $\theta_{\text{GD}}^{t + 1} = \theta_{\text{GD}}^{t} - \eta \nabla f(\theta_{\text{GD}}^{t}) = \theta_{\text{GD}}^{t}(1 - \eta \|\theta_{\text{GD}}^{t}\|^{2p - 2})$. As $\theta_{\text{GD}}^{t}$ approaches $\theta^{*} = 0$, the contraction coefficient $(1 - \eta \|\theta_{\text{GD}}^{t}\|^{2p - 2})$ goes to 1 independent of the fixed-step size $\eta$, which indicates that the convergence rate of the GD iterates to the optimal parameter $\theta^{*}$ is sub-linear (or to be more concrete, the rate is $\mathcal{O}(t^{-1/(2p - 2})$ as we demonstrate later in Proposition~\ref{prop:convergence_gd}).

\begin{figure*}[!t]
  \centering \includegraphics[width=0.45\textwidth]{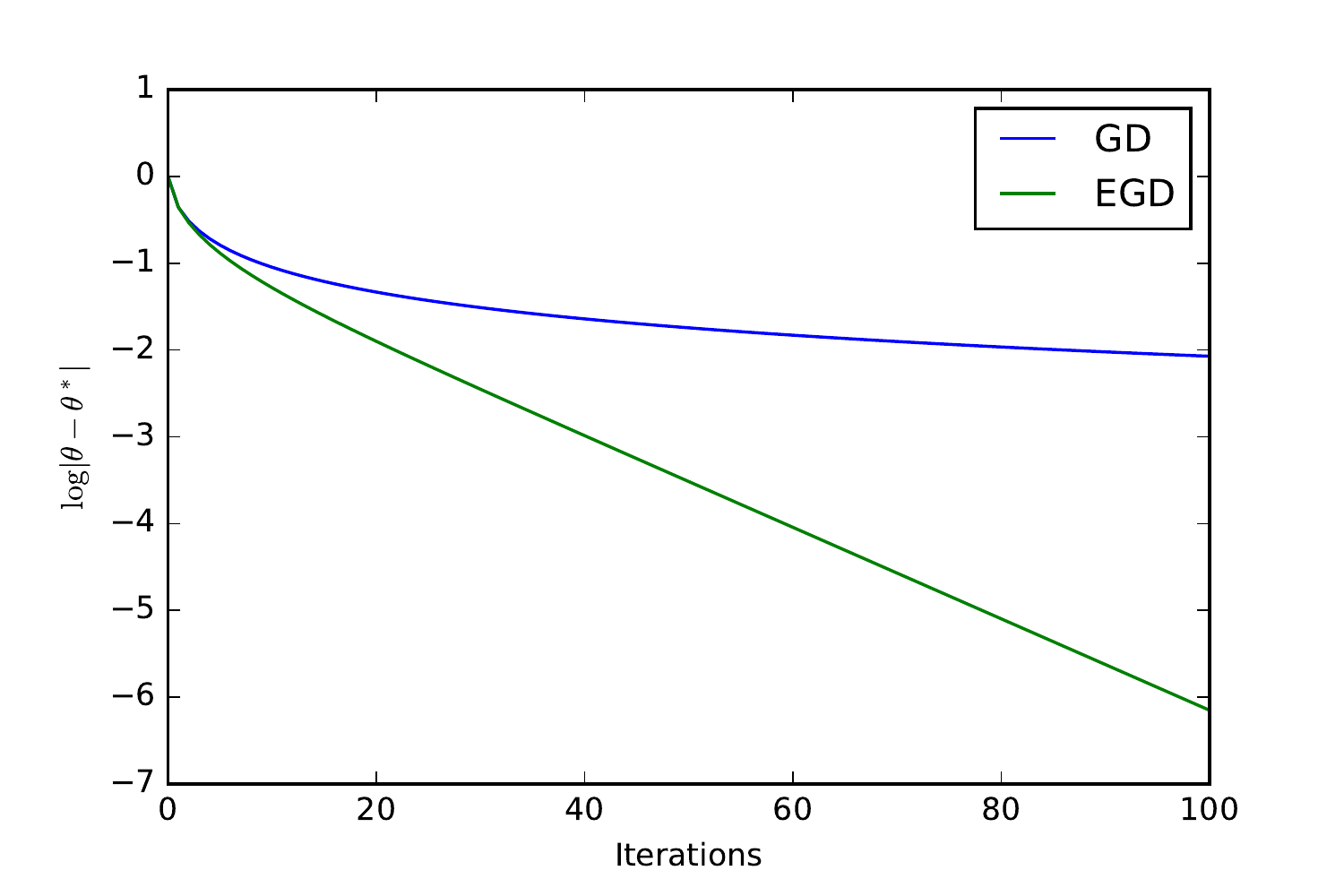}
  \includegraphics[width=0.45\textwidth]{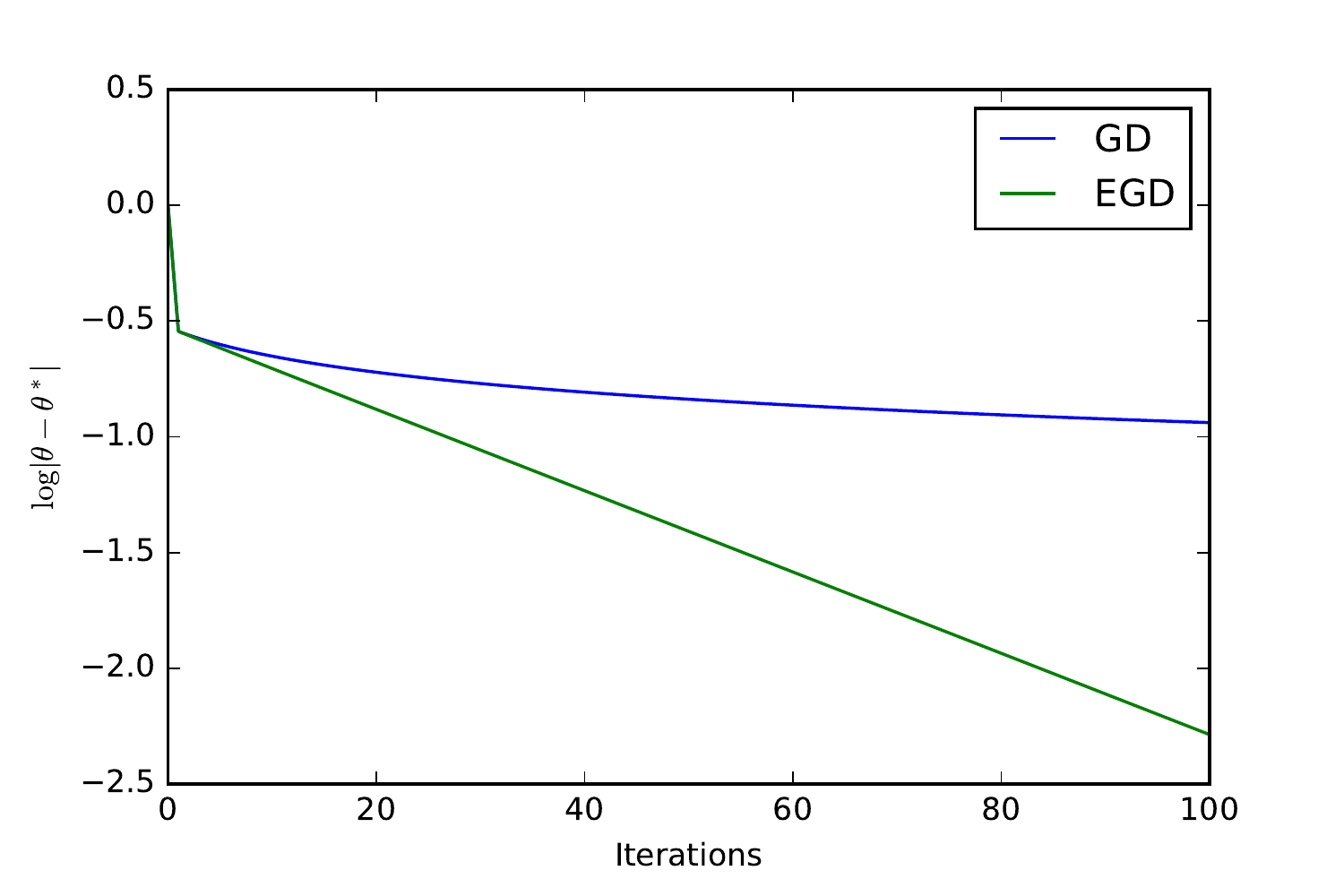}
  \caption{\textit{GD versus EGD iterates for solving the population loss function $f(\theta) = \theta^{2p}/ (2p)$ when $p \in \{2,4\}$}. \textbf{Left}: $p = 2$;  \textbf{Right}: $p = 4$. The EGD iterates converge linearly to the true parameter $\theta^{*} = 0$, while the GD iterates converge to $\theta^{*}$ at a sub-linear rate.}
  \label{fig:population_low_SNR_Generalized_Linear}
\end{figure*}

\noindent\textbf{Exponential step size gradient descent (EGD).} A key insight from that example is that the fixed-step size in the GD algorithm is the bottleneck preventing GD from the linear convergence, as a fixed step-size cannot adapt to the curvature of the convex loss function. To account for this limitation of the fixed step-size GD, consider the example we mentioned before, we need $\eta \propto \|\theta_{\text{GD}}^t\|^{2-2p}$ to make $\theta_{\text{GD}}^t$ converge to $\theta^* = 0$ geometrically, which conversely means $\eta$ should be increased to infinite with an exponential rate as the optimization procedure proceeds. This motivates us to propose an \emph{exponential step size gradient descent} algorithm (EGD), which takes the following form:
\begin{align}
    \theta^{t + 1} : = \theta^{t} - \frac{\eta}{\beta^{t}} \nabla f(\theta^{t}), \label{eq:population_sample_exponential_GD}
\end{align}
where $\beta \in (0, 1]$ is some given scale parameter and $\eta > 0$ is the step size. When $\beta = 1$, the EGD iterates become the standard fixed-step size GD updates. The EGD algorithm had been considered before in the setting of training deep learning models~\cite{Arora_EGD}. 

\vspace{0.5 em}
\noindent
\textbf{Contribution.} Our contribution is three-fold and can be summarized as follows:

\vspace{0.5 em}
\noindent
\textbf{1. Linear convergence of the EGD updates:} We assume that the population loss function $f$ is locally convex and \emph{homogeneous}: the maximum eigenvalue of the population loss function satisfies $\lambda_{\max}(\nabla^2 f(\theta)) \lesssim \|\theta - \theta^*\|^{\alpha}$ while the population loss function satisfies a generalized PL condition: $\|\nabla f(\theta)\| \gtrsim (f(\theta) - f(\theta^*))^{1 - \frac{1}{\alpha + 2}}$ for non-negative constant $\alpha$. When $\alpha > 0$, under suitable conditions on the scaling parameter $\beta$ and the step size we demonstrate that both the population EGD updates and their objective values respectively converge linearly to the optimal solution $\theta^{*}$ and optimal objective value $f(\theta^{*})$. This linear convergence rate is much faster than the sub-linear convergence rate $\mathcal{O}(t^{-1/\alpha})$ of the GD iterates to $\theta^{*}$. We illustrate these behaviors in Figure~\ref{fig:population_low_SNR_Generalized_Linear}.

When $\alpha = 0$, namely, the function $f$ satisfies the Polyak-Łojasiewicz inequality and the standard smoothness condition, we prove that the population EGD iterates and their objective values converge geometrically fast to some neighborhoods around the optimal solution and then diverge. However, as we will see shortly, such behavior is sufficient for parameter estimation in statistical models if we choose a proper scaling parameter $\beta$.

\vspace{0.5 em}
\noindent
\textbf{2. Optimal complexities of the EGD for parameter estimation in statistical models:} Leveraging the linear convergence of the EGD algorithm for solving the optimal solution of $f$, we also would like to demonstrate the benefits of the EGD algorithm over the GD algorithm for the problem of parameter estimation in statistical models. In particular, we can view the population function $f$ in equation~\eqref{eq:population_loss} as an expectation of the \textit{sample loss function} $f_{n}$ with respect to the data, namely, $f(\theta) = \mathbb{E} \brackets{f_{n}(\theta)}$ where $n$ is the sample size. In statistical models, we only have access to the data that are i.i.d. some from unknown distribution $\mathcal{P}_{\theta^{*}}$ and the sample loss function $f_{n}$ (e.g, the negative log-likelihood function or the least-square loss function). 

To obtain an estimation for the true parameter $\theta^{*}$, we minimize the sample loss function $\min_{\theta \in \mathbb{R}^{d}} f_{n}(\theta)$ and run the EGD algorithm for solving that minimization problem. The corresponding sample EGD updates take the form:
\begin{align}
    \theta_{n}^{t + 1} : = \theta_{n}^{t} - \frac{\eta}{\beta^{t}} \nabla f_{n}(\theta_{n}^{t}). \label{eq:sample_exponential_GD}
\end{align}
We prove that under the homogeneous settings of $f$ with $\alpha > 0$, which arise from non-regular/singular settings of statistical models, as long as the uniform concentration bound $\sup_{\|\theta - \theta^{*}\| \leq r} \|\nabla f_{n}(\theta) - \nabla f(\theta)\| \leq r^{\gamma} \varepsilon(n, \delta)$ holds with probability $1 - \delta$ for some $\gamma \geq 0$ and for some noise function $\varepsilon(n, \delta)$, with suitable fixed $\beta$ or sample size dependent $\beta^2 = 1 - C/\log(1/ \varepsilon(n, \delta))$ for some constant $C$, the sample EGD updates $\{\theta_{n}^{t}\}_{t \geq 0}$ reach the final statistical radius $\mathcal{O}(\varepsilon(n, \delta)^{1/(\alpha + \gamma - 1)})$ around the true parameter $\theta^{*}$ up to logarithmic terms after a logarithmic number of iterations $\mathcal{O}(\log(1/\varepsilon(n, \delta)))$ for fixed $\beta$ or $\mathcal{O}(\log^2(1/\varepsilon(n, \delta)))$ number of iterations for sample size dependent $\beta$. We would like to remark that the uniform concentration bound had been utilized widely in the literature~\cite{Siva_2017, hardt16,kuzborskij2018data,charles2018stability,Raaz_Ho_Koulik_2020, Raaz_Ho_Koulik_2018_second, Ho_Instability} to establish the statistical guarantee of optimization algorithms for solving parameter estimation in statistical models. Since the per iteration cost of the EGD algorithm is $\mathcal{O}(n)$, this indicates that the total computational complexity of the EGD algorithm for reaching the final statistical radius around the true parameter is $\mathcal{O}(n\log(1/\varepsilon(n, \delta)))$, which is optimal in terms of $n$ (up to some logarithmic factor). In contrast, the GD iterates converge to the similar statistical radius $\mathcal{O}(\varepsilon(n, \delta)^{1/(\alpha + \gamma - 1)})$ after polynomial number of iterations $\mathcal{O}(\varepsilon(n, \delta)^{-\alpha/(\alpha + \gamma - 1)})$. Under popular statistical models, the noise function $\varepsilon(n, \delta) = \mathcal{O}(\sqrt{d \log (1/\delta)/n})$; therefore, these results indicate that the total computational complexity of the GD algorithm to reach the final statistical radius is $\mathcal{O}(n^{1 + \alpha/(\alpha + \gamma - 1)})$, which is more expensive than the linear computational complexity of the EGD algorithm. 

For the homogeneous settings of $f$ with $\alpha = 0$, which stems from the regular statistical models, with fixed $\beta$, as long as the uniform concentration bound $\sup_{\|\theta - \theta^{*}\| \leq r} \|\nabla f_{n}(\theta) - \nabla f(\theta)\| \leq \varepsilon(n, \delta)$ holds with probability $1 - \delta$, the statistical rate of the EGD iterates consist of two errors: (i) statistical error $\mathcal{O}(\varepsilon(n, \delta))$ and (ii) optimization error $\mathcal{O}(C'/(\beta^{-2}-1))$ where $C'$ is some positive constant. We demonstrate that the non-vanishing optimization error when $\beta$ is fixed can be resolved by balancing the statistical and optimization errors, which leads to the sample size dependent $\beta^2 = 1 - C''/\log(1/ \varepsilon(n, \delta))$ for some constant $C''$, which is consistent with the sample size dependent $\beta$ in the non-regular statistical settings that lead to optimal sample and computational complexities of the EGD. Under that choice of sample size dependent $\beta$, the EGD iterates reach a neighborhood of radius $\mathcal{O}(\varepsilon(n, \delta))$ within the true parameter $\theta^{*}$ after $\mathcal{O}(\log(1/\varepsilon(n, \delta)))$ number of iterations, which is optimal and comparable to those of the fixed-step size GD iterates under the regular statistical models. 

\vspace{0.5 em}
\noindent
\textbf{3. Examples of popular statistical models:} We apply our general theory for the EGD algorithm to two specific models: the generalized linear model and the Gaussian mixture model. For the generalized linear model with polynomial link function $g(r) = r^{p}$ and $p \geq 2$, we demonstrate that under the high signal-to-noise ratio (SNR) regime, i.e., the regular settings and low SNR regime, namely, the non-regular/singular settings, the statistical rates of EGD updates are respectively $\mathcal{O}((d/n)^{1/2})$ and $\mathcal{O}((d/n)^{1/(2p)})$ up to logarithmic terms after logarithmic number of iterations. This is in stark contrast to the GD algorithm, which needs $\mathcal{O}((n/d)^{(p-1)/p})$ iterations to reach the final statistical radius under the low SNR regime. Moving to the Gaussian mixture model, we specifically consider the symmetric two-component Gaussian mixtures, which have been considered extensively before to study the non-asymptotic behaviors of the EM algorithm~\cite{Siva_2017, Caramanis-nips2015, Raaz_Ho_Koulik_2020, Raaz_Ho_Koulik_2018_second}. We prove that when the model is over-specified~\cite{Chen1992, Rousseau-2011}, i.e, $\theta^{*} = 0$, the EGD algorithm reaches the final statistical radius $\mathcal{O}((d/n)^{1/4})$ up to logarithmic terms after logarithmic number of iterations. This is significantly  cheaper than the EM algorithm, which requires $\mathcal{O}(\sqrt{n/d})$ iterations. When the model is exactly-specified, i.e., $\|\theta^{*}\|$ is sufficiently large, the EGD iterates have the statistical rate $\mathcal{O}((d/n)^{1/2})$ up to logarithmic terms after logarithmic number of iterations, which is optimal and comparable to the complexities of the EM algorithm.

\vspace{0.5 em}
\noindent
\textbf{Organization.} In Section~\ref{sec:optimization_rate_EGD}, we establish the linear optimization rate of EGD updates for solving the population loss function under the homogeneous assumption. Leveraging the insight from that linear convergence rate, we study the statistical rate of EGD iterates for parameter estimation under both regular and non-regular/singular statistical models with fixed $\beta$ in Section~\ref{sec:statistics_rate_EGD} and with sample size dependent $\beta$ in Section~\ref{sec:varied_beta}. We then apply the general theory to generalized linear models in Appendix~\ref{sec:examples} and Gaussian mixture models in Appendix~\ref{sec:example_Gaussian_mixture}. We conclude the paper in Section~\ref{sec:discussion} while deferring the proofs and remaining materials to the Supplementary Material. 

\vspace{0.5 em}
\noindent
\textbf{Notation.} For any $n \in \mathbb{N}$, we denote $[n] = \{1, 2, \ldots, n\}$. For any two sequences $\{a_{n}\}$ and $\{b_{n}\}$, the notation $a_{n} = \mathcal{O}(b_{n})$ means that there exists a universal constant $C > 0$ such that $a_{n} \leq C b_{n}$ for all $n \geq 1$. For any matrix $A \in \mathbb{R}^{d \times d}$ where $d \geq 1$, the notation $\lambda_{\max}(A)$ is the maximum eigenvalue of the matrix $A$. For any vector $x \in \mathbb{R}^{d}$, the notation $\|x\|$ is the $\ell_{2}$ norm of the vector $x$.  

\section{Optimization Rate for the EGD Algorithm}
\label{sec:optimization_rate_EGD}
In this section, we study the convergence rate of EGD applied to the population loss function $f$. Our result relies on the homogeneous assumption on the population loss function $f$, which is described in Assumption~\ref{assump:homogeneous}. We want to remark that the homogeneous assumption is not strong and appears in several classes of non-regular statistical models, including generalized linear models~\cite{Carroll-1997, Fienup_82,Shechtman_Yoav_etal_2015, candes_2011,Netrapalli_Prateek_Sanghavi_2015} when the signal-to-noise ratio is low, and mixture models and hierarchical models~\cite{Lindsay-1995, Mclachlan-1988, Blei-et-al, Rousseau-2011} when the number of clusters in the model is over-specified. Finally, in Appendix~\ref{sec:inhomogeneous_settings} we provide a brief discussion on the behavior of the EGD algorithm when the population loss function is locally convex and does not satisfy the homogeneous assumption~\ref{assump:homogeneous}.
\begin{enumerate}[label=(W.1)]
\item \label{assump:homogeneous} (Homogeneous Assumption)
There exist constants $\alpha \geq 0$ and $\rho > 0$ such that the function $f$ is locally convex in $\mathcal{B}(\theta^*, \rho)$ and for all $\theta\in \mathbb{B}(\theta^*, \rho)$, we have
\begin{align}
    \lambda_{\max}(\nabla^2 f(\theta)) & \leq c_1 \|\theta - \theta^*\|^{\alpha}, \label{eq:bound_lambda_max} \\
    \|\nabla f(\theta)\| & \geq c_2(f(\theta) - f(\theta^*))^{1 - \frac{1}{\alpha + 2}},  \label{eq:generalized_PL}
\end{align}
where $c_1, c_{2}$ are some positive universal constants.
\end{enumerate}
\vspace{-0.6em}
A few remarks on the homogeneous assumption~\ref{assump:homogeneous} are in order. First, the condition~\eqref{eq:bound_lambda_max} provides a growth condition for the maximum eigenvalue of the Hessian matrix of the population loss function $f$ when the parameter $\theta$ approaches the true parameter $\theta^{*}$. Note that, when $\alpha = 0$, it reduces to the standard smoothness assumption that is widely used in the optimization literature. For condition~\eqref{eq:generalized_PL}, it is a generalized Łojasiewicz condition and characterizes the local growth of the gradient in terms of high order polynomial functions. When $\alpha = 0$, this condition corresponds to the popular Polyak-Łojasiewicz (PL) inequality~\cite{bubeck2015convex}, which is used to establish the linear convergence of the GD iterates for solving the smooth population loss function $f$. Examples of homogeneous assumption~\ref{assump:homogeneous} when $\alpha = 0$ include the regular statistical models and $\alpha > 0$ include non-regular/singular statistical models, such as the least-square loss function of the generalized linear models with polynomial link functions $g(r) = r^{p}$ where $p \in \mathbb{N}$ (see equation~\eqref{eq:generalized_linear_model} for more details) whose the constant $\alpha = 2p - 2$ and the negative log-likelihood function corresponding to the over-specified symmetric two-component location Gaussian mixture~\eqref{eq:Gaussian_mixture} with constant $\alpha = 2$ in Appendix~\ref{sec:example_Gaussian_mixture}.
\vspace{-0.5 em}
\subsection{When $\alpha > 0$}
We first focus on the homogeneous setting~\ref{assump:homogeneous} with $\alpha > 0$ and demonstrate that the objective values $f(\theta^{t})$ at the population EGD updates converge geometrically to the optimal objective value $f(\theta^{*})$.

\begin{figure*}[!t]
    \centering
    \includegraphics[width=0.45\textwidth]{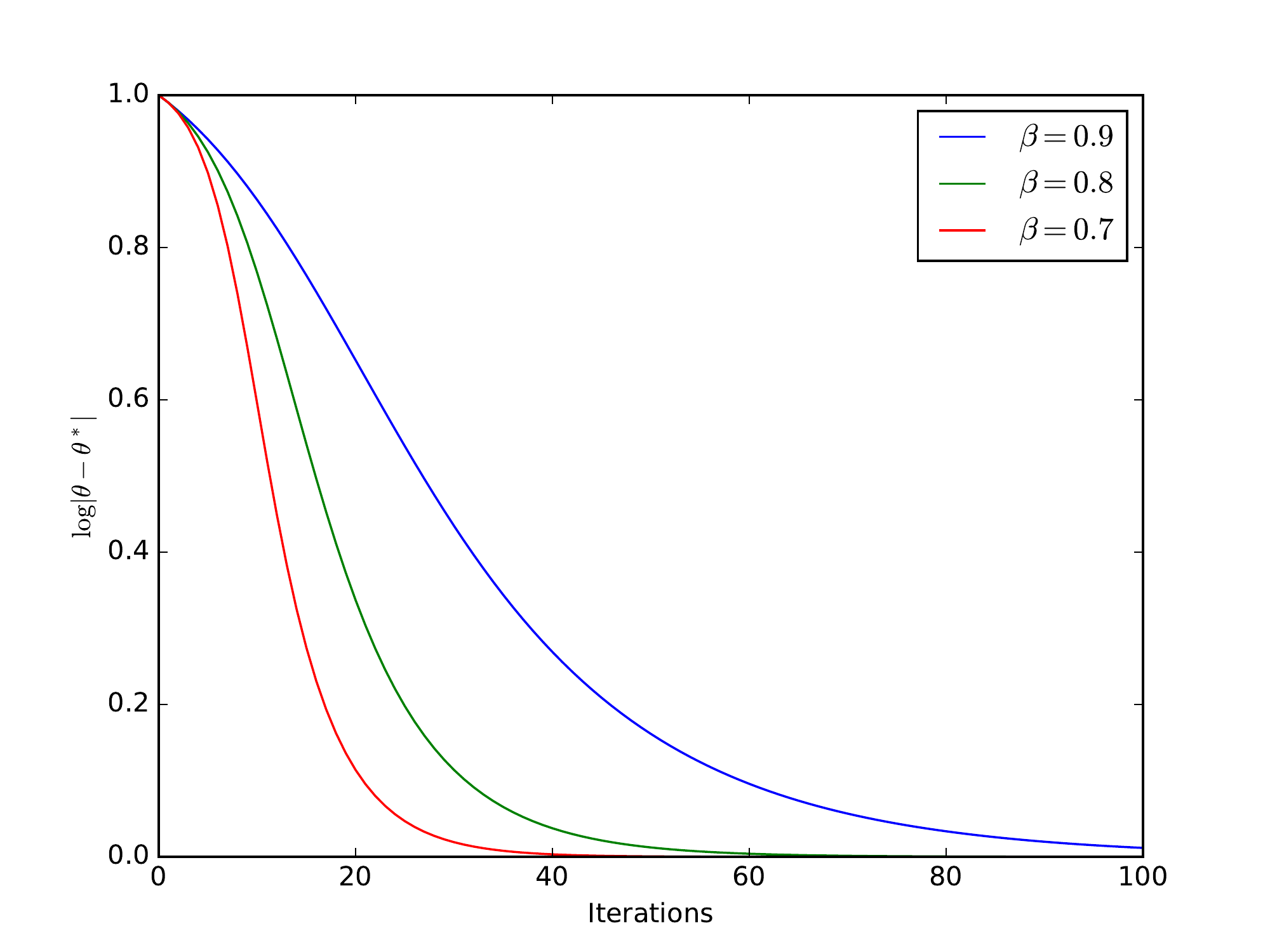}
    \includegraphics[width=0.45\textwidth]{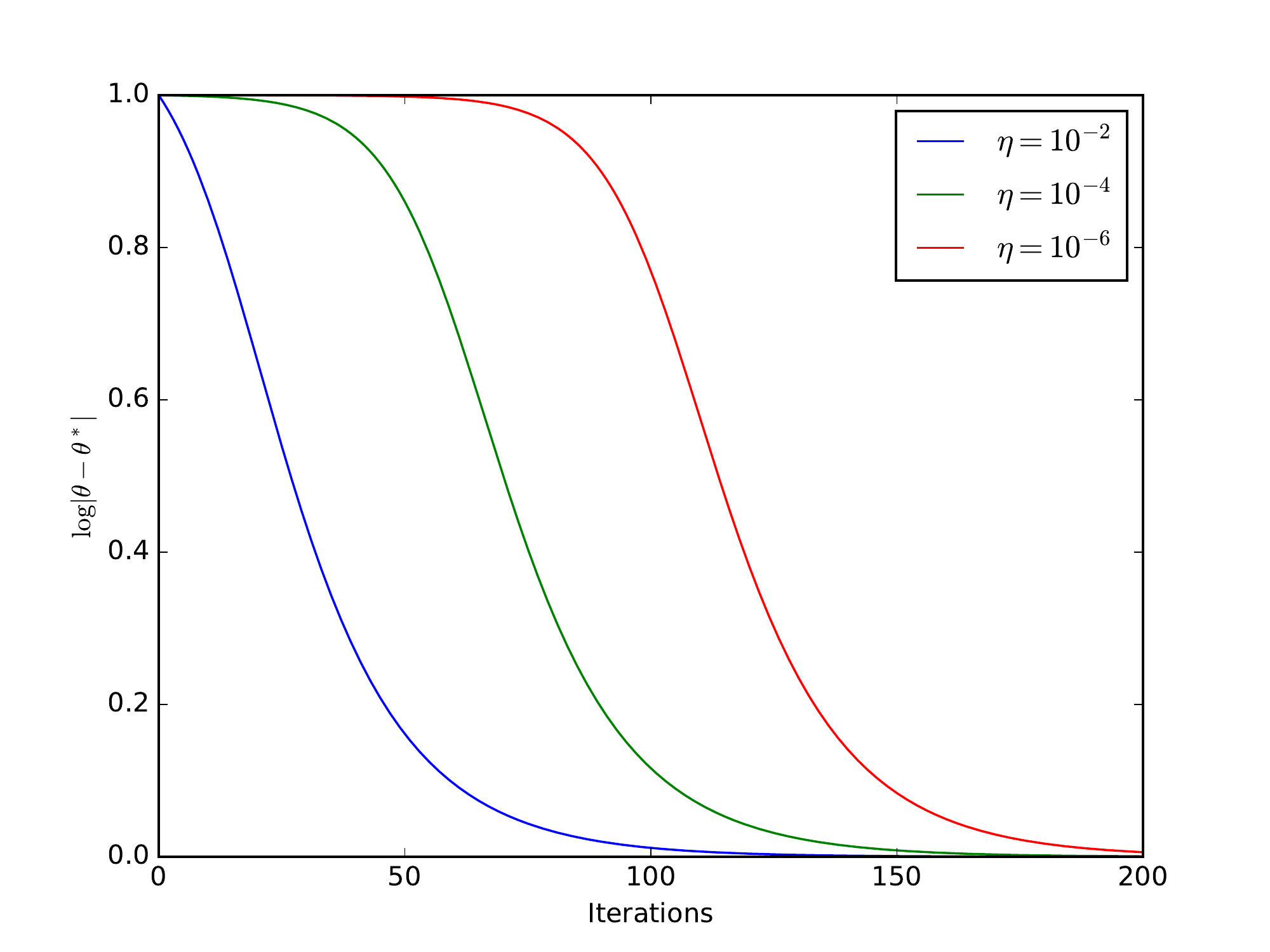}
    \caption{Illustration on the effects of $\eta$ and $\beta$ on the convergence rate of the EGD iterates for solving $f(\theta) = \theta^4/ 4$. \textbf{Left}: Effects of different $\beta$ where $\eta=10^{-2}$. \textbf{Right}: Effects of different $\eta$ where $\beta = 0.9$.}
    \label{fig:effects_hpm}
\end{figure*}
\begin{theorem} 
\label{theorem:homogeneous_setting}
Assume that Assumption~\ref{assump:homogeneous} holds for some $\alpha > 0$. As long as $\beta$ is chosen such that $\frac{1 - \beta^{\frac{\alpha + 2}{\alpha}}}{\beta} \leq \frac{c_{2}^{\alpha + 1}}{2 c_{1}(\alpha + 2)^{\alpha}}$ where $c_{1}$ and $c_{2}$ are universal constants in Assumption~\ref{assump:homogeneous}, then the population EGD iterates $\{\theta^{t}\}_{t \geq 0}$ in equation~\eqref{eq:population_sample_exponential_GD} satisfy
\begin{align*}
    f(\theta^{t}) - f(\theta^{*}) \leq \bar{C} \beta^{ \frac{\alpha + 2}{\alpha}t},
\end{align*}
where $\bar{C}$ is some universal constant such that $\bar{C} \geq f(\theta^{0}) - f(\theta^{*})$ and 
\vspace{-0.6em}
\begin{align}
    \frac{2 (1 - \beta^{\frac{\alpha + 2}{\alpha}})}{c_{2} \beta} \leq \eta \bar{C}^{\alpha/(\alpha + 2)} \leq \frac{c_{2}^{\alpha}}{c_{1} (\alpha + 2)^{\alpha}}. \label{eq:upper_bound_eta}
\end{align}
\end{theorem}
\vspace{-0.6em}
\noindent
Proof of Theorem~\ref{theorem:homogeneous_setting} is in Appendix~\ref{subsec:proof:theorem:homogeneous_setting}. To illustrate the linear convergence of the EGD algorithm in Theorem~\ref{theorem:homogeneous_setting}, we run both the GD and EGD algorithms for solving the population loss function $f(\theta) = \theta^{2p}/(2p)$ when $p \in \{2,4\}$. The experimental results are in Figure~\ref{fig:population_low_SNR_Generalized_Linear}.
Let us now make a few remarks on the results and assumptions of Theorem~\ref{theorem:homogeneous_setting}. 

\vspace{0.5 em}
\noindent
\textbf{On the assumptions on $\eta$ and $\beta$:} Note that when $\alpha > 0$, $\lim_{\beta \to 1} \left(1-\beta^{\frac{\alpha + 2}{\alpha}}\right)/\beta = 0$. Hence, there always exists a value $\beta < 1$ such that the condition $\frac{1 - \beta^{\frac{\alpha + 2}{\alpha}}}{\beta} \leq \frac{c_{2}^{\alpha + 1}}{2 c_{1}(\alpha + 2)^{\alpha}}$ holds. As $\beta$ determines the contraction rate of the objective values of EGD iterates, generally we want $\beta$ to be as small as possible. For the assumption on $\eta$ in inequalities~\eqref{eq:upper_bound_eta}, our proof reveals that the upper bound is used to guarantee the descent of the objective, while the lower bound is used to guarantee the linear contraction. Hence, in practical implementation, a simple but effective strategy is to take sufficiently small $\eta$ at the beginning. After $O(\log 1/\beta)$ iterations, the step-size will be increased to proper scale and the objective will start to decrease geometrically. 

\vspace{0.5 em}
\noindent
\textbf{Illustration of the influence of $\eta$ and $\beta$:} We now illustrate the effects of $\beta$ and $\eta$ on a simple function $f(\theta) = \theta^4/4$. This function satisfies the homogeneous assumption~\ref{assump:homogeneous} with constant $\alpha = 2$. The experimental results are shown in Figure~\ref{fig:effects_hpm}. From these results, we note that the EGD iterates converge to the optimal solution faster as $\beta$ decreases, which aligns with our theoretical prediction. Regarding the effect of the choice of $\eta$, when $\eta$ is set properly, the EGD iterates directly converge to the optimal solution linearly (as shown in the case when $\eta = 10^{-2}$). When $\eta$ is chosen to be small, the EGD iterates will first experience a burn-in phase, then start to converge to the optimal solution linearly.

\vspace{0.5 em}
\noindent\textbf{Linear convergence rate of EGD iterates:} In light of the linear convergence rate of $f(\theta^{t})$ to $f(\theta^{*})$ under the homogeneous assumption~\ref{assump:homogeneous}, we demonstrate in the following corollary that we also have linear convergence rate of the EGD updates $\{\theta_{t}\}_{t \geq 0}$ to the true parameter $\theta^{*}$.
\begin{corollary}
\label{cor:updates_EGD}
Under the same assumptions as in Theorem~\ref{theorem:homogeneous_setting}, we find that
\vspace{-0.2em}
\begin{align*}
    \|\theta^{t} - \theta^{*}\| \leq \frac{(\alpha + 2)^{\alpha + 2}}{c_{2}^{\alpha + 2}} \bar{C} \beta^{ \frac{t}{\alpha}},
\end{align*}
where $\bar{C}$ is a universal constant in Theorem~\ref{theorem:homogeneous_setting} and $c_{2}$ is universal constant in Assumption~\ref{assump:homogeneous}.
\end{corollary}
\vspace{-0.5 em}
The proof of Corollary~\ref{cor:updates_EGD} is in Appendix~\ref{subsec:proof:cor:updates_EGD}. The linear convergence rate of EGD for approximating the optimal solution of the population loss function $f$ provides important insight into the logarithmic number of iterations of EGD for solving parameter estimation in statistical models (see Section~\ref{sec:statistics_rate_EGD} for a detailed statement). 

\vspace{0.5 em}
\noindent
\textbf{Sub-linear convergence rate of fixed-step size GD:} Under the homogeneous assumptions, in Proposition~\ref{prop:convergence_gd} we prove that the GD updates have sub-linear convergence rate to the true parameter $\theta^{*}$. Therefore, by exponentially increasing the step size of the GD algorithm by $\beta^{t}$, the EGD iterates converge to $\theta^{*}$ much faster than the GD iterates.
\begin{proposition}
\label{prop:convergence_gd}
Under Assumptions~\ref{assump:homogeneous}, we can find a universal constant $C'$ such that
\begin{align*}
    \|\theta_{\text{GD}}^t - \theta^*\|\leq C'/(\eta t)^{1/\alpha},
\end{align*}
where $\theta_{\text{GD}}^{t + 1} = \theta_{\text{GD}}^{t} - \eta \nabla f(\theta_{\text{GD}}^{t})$. Furthermore, this bound is tight, namely, there exists function $f$ satisfying Assumptions~\ref{assump:homogeneous} and $\|\theta_{\text{GD}}^t - \theta^*\| \geq \frac{C'}{(\eta t)^{1/\alpha}}$.
\end{proposition}
The proof of Proposition~\ref{prop:convergence_gd} follows the proof argument from~\cite{mei2021leveraging} and subsequently Lemma 4 in Appendix E of~\cite{ren2021towards}; therefore, the proof is omitted.
\vspace{-0.5 em}
\subsection{When $\alpha = 0$}
\label{sec:optimization_rate_PL}
We now study the behaviors of the EGD algorithm when $\alpha = 0$, namely, the population loss function $f$ satisfies the local PL inequality and smooth condition, in the homogeneous assumption~\ref{assump:homogeneous}. 

\begin{figure}[t!]
  \centering
  \includegraphics[width=0.5\textwidth]{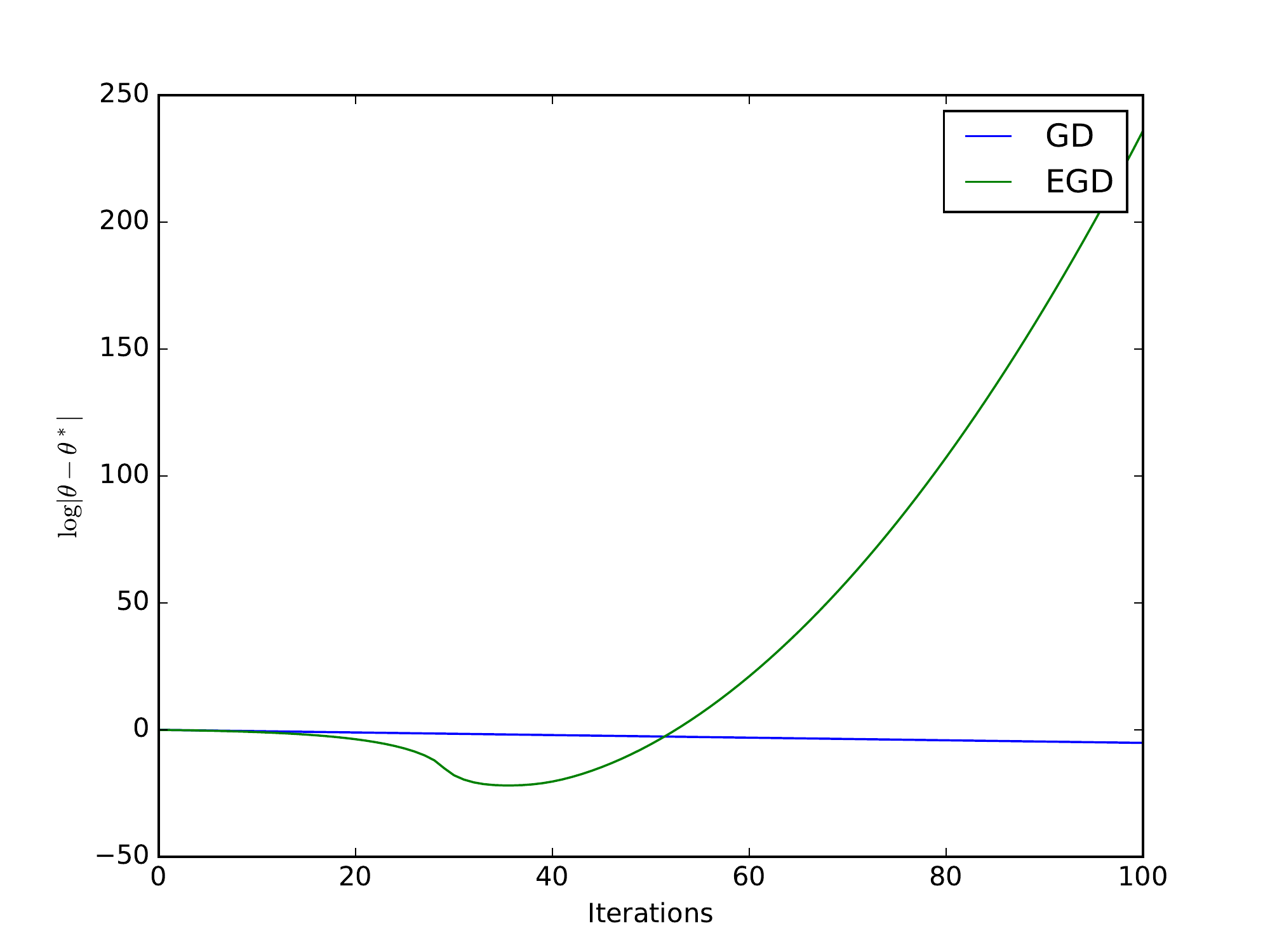}
  \caption{GD and EGD algorithm iterates for solving $f(\theta) = \theta^2/2$, an instance of strongly convex and smooth function. As indicated in the plot, EGD iterates converge faster than GD at the first several iterations, and start to diverge when $\frac{\eta}{\beta^t} > \frac{2}{c_1}$ where $c_{1} = 1$. Here, we choose $\beta = 0.9$ and $\eta = 0.05$.}
  \label{fig:optimization_strongly_convex}
  \vspace{-1 em}
\end{figure}

\begin{theorem}
\label{theorem:optimization_EGD_strongly_convex}
Assume that the function $f$ satisfies the homogeneous assumption~\ref{assump:homogeneous} when $\alpha = 0$, namely, $f$ satisfies the local PL inequality and smooth condition. Assume that the step size $\eta$ satisfies $c_{1} \eta < 2$ where $c_{1}$ is the universal constant in Assumption~\ref{assump:homogeneous}. Then, as long as $t \leq T = \log \parenth{\frac{2}{c_{1} \eta}}/ \log(1/ \beta)$, we have that
\begin{align*}
     f(\theta^{t + 1}) - f(\theta^{*}) \leq \parenth{1 - \frac{c_{2}^2 \eta}{\beta^{t}} + \frac{c_{1} c_{2}^2 \eta^2}{2\beta^{2t}}} \parenth{f(\theta^{t}) - f(\theta^{*})}.
\end{align*}
Here, $c_{1}, c_{2}$ are universal constants in the homogeneous assumption~\ref{assump:homogeneous}. Furthermore, the above recursive inequality leads to
\begin{align*}
    f(\theta^{T}) - f(\theta^{*}) \leq \exp \parenth{-\frac{c_{2}^2 \eta ((c_{1}\eta/ 2)^{-1} - 1)(\beta^{-1} - c_{1} \eta/ 2)}{(\beta^{-2} - 1)}} \parenth{f(\theta^{0}) - f(\theta^{*})}. 
\end{align*}
\end{theorem}
\noindent Proof of Theorem~\ref{theorem:optimization_EGD_strongly_convex} is in Appendix~\ref{sec:proof:theorem:optimization_EGD_strongly_convex}. A few comments on Theorem~\ref{theorem:optimization_EGD_strongly_convex} are in order. First, the result of Theorem~\ref{theorem:optimization_EGD_strongly_convex} indicates that the objective values at the EGD iterates are decreasing within first $T = \log \parenth{\frac{2}{c_{1} \eta}}/ \log(1/ \beta)$ iterations. Furthermore, the objective value at the EGD iterate at step $T$ only gets to a neighborhood of radius 
\begin{align*}
    \exp \parenth{-\frac{c_{2}^2 \eta ((c_{1}\eta/ 2)^{-1} - 1)(\beta^{-1} - c_{1} \eta/ 2)}{(\beta^{-2} - 1)}} \parenth{f(\theta^{0}) - f(\theta^{*})}.
\end{align*}
When $\beta$ approaches 1, the radius approaches 0. After step $T$, the objective value of the EGD iterates will diverge. We illustrate these two-phase behaviors of the EGD iterates in Figure~\ref{fig:optimization_strongly_convex} where we specifically consider the function $f(\theta) = \theta^2/2$. 

Second, since the local strong convexity implies the local PL inequality (or more concretely, $\mu$-strongly convexity implies $\|\nabla f(\theta)\|^2 \geq 2\mu(f(\theta) - f(\theta^*))$), it suggests that these two-phase behaviors will happen when the function $f$ is locally strongly convex and smooth. 

Lastly, compared with the fixed step-size GD which enjoys the rate
\begin{align*}
    f(\theta_{\mathrm{GD}}^{t+1}) - f(\theta^*) \leq \left(1 - c_2^2 \eta + \frac{c_1 c_2^2 \eta^2}{2}\right) (f(\theta_{\mathrm{GD}}^{t}) - f(\theta^*)),
\end{align*}
when $\eta \geq \frac{1}{c_1}$, then the EGD does not have any benefits over the GD. However, when $\eta \ll \frac{1}{c_1}$, then the EGD will increase the step-size to the proper range and converge faster than the GD, then diverge to infinity.
\vspace{-0.5 em}
\section{Statistical Rate of the EGD with Fixed $\beta$}
\label{sec:statistics_rate_EGD}
Leveraging the convergence rate of the EGD algorithm for solving the population loss function $f$, we now demonstrate the benefits of the EGD algorithm over the GD algorithm for the parameter estimation in statistical models. In particular, we can view the population function $f$ in equation~\eqref{eq:population_loss} as $f(\theta) = \mathbb{E} \brackets{f_{n}(\theta)}$ where $n$ is the sample size and the outer expectation is taken with respect to the i.i.d. data from the unknown distribution $\mathcal{P}_{\theta^{*}}$. In statistical models, we only have access to the data and the sample loss function $f_{n}$ (e.g, the negative log-likelihood function or the least-square loss function). To obtain an estimation for $\theta^{*}$, we minimize the sample loss function
\begin{align}
    \min_{\theta \in \mathbb{R}^{d}} f_{n}(\theta). \label{eq:sample_loss_function}
\end{align}
In Sections~\ref{sec:general_theory_statistics_rate_EGD} and~\ref{sec:stats_rate_local_PL}, we respectively provide statistical guarantees for the EGD iterates in solving the sample loss function $f_n$ in the optimization problem~\eqref{eq:sample_loss_function} when the sample size $n$ is fixed and the statistical models are non-regular/singular ($\alpha > 0$), namely, when the Hessian matrix of $f$ is degenerate at the true parameter $\theta^{*}$, and regular ($\alpha = 0$). Then, in Appendices~\ref{sec:examples} and~\ref{sec:example_Gaussian_mixture} we respectively discuss an application of the general theory to a generalized linear model with polynomial link function and a Gaussian mixture model.
\vspace{-0.5 em}
\subsection{Non-regular/Singular Statistical Models: $\alpha > 0$}
\label{sec:general_theory_statistics_rate_EGD}
In order to talk about the statistical rate of the EGD iterates $\{\theta_{n}^{t}\}_{t \geq 0}$ in equation~\eqref{eq:sample_exponential_GD}, we need the following assumption on the stability of the sample gradient $\nabla f_{n}$ around the population gradient $\nabla f$, which was previously used in~\cite{Siva_2017, hardt16,kuzborskij2018data,charles2018stability,Raaz_Ho_Koulik_2020, Raaz_Ho_Koulik_2018_second, Ho_Instability} to establish statistical rates of optimization algorithms.
\vspace{-0.6em}
\begin{enumerate}[label=(W.2)] 
\item \label{assump:stab} (Stability Property)
For a given parameter $\gamma \geq 0$, there exist a noise function $\varepsilon: \mathbb{N} \times (0,1] \to \mathbb{R}^{+}$, a universal constant $c_3 > 0$, and some positive parameter $\rho > 0$ such that 
\begin{align*}
    \sup_{\theta\in \mathbb{B}(\theta^*, r)} \|\nabla f_n(\theta) - \nabla f(\theta)\|\leq c_3 r^\gamma \varepsilon(n, \delta),
\end{align*}
    \vspace{-0.6em}
for all $r \in (0, \rho)$ with probability $1 - \delta$.
\end{enumerate}
\vspace{-0.2em}
The uniform concentration bound~\ref{assump:stab} provides control on the growth of the noise function, which is the difference between the population and sample loss function. When $\gamma = 0$, this reduces to a standard empirical process bound that holds for most regular statistical models~\cite{vanderVaart-Wellner-96}. When $\gamma > 0$, this condition has been used to analyze the behavior of optimization algorithms for solving parameter estimation of non-regular statistical models~\cite{Raaz_Ho_Koulik_2020,Raaz_Ho_Koulik_2018_second, Kwon_minimax, Ho_Instability}. Under this setting of $\gamma$, when $\theta$ approaches $\theta^{*}$ or, equivalently, $r \to 0$, $\nabla f_{n}(\theta)$ approaches $\nabla f(\theta)$ and $\theta^{*}$ becomes the stationary point of $f_{n}$. Examples of Assumption~\ref{assump:stab} include the generalized linear model~\eqref{eq:generalized_linear_model} whose $\gamma = p - 1$ (see equation~\eqref{eq:concentration_generalized_linear}) and the Gaussian mixture model~\eqref{eq:Gaussian_mixture} whose $\gamma = 1$ (see equation~\eqref{eq:concentration_Gaussian_mixtures}). In these examples, the choice of noise function $\varepsilon(n, \delta) = \sqrt{\frac{d + \log(1/\delta)}{n}}$.

Based on the homogeneous assumption~\ref{assump:homogeneous} and the stability property~\ref{assump:stab}, we have the following result regarding the statistical rate of the sample EGD updates for solving problem~\eqref{eq:sample_loss_function} when $\alpha \geq \gamma \geq 1$. 
\begin{theorem}
\label{theorem:statistical_rate_EGD}
Assume that Assumptions~\ref{assump:homogeneous} and~\ref{assump:stab} hold with $\alpha \geq \gamma \geq 1$. Furthermore, the step size $\eta$ and the scaling parameter $\beta$ are chosen such that $\frac{\eta c_{1} \|\theta_{n}^{0} - \theta^{*}\|^{\alpha} \beta}{(\alpha + 1) (\alpha + 2)} + \beta^{2/\alpha} \geq 1$ and $\eta c_{1}(\alpha + 2) 3^{\alpha} \|\theta_{n}^{0} - \theta^{*}\|^{\alpha} \leq \alpha + 1$ where $\theta_{n}^{0}$ is an initialization of the EGD sequence $\{\theta_{n}^{t}\}_{t \geq 0}$. Then, there exist universal constants $C_{1}$ and $C_{2}$ such that as long as the sample size $n$ is large enough such that $t \geq C_{1} \log(1/ \varepsilon(n, \delta))$, with probability $1 - \delta$ we have
\begin{align*}
    \min_{1 \leq k \leq t} \|\theta_{n}^{k} - \theta^{*}\| \leq C_{2} \parenth{\varepsilon(n, \delta)}^{\frac{1}{\alpha + 1 - \gamma}}.
\end{align*}
\end{theorem}
Proof of Theorem~\ref{theorem:statistical_rate_EGD} is in Appendix~\ref{subsec:proof:theorem:statistical_rate_EGD}. We have the following comments:

\vspace{0.5 em}
\noindent\textbf{On the assumption $\alpha \geq \gamma$:} The assumption $\alpha \geq \gamma$ is to guarantee that the signal is stronger than noise in statistical models, which guarantees that the statistical rate of EGD in Theorem~\ref{theorem:statistical_rate_EGD} is meaningful. To see that claim more clearly, we specifically consider $f_{n}(\theta) = \|\theta\|^{\alpha + 2} + \omega \|\theta\|^{\gamma + 1} \sqrt{\frac{d}{n}}$ where $\omega \sim \mathcal{N}(0, 1)$. It is clear that the population loss function $f(\theta) = \|\theta\|^{\alpha + 2}$ and $\theta^{*} = 0$. With $1/2$ probability, we have $\omega < 0$ and the optimal solution $\theta_{n}^{*}$ that minimizes $f_{n}$ satisfies $\|\theta_{n}^{*} - \theta^{*}\| = \mathcal{O} \parenth{(d/n)^{\frac{1}{2(\alpha + 1 - \gamma)}}}$. To guarantee that the statistical rate of $\theta_{n}^{*}$ is meaningful and not faster than the parametric rate, we would need $\alpha + 1 - \gamma \geq 1$, which is equivalent to $\alpha \geq \gamma$. 

\vspace{0.5 em}
\noindent\textbf{On the assumptions for $\eta$ and $\beta$:} Due to the influence of the noise function, the assumptions on $\eta$ and $\beta$ for the statistical rates of EGD iterates are different from those for the optimization rate of the EGD algorithm applied to the population loss function $f$. However, they follow the same principle. As $\lim_{\beta\to 1}\left(1 - \beta^{2/\alpha}\right)/\beta = 0$, there always exists $\beta < 1$ such that the condition $\frac{\eta c_{1} \|\theta_{n}^{0} - \theta^{*}\|^{\alpha} \beta}{(\alpha + 1) (\alpha + 2)} + \beta^{2/\alpha} \geq 1$ holds. For the second condition $\eta c_{1}(\alpha + 2) 3^{\alpha} \|\theta_{n}^{0} - \theta^{*}\|^{\alpha} \leq \alpha + 1$, this is satisfied as long as $\eta$ is sufficiently small. In practice, we can still start from sufficiently small $\eta$, and after $O(\log 1/\beta)$ iterations, the step-size will be increased to a proper scale where the parameter will converge to the statistical radius geometrically.

\vspace{0.5 em}
\noindent\textbf{On the minimum number of iterations:} The statistical guarantee for the EGD iterates is only for some $k < t$ such that $\|\theta^{k} - \theta^{*}\| \leq C_{2}\varepsilon(n, \delta)^{\frac{1}{\alpha + 1 - \gamma}}$
where $t \geq C_{1} \log(1/ \varepsilon(n, \delta))$. It indicates that after reaching the statistical radius, EGD may diverge (see experimental results in Figures~\ref{fig:GLM_Low_SNR} and~\ref{fig:GMM_Low_SNR} for such phenomenon). While this minimum number of iterates may sound limited, it is inherent to any fast and unstable algorithm (See remark after Theorem 2 of~\cite{Ho_Instability}). In principle, we can apply early stopping via cross-validation with an extra computation $\mathcal{O}(nd)$, which does not affect the total computational complexity of the EGD algorithm for reaching the final statistical radius. We illustrate the early stopping via cross-validation in Figures~\ref{fig:GLM_Low_SNR} and~\ref{fig:GMM_middle_SNR} when we run the EGD algorithm for solving parameter estimation in generalize linear models and Gaussian mixture models.

\vspace{0.5 em}
\noindent\textbf{Comparing to fixed-step size GD:} Under similar assumptions as those of the EGD algorithm, we demonstrate in Proposition~\ref{proposition:fixed_step_GD} that the GD iterates reach a similar statistical radius as that of the EGD iterates after $\mathcal{O}(\varepsilon(n, \delta)^{-\frac{\alpha}{\alpha + 1 - \gamma}})$ iterations. Since the EGD and GD have similar per iteration cost, the total computational complexity of the EGD algorithm is much cheaper than that of the GD algorithm for parameter estimation in non-regular statistical models.
\begin{proposition}
\label{proposition:fixed_step_GD}
Assume that Assumptions~\ref{assump:homogeneous} and~\ref{assump:stab} hold with $\alpha \geq \gamma$. Suppose the sample size $n$ satisfies $\varepsilon(n, \delta) \leq C$ for some universal constant $C$. Then there exist universal constant $C_1$ and $C_2$, such that for any fixed $\tau \in \left(0, \frac{1}{\alpha + 1 - \gamma}\right)$, as long as $t \geq C_1 \varepsilon(n, \delta)^{-\frac{\alpha}{\alpha + 1 - \gamma}} \log \frac{1}{\tau}$, we have that
\vspace{-0.6em}
\begin{align*}
    \|\theta_{n, \text{GD}}^t - \theta^*\| \leq C_2 \varepsilon(n, \delta)^{\frac{1}{\alpha + 1 - \gamma} - \tau}.
\end{align*}
\end{proposition}
\vspace{-0.5 em}
\noindent The proof of Proposition~\ref{proposition:fixed_step_GD} is an application of Theorem 1 in~\cite{Ho_Instability} given that the rate of population GD updates to the true parameter $\theta^{*}$ is $\mathcal{O}(t^{-1/\alpha})$ (see Proposition~\ref{prop:convergence_gd}) and the uniform concentration bound of $\nabla f_{n}$ around $\nabla f$ in Assumption~\ref{assump:stab}; therefore, this proof is omitted.
\vspace{-0.5 em}
\subsection{Regular Statistical Models: $\alpha = 0$}
\label{sec:stats_rate_local_PL}
We now study the statistical behaviors of the sample EGD iterates for solving the sample loss function $f_{n}$ when its corresponding population loss function $f$ satisfies the homogeneous assumption~\ref{assump:homogeneous} and the uniform concentration between $\nabla f_{n}$ and $\nabla f$ in Assumption~\ref{assump:stab} is satisfied when $\gamma = 0$. It turns out that the sample EGD iterates also exhibit the similar two-phase behaviors as their population counterparts in Section~\ref{sec:optimization_rate_PL}. 
\begin{theorem}
\label{theorem:statistics_EGD_strongly_convex}
Assume that the homogeneous assumption~\ref{assump:homogeneous} holds when $\alpha = 0$, namely, the population loss function $f$ satisfies the local PL inequality and smoothness condition, and the stability assumption~\ref{assump:stab} holds when $\gamma = 0$. Assume that the step size $\eta$ is such that $\eta c_{1} < 1$ where $c_{1}$ is a universal constant in Assumption~\ref{assump:homogeneous}. Then, by denoting $\bar{T} = \log(1/(\eta c_{1}))/ \log(1/ \beta)$, with probability $1 - \delta$ we have
\begin{align*}
    & \hspace{-2 em} \min_{1 \leq t \leq \bar{T}} \|\theta_{n}^{t} - \theta^{*}\| \leq c \eta \varepsilon(n, \delta) \frac{((\eta c_{1})^{-1} - 1)}{\beta^{-1} - 1} \\
    & + \exp \parenth{- \frac{(1 - \eta c_{1})(\beta^{-1} - c_{1} \eta)}{2(\beta^{-2} - 1)}} \|\theta_{n}^{0} - \theta^{*}\|,
\end{align*}
where $c$ is a universal constant in Assumption~\ref{assump:stab}.
\end{theorem}
Proof of Theorem~\ref{theorem:statistics_EGD_strongly_convex} is in Appendix~\ref{sec:proof:theorem:statistics_EGD_strongly_convex}. We have the following remarks on the result of the statistical rate of the sample EGD iterates $\{\theta_{n}^{t}\}_{t \geq 0}$. First, the upper bound on the statistical rate of the sample EGD iterates consists of two errors: (i) statistical error: $c \eta \varepsilon(n, \delta) \frac{((\eta c_{1})^{-1} - 1)}{\beta^{-1} - 1}$, which is comparable to that of GD iterates; (ii) optimization error: $\exp \parenth{- \frac{(1 - \eta c_{1})(\beta^{-1} - c_{1} \eta)}{2(\beta^{-2} - 1)}} \|\theta_{n}^{0} - \theta^{*}\|$. For standard statistical models, the noise function $\varepsilon(n, \delta) = \sqrt{\frac{d + \log(1/\delta)}{n}}$. Therefore, the statistical error of the sample EGD iterates is of the order $\mathcal{O}(\sqrt{d/n})$. For the optimization error of the sample EGD iterates, as $\beta \to 1$, then the optimization error approaches 0. Second, similar to the population EGD iterates, after the $\bar{T} = \log(1/(\eta c_{1}))/ \log(1/ \beta)$ iterations, the sample EGD iterates will diverge. Therefore, the sample EGD iterates also have two-phase behaviors. 
\vspace{-0.5 em}
\section{Statistical Rate of the EGD algorithm with Sample Size Dependent $\beta$}
\label{sec:varied_beta}
In Theorem~\ref{theorem:statistics_EGD_strongly_convex}, the non-vanishing optimization error $\exp \parenth{- \frac{(1 - \eta c_{1})(\beta^{-1} - c_{1} \eta)}{2(\beta^{-2} - 1)}} \|\theta_{n}^{0} - \theta^{*}\|$ in the statistical rate of the EGD iterates when the population loss function $f$ satisfies the local PL inequality and smoothness condition can be undesirable. In this section, we provide statistical rates of the EGD algorithm when the scaling parameter $\beta$ is varied with the sample size $n$ and the dimension $d$. In particular, when the population loss function satisfies the homogeneous assumption~\ref{assump:homogeneous} with $\alpha = 0$, we would like to choose $\beta$ such that the optimization error $\exp \parenth{- \frac{(1 - \eta c_{1})(\beta^{-1} - c_{1} \eta)}{2(\beta^{-2} - 1)}} \|\theta_{n}^{0} - \theta^{*}\|$ is at the order $\mathcal{O}(\varepsilon(n, \delta))$, which holds when $\beta^2 = 1 - \frac{(1 - \eta c_{1})^2}{2 \log(1/ \varepsilon(n, \delta))}$. It is to guarantee that the EGD iterates can reach the final statistical radius (up to some logarithmic factor) $\mathcal{O}(\log(1/ \varepsilon(n, \delta))) \varepsilon(n, \delta)$ after a logarithmic number of iterations. With that choice of $\beta$, we demonstrate that when the population loss function $f$ satisfies the homogeneous assumption~\ref{assump:homogeneous} with $\alpha > 0$, the EGD iterates still converge to the final statistical radius $\mathcal{O}(\varepsilon(n,\delta)^{\frac{1}{\alpha + 1- \gamma}})$ after $\mathcal{O}(\log^2(1/\varepsilon(n, \delta)))$ number of iterations.
\begin{proposition}
\label{proposition:rate_EGD_varied_beta}
Assume that 
\begin{align}
    \beta^2 = 1 - \frac{(1 - \eta c_{1})^2}{2 \log(1/ \varepsilon(n, \delta))}, \label{eq:varied_beta_value_general}
\end{align}
where $c_{1}$ is the universal constant in the homogeneous assumption~\ref{assump:homogeneous}. Then, the following holds:

(a) (When $\alpha = 0$) When the homogeneous assumption~\ref{assump:homogeneous} holds with $\alpha = 0$ and the stability assumption~\ref{assump:stab} holds when $\gamma = 0$, given the conditions on $\eta$ as those in Theorem~\ref{theorem:statistics_EGD_strongly_convex}, there exist universal constants $C_{1}$ and $C_{2}$ such that as long as $t \geq C_{1} \log(1/ \varepsilon(n, \delta))$ with probability $1 - \delta$ we find that
\begin{align*}
    \min_{1 \leq k \leq t} \|\theta_{n}^{t} - \theta^{*}\| \leq C_{2} \log(1/\varepsilon(n, \delta)) \varepsilon(n, \delta).
\end{align*}
(b) (When $\alpha > 0$) When the homogeneous assumption~\ref{assump:homogeneous} holds with $\alpha > 0$ and the stability assumption~\ref{assump:stab} holds when $\alpha \geq \gamma \geq 1$, given the conditions on $\eta$ as those in Theorem~\ref{theorem:statistical_rate_EGD} there exist universal constants $C_{3}$ and $C_{4}$ such that as long as $t \geq C_{3} \log^2(1/ \varepsilon(n, \delta))$ with probability $1 - \delta$ we find that
\begin{align*}
    \min_{1 \leq k \leq t} \|\theta_{n}^{k} - \theta^{*}\| \leq C_{4} \cdot \parenth{\log(1/\varepsilon(n, \delta)) \varepsilon(n, \delta)}^{\frac{1}{\alpha + 1 - \gamma}}.
\end{align*}
\end{proposition}
\noindent The result of part (a) follows directly from the result of Theorem~\ref{theorem:statistics_EGD_strongly_convex} when $\beta$ takes the value in equation~\eqref{eq:varied_beta_value_general}. The proof of part (b) follows directly the proof of Theorem~\ref{theorem:statistical_rate_EGD} where the lower bound $C_{3} \log^2(1/ \varepsilon(n, \delta))$ on the iteration is to ensure that $\beta^{C_{3} \log^2(1/ \varepsilon(n, \delta))} \geq C' \varepsilon(n, \delta)$ for some constant $C'$ and the extra term $\log(1/\varepsilon(n, \delta))$ in the statistical rate stems from the constant $\frac{\beta^{-1/\alpha}}{(1 - \beta^{\frac{\alpha + 1 - \gamma}{\alpha}})}$ next to the statistical rate of the EGD iterates in Theorem~\ref{theorem:statistical_rate_EGD}. Therefore, the proof of Proposition~\ref{proposition:rate_EGD_varied_beta} is omitted.

A few comments with the results of Proposition~\ref{proposition:rate_EGD_varied_beta} are in order. First, comparing to the behaviors of the GD algorithm in Proposition~\ref{prop:convergence_gd} when the population loss function satisfies the homogeneous assumption~\ref{assump:homogeneous} with $\alpha = 0$, the EGD iterates also reach to the similar statistical radius $\mathcal{O}(\varepsilon(n, \delta))$ (up to some logarithmic factor) after comparable logarithmic number of iterations $\mathcal{O}(\log(1/ \varepsilon(n, \delta)))$. When the population loss function satisfies the homogeneous assumption~\ref{assump:homogeneous} with $\alpha > 0$, the EGD iterates converge to the final statistical radius $\mathcal{O}(\varepsilon(n, \delta)^{\frac{1}{\alpha + 1 - \gamma}})$ (up to some logarithmic factor) after $\mathcal{O}(\log^2(1/ \varepsilon(n, \delta)))$ number of iterations, which is slightly more than that of the EGD iterates with fixed $\beta$ in Theorem~\ref{theorem:statistical_rate_EGD} and much cheaper than that of the GD iterates with fixed step size in Proposition~\ref{prop:convergence_gd}.

\vspace{-0.5 em}
\section{Conclusion}
\label{sec:discussion}
In this paper, we propose using the exponential gradient descent (EGD) algorithm, an adaptation of the gradient descent (GD) algorithm by exponentially increasing the step size by $\beta^{t}$ where $t$ is the iteration number and $\beta \in (0,1]$, with targeted applications for parameter estimation in statistical models. We demonstrate that under homogeneous assumptions~\ref{assump:homogeneous} on the population loss function $f$, the EGD iterates and their objective values have linear convergence rates to the true parameter $\theta^{*}$ and optimal objective value $f(\theta^{*})$ when $\alpha > 0 $ (here, $\alpha$ is a constant in Assumption~\ref{assump:homogeneous}) or to some neighborhoods around these values when $\alpha = 0$. Leveraging these insights, under the uniform concentration~\ref{assump:stab} and the non-singular statistical settings, namely, $\alpha > 0$, with either suitable fixed $\beta$ or sample size dependent $\beta^2 = 1 - C/\log(1/\varepsilon(n, \delta))$ where $C$ is some constant, we prove that the sample EGD iterates reach the final statistical radius $\mathcal{O}(\varepsilon(n, \delta)^{\frac{1}{\alpha + 1 - \gamma}})$ after a logarithmic number of iterations for noise function $\varepsilon(n, \delta)$ and constant $\gamma$ in Assumption~\ref{assump:stab}. This indicates that the EGD algorithm has better computational complexity than the GD algorithm, which is $\mathcal{O}(n \varepsilon(n, \delta)^{-\frac{\alpha}{\alpha + 1 - \gamma}})$, for reaching the similar statistical radius. Under the regular statistical models, i.e., $\alpha = 0$, we prove that with the similar sample size dependent $\beta$ as being chosen in the non-regular settings, the EGD iterates reach to the similar radius of convergence $\mathcal{O}(\varepsilon(n, \delta))$ around the true parameter as that of the GD after logarithmic number of iterations.  
\bibliographystyle{abbrv}
\bibliography{Nhat}
\clearpage

\newpage
\appendix
\onecolumn
\begin{center}
{\bf \Large Supplement for ``An Exponentially Increasing Step-size for Parameter Estimation in Statistical Models"}
\end{center}
In this supplementary material, we present proofs of the key results on the optimization and statistical rates of the EGD algorithm in Appendix~\ref{sec:key_proofs}. We provide proofs for auxiliary results in Appendix~\ref{sec:auxiliary_results}. We discuss an application of the general theory of the EGD to generalized linear models with polynomial link function in Appendix~\ref{sec:examples} and Gaussian mixture models in Appendix~\ref{sec:example_Gaussian_mixture}. Finally, we provide discussion when the population loss function $f$ is inhomogeneous in Appendix~\ref{sec:inhomogeneous_settings} while we discuss the behaviors of the EGD algorithm under the middle SNR regime of the generalized linear models in Appendix~\ref{sec:discussion_middle_SNR}.
\section{Proofs of Key Results} 
\label{sec:key_proofs}
In this Appendix, we provide proofs for key results in the main text.
\subsection{Proof of Theorem~\ref{theorem:homogeneous_setting}}
\label{subsec:proof:theorem:homogeneous_setting}


According to the updates of the EGD iterates in equation~\eqref{eq:population_sample_exponential_GD}, we obtain that
\begin{align}
    f(\theta^{t + 1}) - f(\theta^{*}) & = f\left(\theta^{t} - \frac{\eta}{\beta^{t}} \nabla f(\theta^{t})\right) - f(\theta^{*}) \nonumber \\
    & \leq f(\theta^{t}) - f(\theta^{*}) - \frac{\eta}{\beta^{t}} \| \nabla f(\theta^{t})\|^2 + \frac{c_{1}}{2} \frac{\eta^2}{\beta^{2t}} \|\theta^{t} - \theta^{*}\|^{\alpha} \| \nabla f(\theta^{t})\|^2 \nonumber \\
    & = f(\theta^{t}) - f(\theta^{*}) - \frac{\eta}{\beta^{t}}\left(1 - \frac{c_{1}}{2} \frac{\eta}{\beta^{t}} \|\theta^{t} - \theta^{*}\|^{\alpha} \right) \| \nabla f(\theta^{t})\|^2. \label{eq:key_inequality}
\end{align}
where the inequality is due to Assumption~\ref{assump:homogeneous}. As long as we can guarantee that $c_{1} \frac{\eta}{\beta^{t}} \|\theta^{t} - \theta^{*}\|^{\alpha} \leq 1$, the inequality~\eqref{eq:key_inequality} becomes
\begin{align}
    f(\theta^{t + 1}) - f(\theta^{*}) & \leq f(\theta^{t}) - f(\theta^{*}) - \frac{\eta}{2 \beta^{t}} \| \nabla f(\theta^{t})\|^2 \nonumber \\
    & \leq f(\theta^{t}) - f(\theta^{*}) - \frac{c_{2}^2 \eta}{2 \beta^{t}} \left( f(\theta^{t}) - f(\theta^{*})\right)^{\frac{2\alpha + 2}{\alpha + 2}}, \nonumber \\
    & = \left(f(\theta^{t}) - f(\theta^{*}) \right) \left(1 - \frac{c_{2}^2 \eta}{2 \beta^{t}} (f(\theta^{t}) - f(\theta^{*}))^{\frac{\alpha}{\alpha+ 2}} \right), \label{eq:key_inequality_second}
\end{align}
where the second inequality is due to Assumption~\ref{assump:homogeneous}. for the ease of presentation, we assume the condition $c_{1} \frac{\eta}{\beta^{t}} \|\theta^{t} - \theta^{*}\|^{\alpha} \leq 1$ holds and we will demonstrate it later. From inequality~\eqref{eq:key_inequality_second}, we now demonstrate that there exists universal constant $\bar{C}$ such that 
\begin{align}
    \delta_{t} := f(\theta^{t}) - f(\theta^{*}) \leq \bar{C} \beta^{t \frac{\alpha + 2}{\alpha}}. \label{eq:key_inequality_third}
\end{align}

Indeed, we can prove the inequality~\eqref{eq:key_inequality_third} via an induction argument. When $t = 0$, the inequality~\eqref{eq:key_inequality_third} is satisfied as long as $\bar{C} \geq \delta_{0}$. We assume that inequality~\eqref{eq:key_inequality_third} holds for $t$. Now, we would like to argue that it also holds for $t + 1$. In fact, if $\delta_{t} \leq \bar{C} \beta^{(t + 1) \frac{\alpha + 2}{\alpha}}$, then from inequality~\eqref{eq:key_inequality_second} we have $\delta_{t + 1} \leq \delta_{t} \leq \bar{C} \beta^{(t + 1) \frac{\alpha + 2}{\alpha}}$. Hence, inequality~\eqref{eq:key_inequality_third} holds for $t + 1$. If $\delta_{t} \geq \bar{C} \beta^{(t + 1) \frac{\alpha + 2}{\alpha}}$, we obtain from inequality~\eqref{eq:key_inequality_second} that
\begin{align*}
    \delta_{t + 1} \leq \bar{C} \beta^{t \frac{\alpha + 2}{\alpha}} \left(1 - \frac{c_{2} \eta}{2 \beta^{t}} \bar{C}^{\frac{\alpha}{\alpha + 2}} \beta^{t + 1} \right) = \bar{C} \beta^{t \frac{\alpha + 2}{\alpha}} \left(1 - \frac{c_{2} \eta \beta}{2} \bar{C}^{\frac{\alpha}{\alpha + 2}} \right).
\end{align*}

In order to have $\delta_{t + 1} \leq \bar{C} \beta^{(t + 1) \frac{\alpha + 2}{\alpha}}$, it is sufficient to have $1 - \frac{c_{2} \eta \beta}{2} \bar{C}^{\frac{\alpha}{\alpha + 2}} \leq \beta^{\frac{\alpha + 2}{\alpha}}$. This inequality is equivalent to choose $\bar{C}$ and $\eta$ such that
\begin{align}
    \eta \bar{C}^{\frac{\alpha}{\alpha + 2}} \geq \frac{2 (1 - \beta^{\frac{\alpha + 2}{\alpha}})}{c_{2} \beta}. \label{eq:key_inequality_fourth}
\end{align}
It indicates that as long as $\bar{C}$ and $\eta$ satisfy~\eqref{eq:key_inequality_fourth}, we obtain the conclusion of inequality~\eqref{eq:key_inequality_third} for $t  + 1$. 

Now we demonstrate that $ c_{1} \frac{\eta}{\beta^{t}} \|\theta^{t} - \theta^{*}\|^{\alpha} \leq 1$. Note that, the Assumption~\ref{assump:homogeneous} indicates that
\begin{align}
    c_{1} \frac{\eta}{\beta^{t}} \|\theta^{t} - \theta^{*}\|^{\alpha} \leq c_{1} \frac{(\alpha + 2)^{\alpha}}{c_{2}^{\alpha}} \frac{\eta}{\beta^{t}} (f(\theta^{t}) - f(\theta^{*}))^{\frac{\alpha}{\alpha + 2}}, \label{eq:key_inequality_fifth}
\end{align}
By plugging the inequality~\eqref{eq:key_inequality_third} into
inequality~\eqref{eq:key_inequality_fifth}, we obtain that
\begin{align*}
    c_{1} \frac{\eta}{\beta^{t}} \|\theta^{t} - \theta^{*}\|^{\alpha} \leq c_{1} \frac{(\alpha + 2)^{\alpha}}{c_{2}^{\alpha}} \bar{C}^{\frac{\alpha}{\alpha + 2}} \eta \beta^{t \left(\frac{\alpha (\alpha + 2)}{(\alpha + 2) \alpha} - 1\right)} = c_{1} \frac{(\alpha + 2)^{\alpha}}{c_{2}^{\alpha}} \bar{C}^{\frac{\alpha}{\alpha + 2}} \eta.
\end{align*}
It suggests that as long as we choose $\eta > 0$ such that
\begin{align}
    \eta \bar{C}^{\alpha/(\alpha + 2)} \leq \frac{c_{2}^{\alpha}}{c_{1} (\alpha + 2)^{\alpha}}, \label{eq:key_inequality_sixth}
\end{align}
then we can guarantee that $c_{1} \frac{\eta}{\beta^{t}} \|\theta^{t} - \theta^{*}\|^{\alpha} \leq 1$. In summary, to guarantee the linear convergence of the objective value, the step size $\eta$ and the universal constant $\bar{C}$ need to satisfy inequalities~\eqref{eq:key_inequality_fourth} and~\eqref{eq:key_inequality_sixth}, namely, we need
\begin{align*}
    \frac{2 (1 - \beta^{\frac{\alpha + 2}{\alpha}})}{c_{2} \beta} \leq \eta \bar{C}^{\alpha/(\alpha + 2)} \leq \frac{c_{2}^{\alpha}}{c_{1} (\alpha + 2)^{\alpha}}.
\end{align*}
As a consequence, we obtain the conclusion of the theorem.
\subsection{Proof of Theorem~\ref{theorem:statistical_rate_EGD}}
\label{subsec:proof:theorem:statistical_rate_EGD}
To ease the proof presentation, we denote $\delta_{t} : = \|\theta_{n}^{t} - \theta^{*}\|$ for all $t \geq 0$. From the triangle inequality, we obtain that
\begin{align}
    \delta_{t + 1} = \|\theta_{n}^{t + 1} - \theta^{*}\| = \|F_{n}(\theta_{n}^{t}) - \theta^{*}\| & \leq \|F_{n}(\theta_{n}^{t}) - F(\theta_{n}^{t})\| + \|F(\theta_{n}^{t}) - \theta^{*}\| \nonumber \\
    & = \frac{\eta}{\beta^{t}} \|\nabla f_{n}(\theta_{n}^{t}) - \nabla f(\theta_{n}^{t})\| + \|F(\theta_{n}^{t}) - \theta^{*}\|. \label{eq:triangle_inequality}
\end{align}
According to the stability condition~\ref{assump:stab} of gradients between the sample and population loss functions , we have $\|\nabla f_{n}(\theta_{n}^{t}) - \nabla f(\theta_{n}^{t})\| \leq c \|\theta_{n}^{t} - \theta^{*}\|^{\gamma} \varepsilon(n, \delta)$ for some universal constant $c > 0$. Furthermore, direct calculation yields that
\begin{align*}
    \|F(\theta_{n}^{t}) - \theta^{*}\|^2 & = \left\|\theta_{n}^{t} - \frac{\eta}{\beta^{t}} \nabla f(\theta_{n}^{t}) - \theta^{*}\right\|^2 \\
    & = \|\theta_{n}^{t} - \theta^{*}\|^2 - 2 \frac{\eta}{\beta^{t}} (\theta_{n}^{t} - \theta^{*})^{\top} \nabla f(\theta_{n}^{t}) + \frac{\eta^2}{\beta^{2t}} \|\nabla f (\theta_{n}^{t}) \|^2. 
\end{align*}
As $f$ is a locally convex function in $\mathbb{B}(\theta^{*}, \rho)$, we find that
\begin{align*}
    f(\theta^{*}) \geq f(\theta_{n}^{t}) + (\theta^{*} - \theta_{n}^{t})^{\top} \nabla f(\theta_{n}^{t}). 
\end{align*}
Furthermore, based on the homogeneous assumption~\ref{assump:homogeneous} of population loss function, from Equation (33) in \cite{ren2021towards}, we have that
\begin{align*}
    \|\nabla f(\theta_{n}^{t})\| \leq \frac{c_1}{\alpha + 1} \|\theta_{n}^{t} - \theta^{*}\|^{\alpha + 1}.
\end{align*}
Putting the above results together, we obtain that
\begin{align*}
    \|F(\theta_{n}^{t}) - \theta^{*}\|^2 \leq \|\theta_{n}^{t} - \theta^{*}\|^2 - 2 \frac{\eta}{\beta^{t}} (f(\theta_{n}^{t}) - f(\theta^{*})) + \frac{\eta^2 c_1^2}{\beta^{2t}(\alpha + 1)^2} \|\theta_{n}^{t} - \theta^{*}\|^{2(\alpha + 1)}.
\end{align*}
Due to the homogeneous assumption~\ref{assump:homogeneous}, we also have $f(\theta_{n}^{t}) - f(\theta^{*}) \leq \frac{c_1}{(\alpha + 1)(\alpha + 2)}\|\theta_{n}^{t} - \theta^{*}\|^{\alpha + 2}$. Therefore, we find that
\begin{align}
    \|F(\theta_{n}^{t}) - \theta^{*}\|^2 & \leq \|\theta_{n}^{t} - \theta^{*}\|^2 \parenth{ 1 - \frac{2 \eta c_1}{\beta^{t}(\alpha + 1)(\alpha + 2)} \|\theta_{n}^{t} - \theta^{*}\|^{\alpha} + \frac{\eta^2 c_1^2}{\beta^{2t}(\alpha + 1)^2} \|\theta_{n}^{t} - \theta^{*}\|^{2 \alpha}} \nonumber \\
    & = \delta_{t}^{2} \parenth{1 - \frac{2 \eta c_1}{\beta^{t}(\alpha + 1)(\alpha + 2)} \delta_{t}^{\alpha} + \frac{\eta^2 c_1^2}{\beta^{2t}(\alpha + 1)^2} \delta_{t}^{2\alpha}}. \label{eq:key_inequality_statistical_rate_initial}
\end{align}
As long as we can choose $\eta$ and $\beta$ such that $\eta \frac{c_1(\alpha + 2)}{(\alpha + 1)} \delta_{t}^{\alpha} \leq \beta^{t}$, we have
\begin{align*}
    \|F(\theta_{n}^{t}) - \theta^{*}\|^2 \leq \delta_{t}^2 \parenth{ 1 - \frac{\eta c_1}{\beta^{t}(\alpha + 1)(\alpha + 2)} \delta_{t}^{\alpha}}.
\end{align*}
The condition $\eta \frac{c_1(\alpha + 2)}{(\alpha + 1)} \delta_{t}^{\alpha} \leq \beta^{t}$ will be revisited after we establish the bound on $\delta_{t}$ in equation~\eqref{eq:gamma_equal_one}. Collecting the above inequalities to the bound~\eqref{eq:triangle_inequality} leads to
\begin{align}
    \delta_{t + 1} \leq \frac{c \eta}{\beta^{t}} \delta_{t}^{\gamma} \varepsilon(n, \delta) + \delta_{t} \parenth{1 - \frac{\eta c_1}{\beta^{t}(\alpha + 1)(\alpha + 2)} \delta_{t}^{\alpha}}^{1/2}. \label{eq:key_inequality_statistical_rate}
\end{align}
We will first consider the setting $\gamma = 1$ to illustrate the proof technique to bound $\delta_{t}$ from equation~\eqref{eq:key_inequality_statistical_rate}. Then, we generalize that technique to general setting  $\gamma \geq 1$. 
\subsubsection{When $\gamma = 1$}
When $\gamma = 1$, we can rewrite equation~\eqref{eq:key_inequality_statistical_rate} as follows:
\begin{align}
    \delta_{t + 1} \leq \delta_{t} \parenth{\frac{c \eta}{\beta^{t}} \varepsilon(n, \delta) + \parenth{1 - \frac{\eta c_1}{\beta^{t}(\alpha + 1)(\alpha + 2)} \delta_{t}^{\alpha}}^{1/2}}. \label{eq:gamma_equal_one}
\end{align}
We will prove by induction that 
\begin{align}
    \delta_{t} \leq \bar{C}_{t} \beta^{\frac{t}{\alpha}} \label{eq:gamma_equal_one_first_bound}
\end{align}
for any $t \leq T : = \frac{- \log \varepsilon(n, \delta) + \log C_{1}}{\log(1/\beta)} - \frac{1}{\alpha}$ where $\bar{C}_{t + 1} = \bar{C}_{t} \parenth{1 + \frac{c \eta \varepsilon(n, \delta)}{\beta^{t + 1/\alpha}}}$ for any $t \geq 0$ and $C_{1}$ is some constant chosen later. Indeed, when $t = 0$, inequality~\eqref{eq:gamma_equal_one_first_bound} is obviously true when we choose $\bar{C}_{0}$ such that $\delta_{0} = \|\theta_{n}^{0} - \theta^{*}\| = \bar{C}_{0}$. Assume that the hypothesis holds for $t$. We will prove that it also holds for $t + 1$. Indeed, we have two possibilities. When $\delta_{t} \leq \bar{C}_{t} \beta^{\frac{t + 1}{\alpha}}$, then the bound in equation~\eqref{eq:gamma_equal_one} indicates that
\begin{align*}
    \delta_{t + 1} \leq \delta_{t} \parenth{\frac{c \eta}{\beta^{t}} \varepsilon(n, \delta) + 1} \leq \bar{C}_{t} \parenth{\frac{c \eta}{\beta^{t + 1/\alpha}} \varepsilon(n, \delta) + 1} \beta^{\frac{t + 1}{\alpha}} = \bar{C}_{t + 1} \beta^{\frac{t + 1}{\alpha}},
\end{align*}
which indicates that the inequality~\eqref{eq:gamma_equal_one_first_bound} holds for $t  + 1$. When $\delta_{t} > \bar{C}_{t} \beta^{\frac{t + 1}{\alpha}}$, the bound in equation~\eqref{eq:gamma_equal_one} leads to
\begin{align*}
    \delta_{t + 1} \leq & \bar{C}_{t} \beta^{\frac{t}{\alpha}} \parenth{\frac{c \eta}{\beta^{t}} \varepsilon(n, \delta) + \parenth{1 - \frac{\eta c_1}{\beta^{t}(\alpha+1)(\alpha + 2)} \bar{C}_{t}^{\alpha} \beta^{t+1}}^{1/2}} \\
    = & \bar{C}_{t} \beta^{\frac{t}{\alpha}} \parenth{\frac{c \eta}{\beta^{t}} \varepsilon(n, \delta) + \left(1 -  \frac{\eta c_1}{(\alpha + 1)(\alpha + 2)} \bar{C}_{t}^{\alpha} \beta\right)^{1/2}}. 
\end{align*}
As long as we can choose $\beta$ and $\eta$ such that $\left(1 - \eta \frac{c_1}{(\alpha + 1)(\alpha + 2)} \bar{C}_{t}^{\alpha} \beta\right)^{1/2} \leq \beta^{1/\alpha}$, then the inequality in the above display becomes
\begin{align*}
    \delta_{t + 1} \leq \bar{C}_{t} \beta^{\frac{t}{\alpha}} \parenth{\frac{c \eta}{\beta^{t}} \varepsilon(n, \delta) + \beta^{1/\alpha}} = \bar{C}_{t + 1} \beta^{\frac{t + 1}{\alpha}}.
\end{align*}
Since $\bar{C}_{t} \geq \bar{C}_{0}$ for all $t \leq T$, the inequality $\left(1 - \eta \frac{c_1}{(\alpha + 1)(\alpha + 2)} \bar{C}_{t}^{\alpha} \beta\right)^{1/2} \leq \beta^{1/\alpha}$ holds as long as $\left(1 - \eta \frac{c_1}{(\alpha + 1)(\alpha + 2)}\bar{C}_{0}^{\alpha} \beta\right)^{1/2} \leq \beta^{1/\alpha}$, which is valid based on our hypothesis. As a consequence, we obtain the conclusion of the induction argument. 

Now, we would like to upper bound $\bar{C}_{T}$ when $t = T$. Indeed, from the formulation of $\bar{C}_{T}$, we have
\begin{align*}
    \bar{C}_{T} = \bar{C}_{0} \prod_{j = 0}^{T - 1} \parenth{1 + \frac{c \eta \varepsilon(n, \delta)}{\beta^{j + 1/\alpha}}}.
\end{align*}
Since $\log(\cdot)$ is a concave function in the domain $(0, \infty)$, an application of Jensen inequality leads to
\begin{align*}
    \frac{1}{T} \sum_{j = 0}^{T - 1} \log \parenth{1 + \frac{c \eta \varepsilon(n, \delta)}{\beta^{j + 1/\alpha}}} & \leq \log \parenth{\frac{1}{T} \sum_{j = 0}^{T - 1} \parenth{1 + \frac{c \eta \varepsilon(n, \delta)}{\beta^{j + 1/\alpha}}}} \\
    & \leq \log \parenth{1 + \frac{c \eta \varepsilon(n, \delta)}{(1 - \beta) T \beta^{T - 1 + 1/ \alpha}}} \\
    & \leq \log \parenth{1 + \frac{\beta c \eta}{C_{1}(1 - \beta) T}},
\end{align*}
where the final inequality is due to $T  = \frac{- \log \varepsilon(n, \delta) + \log C_{1}}{\log(1/\beta)} - \frac{1}{\alpha}$, which indicates that $\beta^{T + 1/ \alpha} = C_{1} \varepsilon(n, \delta)$. Collecting these results, we obtain that
\begin{align}
    \bar{C}_{T} \leq \bar{C}_{0} \parenth{1 + \frac{\beta c \eta}{C_{1} (1 - \beta) T}}^{T} \leq \bar{C}_{0} \exp \parenth{\frac{\beta c \eta}{C_{1} (1 - \beta)}}. \label{eq:gamma_equal_one_second_bound}
\end{align}
Given the bound in equation~\eqref{eq:gamma_equal_one_first_bound}, the condition $\eta \frac{c_1(\alpha + 2)}{(\alpha + 1)} \delta_{t}^{\alpha} \leq \beta^{t}$ before equation~\eqref{eq:key_inequality_statistical_rate} is satisfied as long as $\eta \frac{c_1(\alpha + 2)}{(\alpha + 1)} \bar{C}_{T}^{\alpha} \beta^{t} \leq \beta^{t} $, which is equivalent to $\eta \frac{c_1(\alpha + 2)}{(\alpha + 1)} \bar{C}_{T}^{\alpha} \leq 1$. Given the bound of $\bar{C}_{T}$ in equation~\eqref{eq:gamma_equal_one_second_bound}, the previous bound is sufficient as long as we choose $C_{1}, \eta, \beta$ such that
\begin{align*}
    \frac{\eta c_1(\alpha + 2)}{(\alpha + 1)} \bar{C}_{0}^{\alpha} \exp \parenth{\frac{\alpha \beta c \eta}{C_{1} (1 - \beta)}} \leq 1.
\end{align*}
We can choose $C_{1} = \frac{\alpha \beta c  \eta}{1 - \beta}$ and the above inequality is valid from the hypothesis. 
Now, combining the results from equations~\eqref{eq:gamma_equal_one_first_bound} and~\eqref{eq:gamma_equal_one_second_bound}, we arrive at the following bound
\begin{align*}
     \min_{1 \leq t \leq T} \|\theta_{n}^{t} - \theta^{*}\| = \min_{1 \leq t \leq T} \delta_{t} \leq \bar{C}_{T} \beta^{\frac{T}{\alpha}} & \leq \bar{C}_{0} \exp \parenth{\frac{\beta c \eta}{C_{1} (1 - \beta)}} \parenth{C_{1} \beta^{-1/\alpha}  \varepsilon(n, \delta)}^{1/\alpha}. \\
     & = \bar{C}_{0} \exp (1/ \alpha) \parenth{\frac{\alpha \beta^{1 - 1/\alpha} c \eta}{(1 - \beta)}}^{1/\alpha} \varepsilon(n, \delta)^{1/\alpha}.
\end{align*}
As a consequence, we obtain the conclusion of the theorem when $\gamma = 1$.
\subsubsection{When $\gamma > 1$}
When $\gamma > 1$, equation~\eqref{eq:key_inequality_statistical_rate} becomes
\begin{align}
    \delta_{t + 1} \leq \delta_{t} \parenth{\frac{c \eta}{\beta^{t}} \delta_{t}^{\gamma - 1} \varepsilon(n, \delta) + \parenth{1 - \frac{\eta c_1}{\beta^{t}(\alpha + 1)(\alpha + 2)} \delta_{t}^{\alpha}}^{1/2}}. \label{eq:gamma_bigger_one}
\end{align}
Similar to the proof when $\gamma = 1$, we will prove by induction that
\begin{align}
    \delta_{t} \leq \widetilde{C}_{t} \beta^{\frac{t}{\alpha}} \label{eq:gamma_bigger_one_first_bound}
\end{align}
for any $t \leq \tilde{T} : = \frac{- \log \varepsilon(n, \delta) + \log C_{2}}{(\alpha + 1 - \gamma) \log(1/\beta)} - \frac{1}{\alpha (\alpha + 1 - \gamma)}$ where we define $$\widetilde{C}_{t + 1} = \widetilde{C}_{t} \parenth{1 + c \eta \varepsilon(n, \delta)\widetilde{C}_{t}^{\gamma - 1}\beta^{\frac{t (\gamma - 1 - \alpha)}{\alpha} - 1/\alpha}}$$ for any $t \geq 0$ and $C_{2}$ is some constant chosen later. When $t = 0$, the hypothesis is true as long as we choose $\widetilde{C}_{0} = \|\theta_{n}^{0} - \theta^{*}\|$. We assume that inequality~\eqref{eq:gamma_bigger_one_first_bound} is true for $t$. We will argue that this inequality is also true for $t + 1$. Indeed, we have two different cases. When $\delta_{t} \leq \widetilde{C}_{t} \beta^{\frac{t + 1}{\alpha}}$, then from inequality~\eqref{eq:gamma_bigger_one} we obtain that
\begin{align*}
    \delta_{t + 1} \leq \widetilde{C}_{t} \beta^{\frac{t + 1}{\alpha}} \parenth{\frac{c \eta}{\beta^{t}} \widetilde{C}_{t}^{\gamma - 1} \beta^{\frac{t(\gamma - 1)}{\alpha}} \varepsilon(n, \delta) + 1} = \widetilde{C}_{t + 1} \beta^{\frac{t + 1}{\alpha}}. 
\end{align*}
Therefore, the inequality~\eqref{eq:gamma_bigger_one_first_bound} holds for $t + 1$. When $\delta_{t} \in (\widetilde{C}_{t} \beta^{\frac{t + 1}{\alpha}}, \widetilde{C}_{t} \beta^{\frac{t}{\alpha}}]$, we find that
\begin{align*}
    \delta_{t + 1} \leq \widetilde{C}_{t} \beta^{\frac{t}{\alpha}} \parenth{\frac{c \eta}{\beta^{t}} \widetilde{C}_{t}^{\gamma - 1} \beta^{\frac{t(\gamma - 1)}{\alpha}} \varepsilon(n, \delta) + \parenth{1 - \frac{\eta c_1}{(\alpha + 1)(\alpha + 2)} \widetilde{C}_{t}^{\alpha} \beta}^{1/2}}.
\end{align*}
Since $\widetilde{C}_{t} \geq \widetilde{C}_{0}$ for all $t \leq \tilde{T}$, we have $\parenth{1 - \frac{\eta c_1}{(\alpha + 1)(\alpha + 2)} \widetilde{C}_{t}^{\alpha} \beta}^{1/2} \leq \parenth{1 - \frac{\eta c_1}{(\alpha + 1)(\alpha + 2)}  \widetilde{C}_{0}^{\alpha} \beta}^{1/2} \leq \beta^{1/\alpha}$ where the final inequality is based on the hypothesis. Given these bounds, we obtain that
\begin{align*}
    \delta_{t + 1} \leq \widetilde{C}_{t} \beta^{\frac{t}{\alpha}} \parenth{\frac{c \eta}{\beta^{t}} \widetilde{C}_{t}^{\gamma - 1} \beta^{\frac{t(\gamma - 1)}{\alpha}} \varepsilon(n, \delta) + \beta^{1/\alpha}} & = \widetilde{C}_{t} \beta^{\frac{t + 1}{\alpha}} \parenth{\frac{c \eta}{\beta^{t + 1/\alpha}} \widetilde{C}_{t}^{\gamma - 1} \beta^{\frac{t(\gamma - 1)}{\alpha}} \varepsilon(n, \delta) + 1} \\
    & = \widetilde{C}_{t + 1} \beta^{\frac{t + 1}{\alpha}}.
\end{align*}
Hence, we obtain the conclusion of inequality~\eqref{eq:gamma_bigger_one_first_bound} for $t + 1$. As a consequence, we reach the conclusion of the induction argument.

We now proceed to bound $\widetilde{C}_{\tilde{T}}$. We will prove by induction that
\begin{align}
\widetilde{C}_{t} \leq 3 \widetilde{C}_{0} \label{eq:gamma_bigger_one_second_bound}
\end{align}
for all $0 \leq t \leq \tilde{T}$ when we choose $C_{2} = \frac{3^{\gamma - 1} c \cdot \widetilde{C}_{0}^{\gamma - 1} \eta}{(1 - \beta^{\frac{\alpha + 1 - \gamma}{\alpha}})}$. Indeed, the bound~\eqref{eq:gamma_bigger_one_second_bound} holds when $t = 0$. We assume that this bound holds for $t$ for $0 \leq t \leq \tilde{T} - 1$. We will demonstrate that this bound also holds for $t + 1$. From the formulation of $\widetilde{C}_{t + 1}$, we find that
\begin{align*}
    \widetilde{C}_{t + 1} & = \widetilde{C}_{0} \prod_{j = 0}^{t} \parenth{1 + c \eta \varepsilon(n, \delta)\widetilde{C}_{j}^{\gamma - 1}\beta^{\frac{j (\gamma - 1 - \alpha)}{\alpha} - 1/\alpha}} \\
    & \leq \widetilde{C}_{0} \prod_{j = 0}^{t} \parenth{1 + 3^{\gamma - 1} c \eta \varepsilon(n, \delta)\widetilde{C}_{0}^{\gamma - 1}\beta^{\frac{j (\gamma - 1 - \alpha)}{\alpha} - 1/\alpha}}
\end{align*}
An application of Jensen inequality leads to
\begin{align*}
    \log(\widetilde{C}_{t + 1}) & = \log(\widetilde{C}_{0}) + \sum_{j = 0}^{t} \log \parenth{1 + 3^{\gamma - 1} c \eta \varepsilon(n, \delta)\widetilde{C}_{0}^{\gamma - 1}\beta^{\frac{j (\gamma - 1 - \alpha)}{\alpha} - 1/\alpha}} \\
    & \leq \log(\widetilde{C}_{0}) + (t + 1) \log \parenth{\frac{1}{t + 1} \sum_{j = 0}^{t} \parenth{1 + 3^{\gamma - 1} c \eta \varepsilon(n, \delta)\widetilde{C}_{0}^{\gamma - 1}\beta^{\frac{j (\gamma - 1 - \alpha)}{\alpha} - 1/\alpha}}} \\
    & \leq \log(\widetilde{C}_{0}) + (t + 1) \log \parenth{1 + \frac{3^{\gamma - 1} c \eta \varepsilon(n, \delta) \widetilde{C}_{0}^{\gamma - 1}}{(1 - \beta^{\frac{\alpha + 1 - \gamma}{\alpha}}) (t + 1)  \beta^{\frac{t(\alpha + 1 - \gamma)}{\alpha} + \frac{1}{\alpha}}} }.
\end{align*}
From the formulation of $\tilde{T}$, we have $\beta^{\frac{\tilde{T}(\alpha + 1 - \gamma)}{\alpha} + \frac{1}{\alpha}} = C_{2} \varepsilon(n, \delta)$. It indicates that $\beta^{\frac{t(\alpha + 1 - \gamma)}{\alpha} + \frac{1}{\alpha}} \geq C_{2} \varepsilon(n, \delta)$ for all $0 \leq t \leq \tilde{T} - 1$. Putting these bounds together, we have
\begin{align*}
    \log(\widetilde{C}_{t + 1}) & \leq \log(\widetilde{C}_{0}) + (t + 1) \log \parenth{1 + \frac{3^{\gamma - 1} c \eta \widetilde{C}_{0}^{\gamma - 1}}{C_{2} (1 - \beta^{\frac{\alpha + 1 - \gamma}{\alpha}}) (t + 1)}} \\
    & = \log(\widetilde{C}_{0}) + (t + 1) \log \parenth{1 + \frac{1}{t + 1}}.
\end{align*}
From this bound, we find that $\widetilde{C}_{t + 1} \leq \widetilde{C}_{0} \parenth{1 + \frac{1}{t + 1}}^{t + 1} \leq 3 \widetilde{C}_{0}$. Therefore, we obtain the bound~\eqref{eq:gamma_bigger_one_second_bound} for $t + 1$. As a consequence, the conclusion of the bound~\eqref{eq:gamma_bigger_one_second_bound} holds for any $0 \leq t \leq \tilde{T}$. 

We now revisit the condition $\eta \frac{c_1(\alpha + 2)}{(\alpha + 1)} \delta_{t}^{\alpha} \leq \beta^{t}$ before equation~\eqref{eq:key_inequality_statistical_rate}. Given that $\delta_{t}^{\alpha} \leq \widetilde{C}_{\tilde{T}}^{\alpha} \beta^{t}$ and $\widetilde{C}_{\tilde{T}} \leq 3 \widetilde{C}_{0}$, this condition is satisfied as long as $\frac{\eta c_1(\alpha + 2)}{(\alpha + 1)}3^{\alpha} \widetilde{C}_{0}^{\alpha} \beta^{t} \leq \beta^{t} $, which is equivalent to $\frac{\eta c_1(\alpha + 2)}{(\alpha + 1)} 3^{\alpha} \widetilde{C}_{0}^{\alpha} \leq 1$. This inequality is satisfied according to the hypothesis of the theorem.

Combining the results from equations~\eqref{eq:gamma_bigger_one} and~\eqref{eq:gamma_bigger_one_second_bound}, we find that
\begin{align*}
     \min_{1 \leq t \leq T} \|\theta_{n}^{t} - \theta^{*}\| = \min_{1 \leq t \leq T} \delta_{t} \leq \widetilde{C}_{\tilde{T}} \beta^{\frac{\tilde{T}}{\alpha}} \leq 3 \widetilde{C}_{0} \parenth{C_{2} \beta^{-\frac{1}{\alpha}} \varepsilon(n, \delta)}^{\frac{1}{\alpha + 1 - \gamma}},
\end{align*}
where $C_{2} = \frac{3^{\gamma - 1} c \cdot \widetilde{C}_{0}^{\gamma - 1} \eta}{(1 - \beta^{\frac{\alpha + 1 - \gamma}{\alpha}})}$. As a consequence, we obtain the conclusion of the theorem when $\gamma > 1$.
\subsection{Proof of Theorem~\ref{theorem:optimization_EGD_strongly_convex}}
\label{sec:proof:theorem:optimization_EGD_strongly_convex}
We utilize the similar proof argument as that of Theorem~\ref{theorem:homogeneous_setting}. In particular, from the updates of the EGD algorithm in equation~\eqref{eq:population_sample_exponential_GD}, we find that
\begin{align}
    f(\theta^{t + 1}) - f(\theta^{*}) & = f\left(\theta^{t} - \frac{\eta}{\beta^{t}} \nabla f(\theta^{t})\right) - f(\theta^{*}) \nonumber \\
    & \leq f(\theta^{t}) - f(\theta^{*}) - \frac{\eta}{\beta^{t}} \| \nabla f(\theta^{t})\|^2 + \frac{c_{1}}{2} \frac{\eta^2}{\beta^{2t}} \| \nabla f(\theta^{t})\|^2 \nonumber \\
    & = f(\theta^{t}) - f(\theta^{*}) - \frac{\eta}{\beta^{t}}\left(1 - \frac{c_{1}}{2} \frac{\eta}{\beta^{t}} \right) \| \nabla f(\theta^{t})\|^2. \label{eq:key_inequality_strongly_convex}
\end{align}
From the hypothesis, as we have $t \leq T = \log \parenth{\frac{2}{c_{1} \eta}}/ \log(1/ \beta)$, we can guarantee that $1 - \frac{c_{1} \eta}{2 \beta^{t}} \geq 0$. Then due to the local PL inequality in Assumption~\ref{assump:homogeneous} when $\alpha = 0$ we can further bound the RHS of inequality~\eqref{eq:key_inequality_strongly_convex} as follows:
\begin{align}
    f(\theta^{t + 1}) - f(\theta^{*}) & \leq \parenth{1 - \frac{c_{2}^2 \eta}{\beta^{t}} \parenth{1 - \frac{c_{1} \eta}{2\beta^{t}}}} \parenth{f(\theta^{t}) - f(\theta^{*})}. \label{eq:key_inequality_strongly_convex_first}
\end{align}
By repeating the inequality~\eqref{eq:key_inequality_strongly_convex_first} from 0 to $T = \log \parenth{\frac{2}{c_{1} \eta}}/ \log(1/ \beta)$, we have
\begin{align*}
    f(\theta^{T}) - f(\theta^{*}) \leq \parenth{\prod_{t = 0}^{T - 1} \parenth{1 - \frac{c_{2}^2 \eta}{\beta^{t}} + \frac{c_{1}c_{2}^2\eta^2}{2 \beta^{2t}}}} \parenth{f(\theta^{0}) - f(\theta^{*})}.
\end{align*}
As application of the Jensen inequality leads to
\begin{align*}
    \frac{1}{T} \sum_{t = 0}^{T - 1} \log \parenth{1 - \frac{c_{2}^2 \eta}{\beta^{t}} + \frac{c_{1}c_{2}^2\eta^2}{2 \beta^{2t}}} & \leq \log \parenth{\frac{1}{T} \sum_{t = 0}^{T - 1} \parenth{1 - \frac{c_{2}^2 \eta}{\beta^{t}} + \frac{c_{1}c_{2}^2\eta^2}{2 \beta^{2t}}}} \\
    & = \log \parenth{1 - \frac{c_{2}^2 \eta}{T} \frac{(\beta^{-T} - 1)}{(\beta^{-1} - 1)} + \frac{c_{1} c_{2}^2 \eta^2}{2 T} \frac{(\beta^{-2T} - 1)}{(\beta^{-2} - 1)}} \\
    & = \log \parenth{1 - \frac{c_{2}^2 \eta}{T} \frac{((c_{1}\eta/ 2)^{-1} - 1)}{(\beta^{-1} - 1)} + \frac{c_{1} c_{2}^2 \eta^2}{2 T} \frac{((c_{1} \eta/ 2)^{-2} - 1)}{(\beta^{-2} - 1)}} \\
    & = \log \parenth{1 - \frac{c_{2}^2 \eta}{T} \frac{((c_{1}\eta/ 2)^{-1} - 1)(\beta^{-1} - c_{1} \eta/ 2)}{(\beta^{-2} - 1)}}.
\end{align*}
Putting these results together, we find that
\begin{align*}
    f(\theta^{T}) - f(\theta^{*}) & \leq \parenth{1 - \frac{c_{2}^2 \eta}{T} \frac{((c_{1}\eta/ 2)^{-1} - 1)(\beta^{-1} - c_{1} \eta/ 2)}{(\beta^{-2} - 1)}}^{T} \parenth{f(\theta^{0}) - f(\theta^{*})} \\
    & \leq \exp \parenth{-\frac{c_{2}^2 \eta ((c_{1}\eta/ 2)^{-1} - 1)(\beta^{-1} - c_{1} \eta/ 2)}{(\beta^{-2} - 1)}} \parenth{f(\theta^{0}) - f(\theta^{*})},
\end{align*}
where the final inequality is due to the fact that $T \geq \frac{c_{2}^2 \eta ((c_{1}\eta/ 2)^{-1} - 1)(\beta^{-1} - c_{1} \eta/ 2)}{(\beta^{-2} - 1)}$. As a consequence, we obtain the conclusion of the theorem.
\subsection{Proof of Theorem~\ref{theorem:statistics_EGD_strongly_convex}}
\label{sec:proof:theorem:statistics_EGD_strongly_convex}
The proof of Theorem~\ref{theorem:statistics_EGD_strongly_convex} utilizes the proof argument of Theorem~\ref{theorem:statistical_rate_EGD}. In particular, we denote $\delta_{t} : = \|\theta_{n}^{t} - \theta^{*}\|$ for all $t \geq 0$. From the triangle inequality and the stability condition~\ref{assump:stab} with constant $\gamma = 0$, we obtain that
\begin{align}
    \delta_{t + 1} \leq \frac{c \eta}{\beta^{t}} \varepsilon(n, \delta) + \|F(\theta_{n}^{t}) - \theta^{*}\|, \label{eq:triangle_inequality_strongly_convex}
\end{align}
where $c$ is universal constant in Assumption~\ref{assump:stab}. Furthermore, with similar argument as that of the proof of Theorem~\ref{theorem:statistical_rate_EGD} to bound $\|F(\theta_{n}^{t}) - \theta^{*}\|^2$, we obtain that 
\begin{align}
    \|F(\theta_{n}^{t}) - \theta^{*}\|^2 & \leq \|\theta_{n}^{t} - \theta^{*}\|^2 \parenth{ 1 - \frac{\eta c_1}{\beta^{t}} + \frac{\eta^2 c_1^2}{\beta^{2t}}} = \delta_{t}^2 \parenth{ 1 - \frac{\eta c_1}{\beta^{t}} + \frac{\eta^2 c_1^2}{\beta^{2t}}}, \label{eq:key_inequality_statistical_rate_strongly_convex}
\end{align}
where $c_{1}$ is universal constant in the homogeneous assumption~\ref{assump:homogeneous}. Putting these results together, we find that
\begin{align*}
    \delta_{t + 1} \leq \frac{c \eta}{\beta^{t}} \varepsilon(n, \delta) + \delta_{t} \parenth{ 1 - \frac{\eta c_1}{\beta^{t}} + \frac{\eta^2 c_1^2}{\beta^{2t}}}^{1/2}.
\end{align*}
The recursive inequality leads to
\begin{align*}
    \delta_{t + 1} & \leq \frac{c \eta}{\beta^{t}} \varepsilon(n, \delta) + \parenth{\frac{c \eta}{\beta^{t - 1}} \varepsilon(n, \delta) + \delta_{t - 1} \parenth{ 1 - \frac{\eta c_1}{\beta^{t - 1}} + \frac{\eta^2 c_1^2}{\beta^{2(t - 1)}}}^{1/2}} \parenth{ 1 - \frac{\eta c_1}{\beta^{t}} + \frac{\eta^2 c_1^2}{\beta^{2t}}}^{1/2} \nonumber \\
    & \leq c \eta \varepsilon(n, \delta) \parenth{\frac{1}{\beta^{t}} + \frac{1}{\beta^{t - 1}}} + \delta_{t - 1} \parenth{ 1 - \frac{\eta c_1}{\beta^{t - 1}} + \frac{\eta^2 c_1^2}{\beta^{2(t - 1)}}}^{1/2} \parenth{ 1 - \frac{\eta c_1}{\beta^{t}} + \frac{\eta^2 c_1^2}{\beta^{2t}}}^{1/2}.
\end{align*}
By repeating the above inequality, we eventually obtain that
\begin{align*}
    \delta_{\bar{T}} & \leq c \eta \varepsilon(n, \delta) \parenth{\sum_{t = 0}^{\bar{T} - 1} \frac{1}{\beta^{t}}} + \delta_{0} \prod_{t = 0}^{\bar{T} - 1} \parenth{ 1 - \frac{\eta c_1}{\beta^{t}} + \frac{\eta^2 c_1^2}{\beta^{2t}}}^{1/2} \\
    & = c \eta \varepsilon(n, \delta) \frac{\beta^{-\bar{T}} - 1}{\beta^{-1} - 1} + \delta_{0} \prod_{t = 0}^{\bar{T} - 1} \parenth{ 1 - \frac{\eta c_1}{ \beta^{t}} + \frac{\eta^2 c_1^2}{\beta^{2t}}}^{1/2} \\
    & \leq c \eta \varepsilon(n, \delta) \frac{(\eta c_{1})^{-1} - 1)}{\beta^{-1} - 1} + \delta_{0} \exp \parenth{- \frac{(1 - \eta c_{1})(\beta^{-1} - c_{1} \eta)}{2(\beta^{-2} - 1)}},
\end{align*}
where the final inequality is due to $\beta^{T} = \eta c_{1}$ and the Jensen inequality, which indicates that
\begin{align*}
    \prod_{t = 0}^{\bar{T} - 1} \parenth{ 1 - \frac{\eta c_1}{ \beta^{t}} + \frac{\eta^2 c_1^2}{\beta^{2t}}}^{1/2} & \leq \parenth{1 -\frac{\eta c_{1}}{\bar{T}}\frac{(\beta^{-\bar{T}} - 1)}{(\beta^{-1}-1)} + \frac{\eta^2 c_{1}^2}{\bar{T}}\frac{(\beta^{-2\bar{T}} - 1)}{(\beta^{-2}-1)}}^{\bar{T}/2} \\
    & \leq \exp \parenth{- \frac{(1 - \eta c_{1})(\beta^{-1} - c_{1} \eta)}{2(\beta^{-2} - 1)}}.
\end{align*}
As a consequence, we obtain the conclusion of the theorem.
\section{Auxiliary Results}
\label{sec:auxiliary_results}
In this Appendix, we provide proofs for the remaining results in the paper.
\subsection{Proof of Corollary~\ref{cor:updates_EGD}}
\label{subsec:proof:cor:updates_EGD}
To prove Corollary~\ref{cor:updates_EGD}, it is sufficient to demonstrate that under Assumption~\ref{assump:homogeneous}, we have
\begin{align}
    \|\theta - \theta^{*}\|^{\alpha + 2} \leq \frac{(\alpha + 2)^{\alpha + 2}}{c_{2}^{\alpha + 2}} \parenth{f(\theta) - f(\theta^{*})}, \label{eq:objective_to_updates}
\end{align}
for any $\theta \in \mathbb{B}(\theta^{*}, \rho)$ where $\rho$ and $c_{2}$ are constants in Assumption~\ref{assump:homogeneous}. The proof of equation~\eqref{eq:objective_to_updates} follows from the idea of using gradient flow~\cite{bolte2017error}. In particular, we consider the following gradient flow:
\begin{align}
    \frac{d\theta(t)}{dt} = - \nabla f(\theta(t)). \label{eq:gradient_flow}
\end{align}
Direct calculation yields that
\begin{align*}
    \frac{d\|\theta(t) - \theta^*\|_2^2}{dt} = 2 \parenth{\theta(t) - \theta^*}^{\top} \frac{d\theta(t)}{dt}. 
\end{align*}
From the gradient flow~\eqref{eq:gradient_flow} and the convexity of the function $f$ in $\mathbb{B}(\theta^{*}, \rho)$, we obtain that
\begin{align*}
    \frac{d\|\theta(t) - \theta^*\|_2^2}{dt} = - 2 \parenth{\theta(t) - \theta^*}^{\top} \nabla f(\theta(t)) \leq 0.
\end{align*}
From this inequality, it indicates that $\theta(t) \in \mathbb{B}(\theta^{*}, \rho)$ as long as we have $\theta(0) \in \mathbb{B}(\theta^{*}, \rho)$. Therefore, to obtain the conclusion of inequality~\eqref{eq:objective_to_updates}, it is sufficient to demonstrate that
\begin{align*}
    \|\theta(0) - \theta^{*}\|^{\alpha + 2} \leq \frac{(\alpha + 2)^{\alpha + 2}}{c_{2}^{\alpha + 2}} \parenth{f(\theta(0)) - f(\theta^{*})}.
\end{align*}
Indeed, simple algebra indicates that
\begin{align*}
    \left(f(\theta(0)) - f(\theta^*)\right)^{\frac{1}{\alpha + 2}} = \int_{\infty}^{0}d(f(\theta(t)) - f(\theta^*))^{\frac{1}{\alpha + 2}}
    & \geq  \int_{0}^{\infty} \frac{c_2}{\alpha + 2}\|\nabla f(\theta(t))\| dt \\
    & = \int_{0}^\infty \frac{c_2}{\alpha + 2} \left\|\frac{d\theta(t)}{dt}\right\| dt\\
    & = \frac{c_2}{\alpha + 2}\|\theta(0) - \theta^*\|,
\end{align*}
where the second inequality is due to the generalized PL condition~\eqref{eq:generalized_PL} while the third equation is due to the formulation of the gradient flow~\eqref{eq:gradient_flow}. As a consequence, we obtain the conclusion of inequality~\eqref{eq:objective_to_updates}, which directly leads to the conclusion of Corollary~\ref{cor:updates_EGD}.
\section{Generalized Linear Models}
\label{sec:examples}

In this appendix, we provide an application of the general theory to establish the statistical behaviors of the EGD iterates for solving parameter estimation for generalized linear models~\cite{Carroll-1997} with polynomial link function. Here, we assume that $(Y_{1}, X_{1}), (Y_{2}, X_{2}), \ldots, (Y_{n}, X_{n})$ are i.i.d. samples from the following parametric model:
\begin{align}
    Y_{i} = (X_{i}^{\top}\theta^{*})^{p} + \varepsilon_{i}, \label{eq:generalized_linear_model}
\end{align}
and $\theta^{*}$ is a true but unknown parameter, $p \in \mathbb{N}$ is a given power, and $\varepsilon_{1}, \ldots, \varepsilon_{n}$ are independent random noise vectors with zero mean and variance $\sigma^2$. We consider the random design setting where $X_{1}, \ldots, X_{n}$ are i.i.d. samples from $\mathcal{N}(0, I_{d})$. When $p = 1$, the generalized linear model~\eqref{eq:generalized_linear_model} reduces to the traditional linear regression model. When $p = 2$, it becomes the phase retrieval problem~\cite{Fienup_82,Shechtman_Yoav_etal_2015, candes_2011,Netrapalli_Prateek_Sanghavi_2015}. To estimate $\theta^{*}$, we minimize the sample least-square loss function:
\vspace{-0.6em}
\begin{align}
    \min_{\theta \in \mathbb{R}^{d}} \overline{\mathcal{L}}_{n}(\theta) : = \frac{1}{n} \sum_{i = 1}^{n} \parenth{Y_{i} - (X_{i}^{\top} \theta)^p}^2. \label{eq:sample_likelihood_generalized_linear}
\end{align}
As highlighted in Section~\ref{sec:general_theory_statistics_rate_EGD}, a key ingredient to analyze the EGD iterates for solving the sample least-square loss function $\overline{\mathcal{L}}_{n}$ is to understand the homogeneity of the the corresponding population least-square loss function, which is given by:
\begin{align}
    \overline{\mathcal{L}}(\theta) : = \Exs \brackets{\overline{\mathcal{L}}_{n}(\theta)} = \Exs_{X} \brackets{(Y - (X^{\top} \theta)^{p})^2}, \label{eq:population_likelihood_generalized_linear}
\end{align}
where the first outer expectation is taken with respect to the data $(X_{1}, Y_{1}), \ldots, (X_{n}, Y_{n})$ and the second outer expectation is taken with respect to $(X, Y)$ such that $X \sim \mathcal{N}(0, I_{d})$ and $Y = (X^{\top} \theta^{*})^p + \varepsilon$ where $\varepsilon$ has zero mean and variance $\sigma^2$. 

There are three important regimes for the model~\eqref{eq:sample_likelihood_generalized_linear}: (1) High signal-to-noise ratio (SNR) regime when $\|\theta^{*}\|/ \sigma \geq C$ where $C$ is some universal constant; (2) Middle SNR regime when $c \leq \|\theta_{n}^{*} - \theta^{*}\| \leq \|\theta^{*}\|/ \sigma \leq C$ where $\theta_{n}^{*}$ is the optimal solution of the sample least-square problem~\eqref{eq:sample_likelihood_generalized_linear} and $c$ is some universal constant; (3) Low SNR regime where $\|\theta^{*}\|/ \sigma \leq c \|\theta_{n}^{*} - \theta^{*}\|$. In this appendix, we focus on the high and low SNR regimes while deferring the discussion of middle SNR regime to Appendix~\ref{sec:discussion_middle_SNR}.


\subsection{High SNR Regime} 
\textbf{Fixed-step size GD in the high SNR regime:} Under the high SNR regime, the population least-square function $\overline{\mathcal{L}}$ is locally strongly convex and smooth (see Section 3.1 in~\cite{ren2021towards}). Furthermore, there exist universal constants $C_{1}$ and $C_{2}$ such that with probability $1 - \delta$: $\sup_{\theta\in \mathbb{B}(\theta^*, r)} \|\nabla \overline{\mathcal{L}}_n(\theta) - \nabla \overline{\mathcal{L}}(\theta)\|\leq C_{2} \sqrt{\frac{d + \log(1/\delta)}{n}}$
for any $r > 0$ as long as $n \geq C_{1} (d \log(d/ \delta))^{2p}$. An application of the results for fast and stable operators from~\cite{Ho_Instability} indicate that the statistical rate of the GD algorithm is $\mathcal{O}((d/n)^{1/2})$ and this rate is achieved after $\mathcal{O}(\log(n/d))$ number of iterations. 

\vspace{0.5 em}
\noindent
\textbf{The EGD algorithm with fixed $\beta$ in the high SNR regime:} Under the strong signal-to-noise regime, namely, $\|\theta^{*}\|/\sigma \geq C$ for some sufficiently large $C$, the population loss function satisfies the homogeneous assumption~\ref{assump:homogeneous} with constant $\alpha = 0$ (See Section 3.1 in~\cite{ren2021towards}). Furthermore, the stability assumption~\ref{assump:stab} is satisfied with $\gamma = 0$ and the noise function $\varepsilon(n, \delta) = \sqrt{\frac{d + \log(1/\delta)}{n}}$:
\begin{align*}
    \sup_{\theta\in \mathbb{B}(\theta^*, r)} \|\nabla \overline{\mathcal{L}}_n(\theta) - \nabla \overline{\mathcal{L}}(\theta)\|\leq c_3 \sqrt{\frac{d + \log(1/\delta)}{n}}
\end{align*}
for any $r > 0$ as long as the sample size $n \geq c_{4} (d \log(d/ \delta))^{2p}$. A direct application of Theorem~\ref{theorem:statistics_EGD_strongly_convex} for the statistical behaviors of the EGD algorithm leads to the following corollary.
\begin{corollary}
\label{corollary_generalized_linear_strong_signal}
In the strong signal-to-noise regime of the generalized linear model~\eqref{eq:generalized_linear_model}, i.e., $\|\theta^{*}\|/\sigma \geq C$ for some universal constant $C$, and the initialization $\theta_{n}^{0} \in \mathbb{B}(\theta^{*}, \rho)$ for some $\rho > 0$, there exist universal constants $C_{1}$ and $C_{2}$ such that as long as $n \geq C_{1} (d \log(d/ \delta))^{2p}$, the step size $\eta < 1/ c_{1}$ and $\bar{T} = \log(1/(\eta c_{1}))/ \log(1/\beta)$ where $c_{1}$ is the universal constant in the homogeneous assumption~\ref{assump:homogeneous}, with probability $1 - \delta$ we find that
\begin{align*}
    \min_{1 \leq t \leq \bar{T}} \|\theta_{n}^{t} - \theta^{*}\| \leq C_{2} \parenth{\frac{d}{n}}^{1/2} + \exp \parenth{- \frac{(1 - \eta c_{1})(\beta^{-1} - c_{1} \eta)}{2(\beta^{-2} - 1)}} \|\theta_{n}^{0} - \theta^{*}\|.
\end{align*}
\end{corollary}
As indicated in Section~\ref{sec:stats_rate_local_PL}, $C_{3} \parenth{\frac{d}{n}}^{1/2}$ and $\exp \parenth{- \frac{(1 - \eta c_{1})(\beta^{-1} - c_{1} \eta)}{2(\beta^{-2} - 1)}} \|\theta_{n}^{0} - \theta^{*}\|$ are respectively the statistical and optimization errors of the sample EGD iterates. As the optimization error does not vanish to 0, within the first $\bar{T}$ iterations the sample EGD iterates only converge to a neighborhood of $\theta^{*}$ with radius being the sum of the statistical and optimization errors. After $\bar{T}$ iterations, the sample EGD iterates diverge. On the other hand, when the GD iterates converge to the final statistical radius $\mathcal{O}(\sqrt{d/n})$ within the true parameter $\theta^{*}$ after $\mathcal{O}(\log(n/d))$ iterations.

\begin{figure}[!t]
    \centering
    \includegraphics[width=0.48\textwidth]{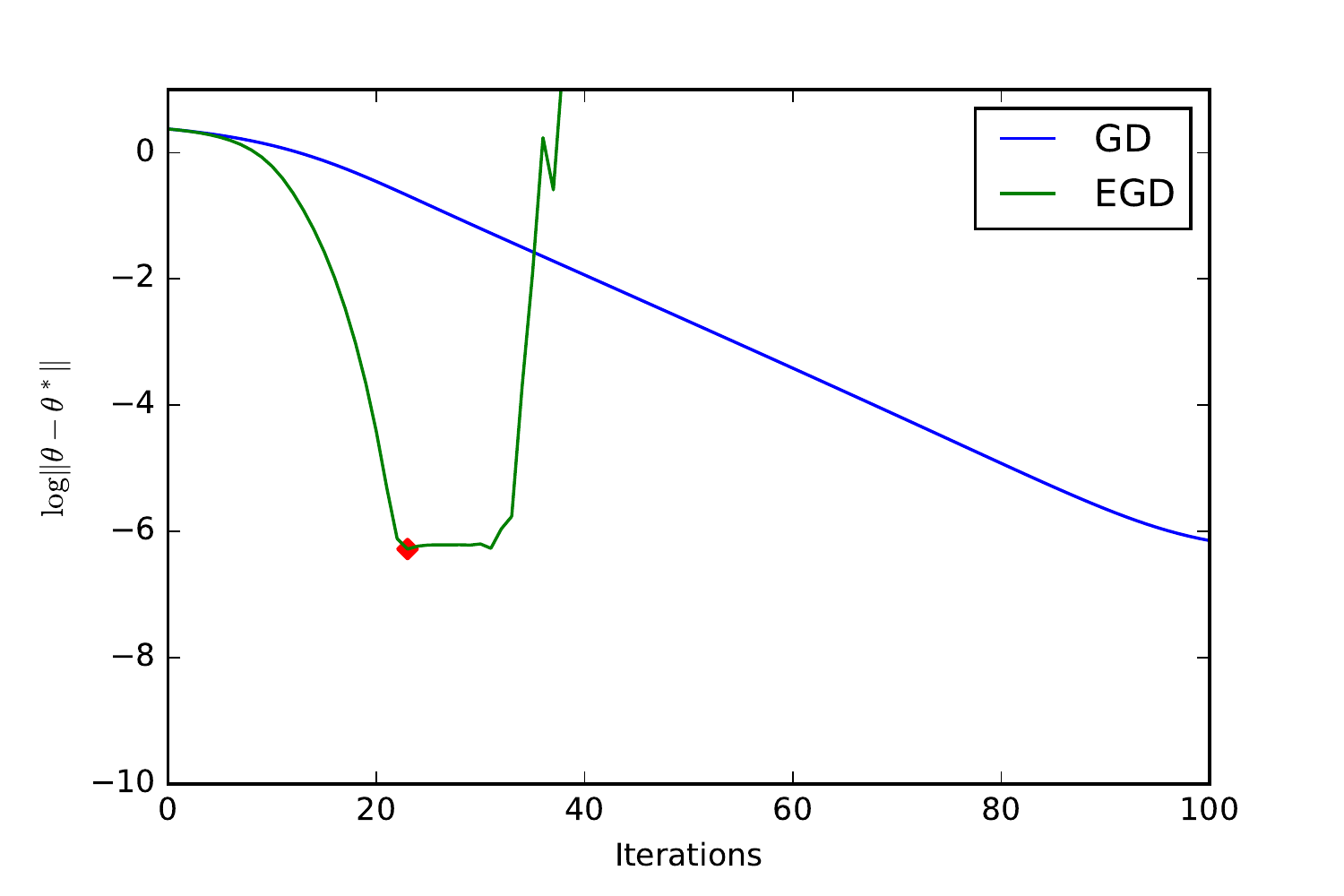}
    \includegraphics[width=0.48\textwidth]{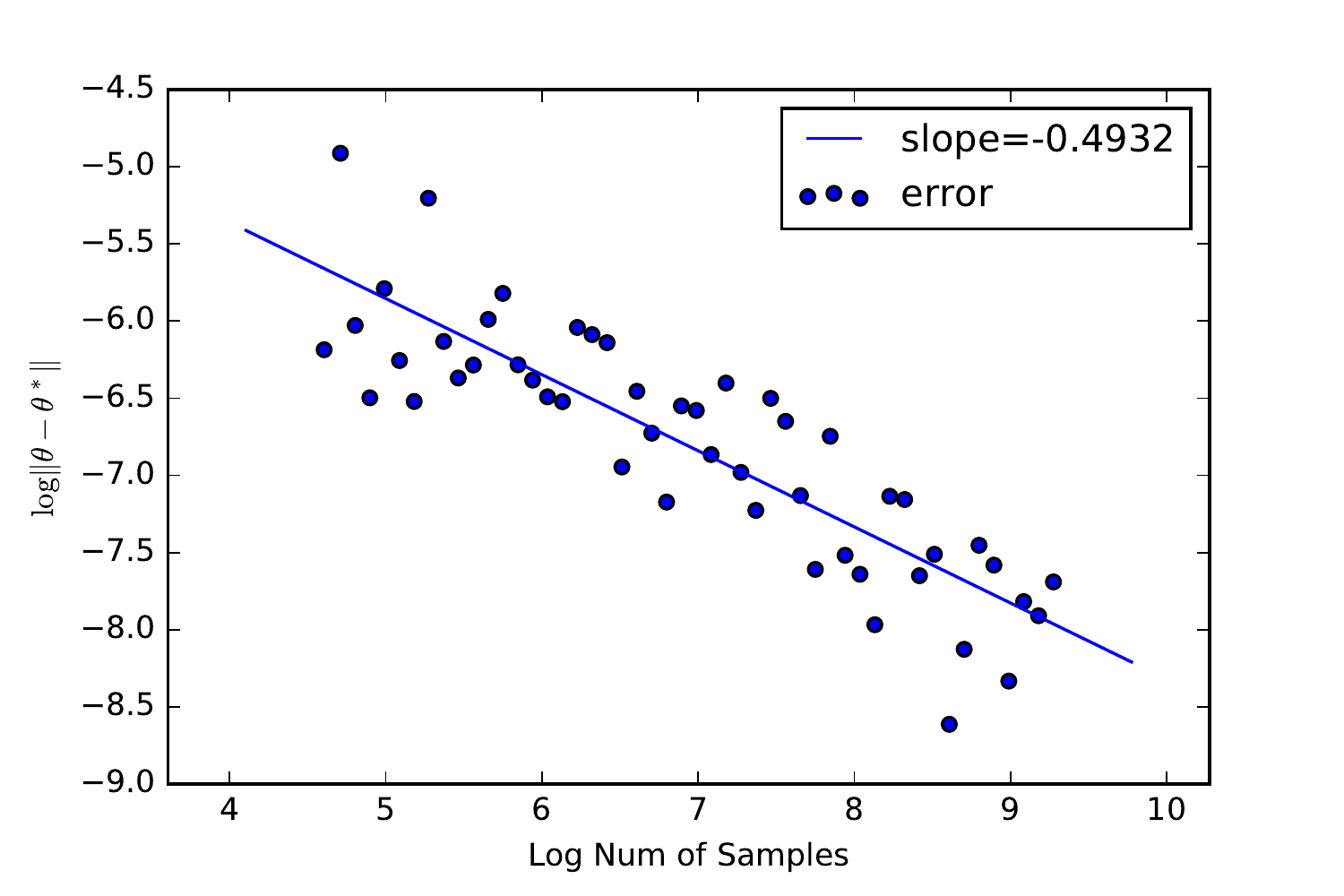}
    \caption{\textit{High SNR regime for GLM: Computational complexity and statistical rates of the GD and EDG iterates for solving parameter estimation under the high SNR regime of the generalized linear model~\eqref{eq:generalized_linear_model} when $d = 4$ and $p = 2$}. \textbf{Left:} The GD iterates converge to the statistical radius linearly, while the EGD iterates converge to the statistical radius with a linear rate faster than GD iterates, then diverge. The red diamond shows the EGD iterate with the minimum validation loss when we use 90\% of the data to compute the gradient for the GD and EGD iterates, and use the remaining 10\% of the data to perform cross validation. \textbf{Right:} Both GD and EGD can find a solution within the statistical radius $\mathcal{O}(n^{-1/2})$.}
    \label{fig:GLM_Strong_SNR}
\end{figure}

\vspace{0.5 em}
\noindent
\textbf{The EGD algorithm with sample size dependent $\beta$ in the high SNR regime:} In Corollary~\ref{corollary_generalized_linear_strong_signal}, when the scaling parameter $\beta$ is fixed, the appearance of the non-vanishing optimization error $\exp \parenth{- \frac{(1 - \eta c_{1})(\beta^{-1} - c_{1} \eta)}{2(\beta^{-2} - 1)}} \|\theta_{n}^{0} - \theta^{*}\|$ can be undesirable. When we choose $\beta^2 = 1 - \frac{C}{\log (n/d)}$ for some constant $C$, under the high SNR regime of the generalized linear models~\eqref{eq:generalized_linear_model} the EGD iterates are able to reach to the final statistical radius $\mathcal{O}((d/n)^{1/2})$ (up to some logarithmic factor) after $\mathcal{O}(\log(n/d))$ logarithmic number of iterations, which is comparable to the GD iterates. We summarize this result in the following corollary.
\begin{corollary}
\label{corollary_generalized_linear_varied_beta_high_SNR}
Given the generalized linear models~\eqref{eq:generalized_linear_model} and the initialization $\theta_{n}^{0} \in \mathbb{B}(\theta^{*}, \rho)$ for some radius $\rho > 0$, by choosing the scaling parameter $\beta$ such that 
\begin{align}
\beta^2 = 1 - \frac{(1 - \eta c_{1})^2}{2 \log (n/d)} \label{eq:varied_beta_value}
\end{align}
where $c_{1}$ is the universal constant in the homogeneous assumption~\ref{assump:homogeneous}, there exist universal constants $\{C_{i}\}_{i = 1}^{3}$ such that as long as $\|\theta^{*}\|/ \sigma \geq C_{1}$, the sample size $n \geq C_{2}(d \log(d/\delta))^{2p}$, and $t \geq C_{3} \log(n/(d + \log(1/\delta))$ with probability $1 - \delta$ the following holds:$$\min_{1 \leq k \leq t} \|\theta_{n}^{t} - \theta^{*}\| \leq C_{4} \log \parenth{\frac{n}{d + \log(1/\delta)}}\parenth{\frac{d + \log(1/\delta)}{n}}^{1/2}.$$
\end{corollary}
\vspace{0.5 em}
\noindent
\textbf{Experimental results:} To illustrate the statistical behaviors of the EGD iterates, we consider the parameter estimation of these iterates in the phase retrieval setting, which corresponds to the case where the power of the link function is chosen to be $p = 2$. We run both the EGD and GD algorithm for solving the sample least-square loss function $\overline{\mathcal{L}}_{n}$ in equation~\eqref{eq:sample_likelihood_generalized_linear}. For both the strong and low SNR regimes, we use $d = 4$, $\beta = 0.9$ and $\eta = 0.001$. The experiment results are shown in Figures~\ref{fig:GLM_Low_SNR} and~\ref{fig:GLM_Strong_SNR}. To illustrate that we can obtain the early stopping behavior of the EGD iterates under both the strong and low SNR regimes via cross-validation, for all of the experiments we use 90\% of the data to compute the gradient for the GD and EGD iterates, and use the remaining 10\% of the data to perform cross validation. The red diamonds in Figures~\ref{fig:GLM_Low_SNR} and~\ref{fig:GLM_Strong_SNR} show the EGD iterate with the minimum validation loss. As we observe from these plots, the red diamonds align quite well with the final statistical radius under each regime of the generalized linear models.

\subsection{Low SNR Regime} To simplify the proof argument, we assume $\theta^{*} = 0$ for the low SNR regime. Under this setting, the population least-square loss function~\eqref{eq:population_likelihood_generalized_linear} admits the following closed-form expression: $\overline{\mathcal{L}}(\theta) : = \frac{\sigma^2 + (2p - 1)!! \|\theta - \theta^{*}\|^{2p}}{2}$.

From simple calculations, the population least-square loss function satisfies the homogeneous Assumption~\ref{assump:homogeneous} when $\alpha = 2p - 2$ for some radius $\rho > 0$, namely, we have
\begin{align}
    \lambda_{\max}(\nabla^2 \overline{\mathcal{L}}(\theta)) \leq c_1 \|\theta - \theta^*\|^{2p - 2}, \quad \quad \|\nabla \overline{\mathcal{L}}(\theta)\| \geq c_2(\overline{\mathcal{L}}(\theta) - \overline{\mathcal{L}}(\theta^*))^{(2p - 1)/(2p)}, \label{eq:homogeneous_generalized_linear}
\end{align}
where $c_{1} = p(2p-1)(2p-1)!!$ and $c_{2} = 2p$. For the stability Assumption~\ref{assump:stab}, the result of Appendix A.2 in~\cite{mou2019diffusion} indicates that this assumption is satisfied with $\gamma = p - 1$ and the noise function $\varepsilon(n, \delta) = \sqrt{\frac{d + \log(1/\delta)}{n}}$, namely, the following uniform concentration bound holds:
\vspace{-0.5 em}
\begin{align}
    \sup_{\theta\in \mathbb{B}(\theta^*, r)} \|\nabla \overline{\mathcal{L}}_n(\theta) - \nabla \overline{\mathcal{L}}(\theta)\|\leq c_3 r^{p - 1} \sqrt{(d + \log(1/\delta))/n} \label{eq:concentration_generalized_linear}
\vspace{-0.6em}
\end{align}
for any $r > 0$ as long as the sample size $n \geq c_{4} (d \log(d/ \delta))^{2p}$. 

\vspace{0.5 em}
\noindent
\textbf{The EGD algorithm with fixed $\beta$ in the low SNR regime:} Based on the results from equations~\eqref{eq:homogeneous_generalized_linear} and~\eqref{eq:concentration_generalized_linear}, a direct application of Theorem~\ref{theorem:statistical_rate_EGD} leads to the following statistical and computational complexities for EGD in solving parameter estimation in generalized linear models~\eqref{eq:generalized_linear_model} when $\theta^{*} = 0$.
\begin{corollary}
\label{corollary_generalized_linear}
In the low SNR regime of the generalized linear model~\eqref{eq:generalized_linear_model}, i.e., $\theta^{*} = 0$, and the initialization $\theta_{n}^{0} \in \mathbb{B}(\theta^{*}, \rho)$ for some $\rho > 0$, the following holds:

(a) (GD algorithm) There exist universal constants $\bar{C}_{1}, \bar{C}_{2}, \bar{C}_{3}$ such that for any fixed $\tau \in \parenth{0, \frac{1}{2p}}$ as long as $n \geq \bar{C}_{1} (d \log(d/ \delta))^{2p}$, and $t \geq \bar{C}_{2} \parenth{n/(d + \log(1/\delta))}^{\frac{p - 1}{p}} \log \frac{1}{\tau}$, we have
\begin{align*}
    \|\theta_{n, \text{GD}}^t - \theta^*\| \leq \bar{C}_{3} \parenth{(d + \log(1/\delta))/n}^{\frac{1}{2p} - \tau}.
\end{align*}  
(b) (EGD algorithm) With the assumptions on $\beta$ and $\eta$ as those in Theorem~\ref{theorem:statistical_rate_EGD} there exist universal constants $C_{1}', C_{2}', C_{3}'$ such that as long as $n \geq C_{1}' (d \log(d/ \delta))^{2p}$, and $t \geq C_{2}' \log(n/ (d + \log(1/\delta))$, with probability $1 - \delta$ we find that
\begin{align*}
    \min_{1 \leq k \leq t} \|\theta_{n}^{k} - \theta^{*}\| \leq C_{3}'\parenth{(d + \log(1/\delta))/n}^{\frac{1}{2p}}.
\end{align*}
\end{corollary}

From Corollary~\ref{corollary_generalized_linear}, the EGD updates reach the final statistical radius $\mathcal{O}((d/n)^{1/(2p)})$ after $\mathcal{O}(\log(n/d))$ iterations. Since each iteration of EGD has a complexity of $\mathcal{O}(n d)$, this implies that the total computational complexity of the EGD algorithm for reaching the final statistical radius of the generalized linear models is $\mathcal{O}(n d \log(n/d))$. On the other hand, the final statistical radius of the GD iterates is also $\mathcal{O}((d/n)^{1/(2p)})$ and is achieved after $\mathcal{O}((n/d)^{(p-1)/p})$ iterations. When $p = 1$ (i.e., the phase retrieval setting), this result of the GD algorithm agrees with the result from~\cite{Candes_Phase_Retrieval}. Therefore, the total computational complexity of the EGD algorithm is much better than that of the GD algorithm, which is of the order $\mathcal{O}(n^{(2p - 1)/p})$ in terms of the sample size $n$. 
\begin{figure*}[!t]
    \centering
    \includegraphics[width=0.48\textwidth]{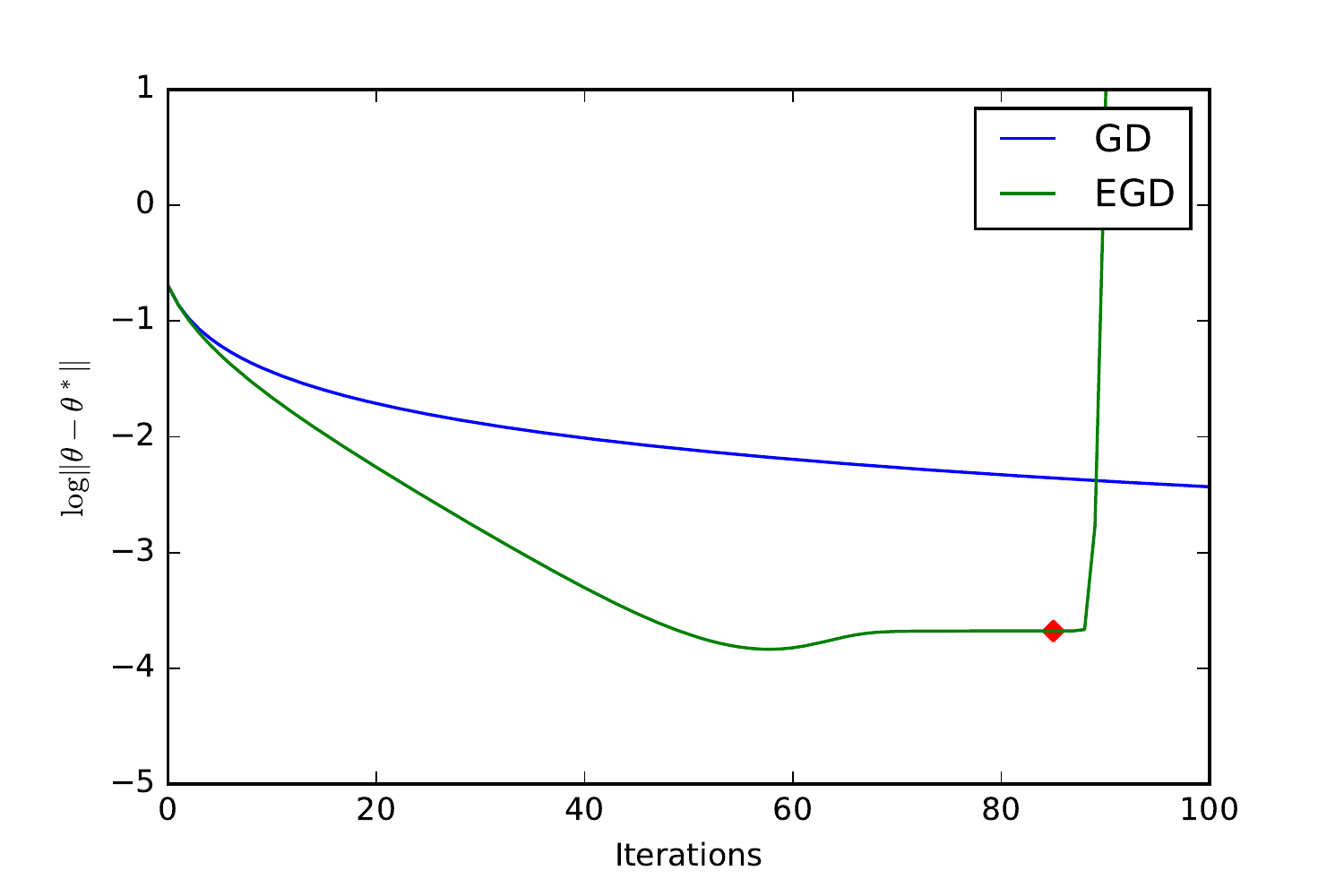}
    \includegraphics[width=0.48\textwidth]{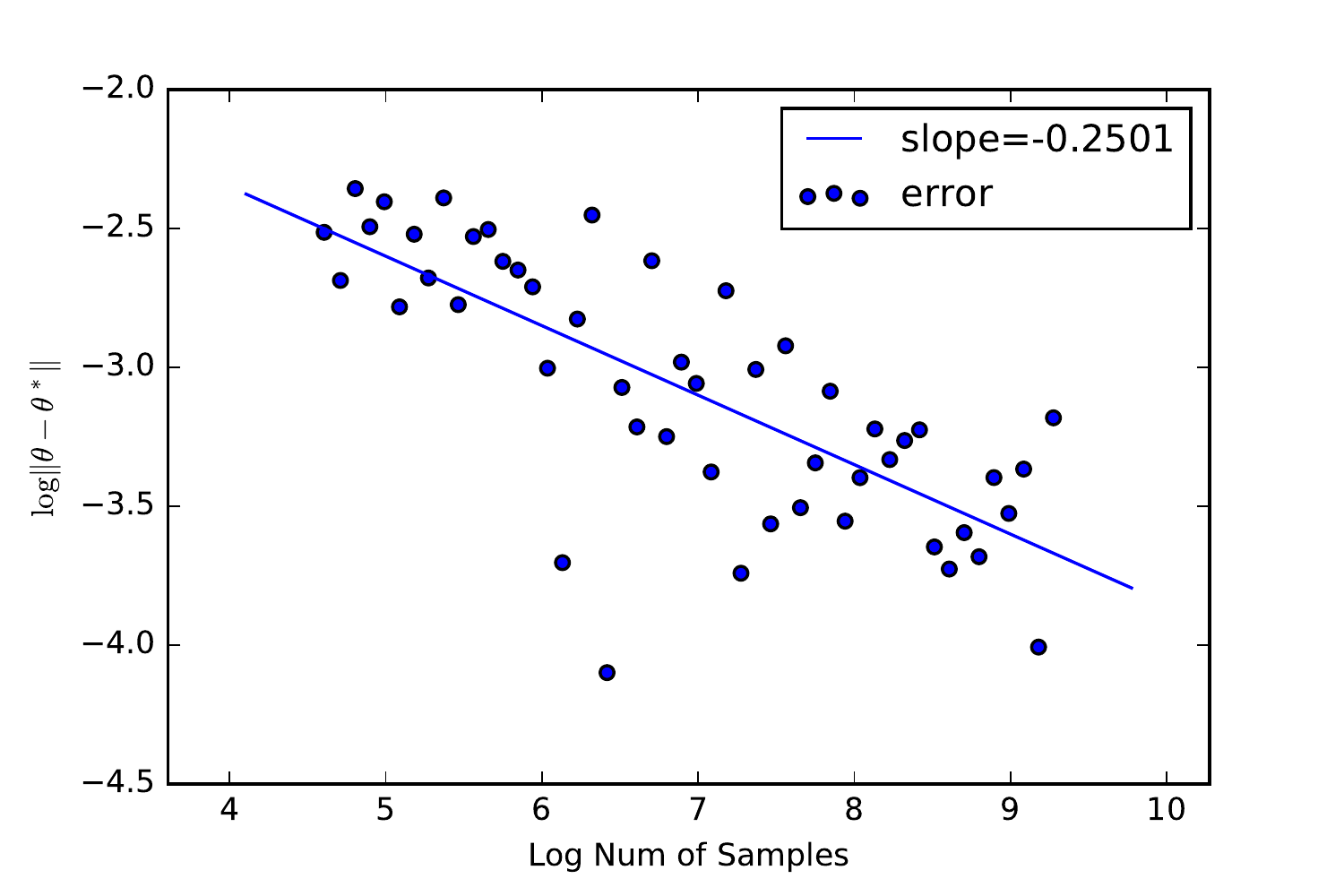}
    \caption{\textit{Computational complexity and statistical rates for GD and EGD iterates for solving parameter estimation in the low SNR regime of the generalized linear models~\eqref{eq:generalized_linear_model} when $d = 4$ and $p = 2$}. \textbf{Left:} The EGD iterates first converge to the statistical radius linearly then diverge, while the GD iterates converge to the statistical radius sub-linearly. The red diamond shows the EGD iterate with the minimum validation loss when we use 90\% of the data to compute the gradient for the GD and EGD iterates, and use the remaining 10\% of the data to perform cross validation. \textbf{Right:} Both GD and EGD can find a solution within the statistical radius $\mathcal{O}(n^{-1/4})$.}
    \label{fig:GLM_Low_SNR}
\end{figure*}

\vspace{0.5 em}
\noindent
\textbf{The EGD algorithm with sample size dependent $\beta$ in the low SNR regime:} Under the low SNR regime of these models, Proposition~\ref{proposition:rate_EGD_varied_beta} in Section~\ref{sec:varied_beta} demonstrates that the EGD iterates still reach the final statistical radius after $\mathcal{O}(\log^2(n/d))$ number of iterations, which is slightly more expensive than the number of iterations $\mathcal{O}(\log(n/d))$ of the EGD iterates when we fix $\beta$ in Corollary~\ref{corollary_generalized_linear} and is still much cheaper than the number of iterations $\mathcal{O}((n/d)^{(p - 1)/p})$ of the GD iterates. 
\begin{corollary}
\label{corollary_generalized_linear_varied_beta_low_SNR}
Given the generalized linear models~\eqref{eq:generalized_linear_model} and the initialization $\theta_{n}^{0} \in \mathbb{B}(\theta^{*}, \rho)$ for some radius $\rho > 0$, by choosing the scaling parameter $\beta$ such that 
\begin{align}
\beta^2 = 1 - \frac{(1 - \eta c_{1})^2}{2 \log (n/d)} \label{eq:varied_beta_value}
\end{align}
where $c_{1}$ is the universal constant in the homogeneous assumption~\ref{assump:homogeneous}, there exist universal constants $\{C_{i}\}_{i = 1}^{2}$ such that as long as $\theta^{*} = 0$, the sample size $n \geq C_{1}(d \log(d/\delta))^{2p}$, $t \geq C_{2} \log^2(n/(d + \log(1/\delta)))$, and with the step size $\eta$ such that $\frac{\eta c_{1} \|\theta_{n}^{0} - \theta^{*}\|^{2p - 2} \beta}{(2p - 1) 2p} + \beta^{1/(p - 1)} \geq 1$ and $2 p \cdot 3^{2p - 2} \cdot \eta c_{1}  \|\theta_{n}^{0} - \theta^{*}\|^{2p - 2} \leq 2p - 1$, with probability $1 - \delta$ the following holds:
$$\min_{1 \leq k \leq t} \|\theta_{n}^{t} - \theta^{*}\| \leq C_{6} \log \parenth{\frac{n}{d + \log(1/\delta)}} \parenth{\frac{d + \log(1/\delta)}{n}}^{1/(2p)}.$$
\end{corollary}

\vspace{0.5 em}
\noindent
\textbf{Experimental results for the low SNR regime:} To illustrate the statistical behaviors of the EGD iterates, we consider the parameter estimation of these iterates in the phase retrieval setting, which corresponds to the case where the power of the link function is chosen to be $p = 2$. We run both the EGD and GD algorithm for solving the sample least-square loss function $\overline{\mathcal{L}}_{n}$ in equation~\eqref{eq:sample_likelihood_generalized_linear}. For the low SNR regimes, we use $d = 4$, $\beta = 0.9$ and $\eta = 0.001$. The experiment results are shown in Figure~\ref{fig:GLM_Low_SNR}. To illustrate that we can obtain the early stopping behavior of the EGD iterates via cross-validation, for all of the experiments we use 90\% of the data to compute the gradient for the GD and EGD iterates, and use the remaining 10\% of the data to perform cross validation. The red diamond in Figure~\ref{fig:GLM_Low_SNR} shows the EGD iterate with the minimum validation loss. As we observe from these plots, the early stopping at the red diamond point aligns quite well with the final statistical radius under the low SNR regime of the generalized linear models.


\vspace{0.5 em}
\noindent\textbf{Practical implications:} In practice, we can run the GD and EGD algorithms simultaneously to estimate $\theta^{*}$ in the generalized linear models. If we observe that the sample GD iterates converge geometrically fast, this indicates that we are in the high SNR regime of the generalized linear model. Therefore, we can just use the GD algorithm for approximating the optimal parameter under that setting. However, if we observe that the sample GD iterates converge slowly while the EGD updates converge geometrically fast to the optimal solution, then it implies that the model is under the low SNR regime and we can just use the EGD algorithm for parameter estimation under that setting. Since the EGD and GD algorithms have similar per iteration cost, the simultaneous running of both the EGD and GD algorithm only slightly increases the per iteration cost.

\section{Additional Examples: Gaussian Mixture Models}
\label{sec:example_Gaussian_mixture}
In this appendix, we discuss an application of the general theory developed in Section~\ref{sec:statistics_rate_EGD} to establish statistical rates of EGD algorithm for solving parameter estimation in Gaussian mixture models. 
Gaussian mixture models have been a popular class of statistical models to study heterogeneity of data~\cite{Lindsay-1995, Mclachlan-1988}. In these models, parameter estimation plays an important role in capturing the behaviors of different sub-populations in the data. A popular approach to obtain parameter estimation in Gaussian mixture models is by using maximum likelihood estimation (MLE). However, since the log-likelihood function of these models is non-concave and has complicated landscape, the MLE does not have a closed-form expression and optimization algorithms are employed to approximate the MLE. The Expectation-Maximization (EM) algorithm is the most popular optimization algorithm for approximating the MLE. The statistical guarantee of the EM algorithm under Gaussian mixture models has been studied extensively~\cite{Siva_2017}. When the number of components is known and the separation of location parameters is large enough, via appropriate initialization in the neighborhood of the true parameter, the EM updates are shown to reach a radius of convergence $\mathcal{O}((d/n)^{1/2})$ around the true location parameters after $\mathcal{O}(\log(n/d))$  iterations~\cite{Siva_2017, Caramanis-nips2015, Hsu-nips2016, Daskalakis_colt2017, Sarkar_nips2017}. However, the number of components in Gaussian mixture models is rarely known in practice. A common approach to deal with that situation is to over-specify the number of components, namely, we choose a large number of components based on our domain knowledge of the data and obtain parameter estimation within that large model~\cite{Chen1992, Rousseau-2011}. This approach is widely regarded as an over-specified Gaussian mixture model. Nevertheless, the EM algorithm was shown to have sub-linear computational complexity to reach the final radius around the true parameter under the over-specified Gaussian mixtures~\cite{Raaz_Ho_Koulik_2020, Raaz_Ho_Koulik_2018_second}.  

To shed light on the behavior of the EGD algorithm under Gaussian mixture models, we specifically consider symmetric two-component location Gaussian mixtures. While these settings may sound simplistic, they have been used extensively in previous works to understand the non-asymptotic behavior of the EM algorithm~\cite{Siva_2017, Caramanis-nips2015, Raaz_Ho_Koulik_2020, Raaz_Ho_Koulik_2018_second}. In particular, we assume that $X_{1}, X_{2}, \ldots, X_{n}$ are i.i.d. data from the true model 
\begin{align}
    \frac{1}{2} \mathcal{N}(-\theta^{*}, \sigma^2 I_{d}) + \frac{1}{2} \mathcal{N}(\theta^{*}, \sigma^2 I_{d}), \label{eq:Gaussian_mixture}
\end{align} 
where $\theta^{*}$ is unknown true parameter and $\sigma$ is a given scale parameter. To estimate $\theta^{*}$, we utilize the symmetric two-component location Gaussian mixtures $\frac{1}{2} \mathcal{N}(-\theta, \sigma^2 I_{d}) + \frac{1}{2} \mathcal{N}(\theta, \sigma^2 I_{d})$.  Then, we have the corresponding negative sample log-likelihood function of the fitted model as follows:
\begin{align}
    \mathcal{L}_{n}(\theta) : = - \frac{1}{n} \sum_{i = 1}^{n} \log \parenth{\frac{1}{2} \phi(X_{i}|-\theta, \sigma^2 I_{d}) + \frac{1}{2} \phi(X_{i}| \theta, \sigma^2 I_{d})}, \label{eq:sample_likelihood_Gaussian}
\end{align}
where $\phi(x|\theta, \sigma^2 I_{d})$ is the density function of the multivariate Gaussian distribution with mean $\theta$ and covariance matrix $\sigma^2 I_{d}$.

\vspace{0.5 em}
\noindent\textbf{Statistical and computational complexities of the EM algorithm:} As the optimal solution of the negative log-likelihood function $\mathcal{L}_{n}$ does not have closed-form expression, the EM algorithm has been widely used to approximate it. Under the strong separation settings, namely, $\|\theta^{*}\|/\sigma \geq C$ for some sufficiently large $C$, the EM updates were shown to reach the optimal statistical radius $\mathcal{O}((d/n)^{1/2})$ around the true location $\theta^{*}$ after $\mathcal{O}(\log(n/d))$ iterations~\cite{Siva_2017}. However, under the over-specified setting, namely, when $\theta^{*} = 0$, the EM iterates converge to the final statistical radius of convergence $\mathcal{O}((d/n)^{1/4})$ around the true parameter $\theta^{*}$ after $\mathcal{O}(\sqrt{n/d})$  iterations~\cite{Raaz_Ho_Koulik_2020}. This indicates that the total computational complexity of the EM algorithm for reaching the final statistical radius when $\theta^{*} = 0$ is $\mathcal{O}(n^{3/2} d^{1/2})$, which is sub-optimal in $n$.

\subsection{The EGD algorithm under the low SNR (over-specified) regime} We now analyze the behavior of the EGD algorithm for solving parameter estimation under the over-specified (or equivalently low SNR) regime of symmetric two-component location Gaussian mixture models~\eqref{eq:Gaussian_mixture}, namely, $\theta^{*} = 0$. It is sufficient to verify Assumptions~\ref{assump:homogeneous} and~\ref{assump:stab}. We first need to define the corresponding negative population log-likelihood function of the symmetric two-component Gaussian mixtures as follows:
\begin{align}
    \mathcal{L}(\theta) : = \Exs_{X} \brackets{\log \parenth{\frac{1}{2} \phi(X|-\theta, \sigma^2 I_{d}) + \frac{1}{2} \phi(X|~\theta, \sigma^2 I_{d})}}, \label{eq:population_likelihood_Gaussian}
\end{align}
where $X \sim \frac{1}{2} \mathcal{N}(X|-\theta^{*}, \sigma^2 I_{d}) + \frac{1}{2} \mathcal{N}(X|~\theta^{*}, \sigma^2 I_{d})$ and $\theta^{*} = 0$. Indeed, from Appendix A.1 in~\cite{ren2021towards}, the homogeneous Assumption~\ref{assump:homogeneous} is satisfied when $\alpha = 2$ and $\rho = \sigma/2$, namely, we have
\begin{align}
    \lambda_{\max}(\nabla^2 \mathcal{L}(\theta)) \leq c_1 \|\theta - \theta^*\|^{2}, \quad \quad \|\nabla \mathcal{L}(\theta)\| \geq c_2(\mathcal{L}(\theta) - \mathcal{L}(\theta^*))^{3/4}, \label{eq:homogeneous_Gaussian_mixtures}
\end{align}
where $c_{1}$ and $c_{2}$ are some universal constants. For Assumption~\ref{assump:stab}, the result of Lemma 1 in~\cite{Raaz_Ho_Koulik_2020} indicates that this assumption is satisfied with $\gamma = 1$ and the noise function $\varepsilon(n, \delta) = \sqrt{\frac{d + \log(1/\delta)}{n}}$, namely, the following uniform concentration bound holds:
\begin{align}
    \sup_{\theta\in \mathbb{B}(\theta^*, r)} \|\nabla \mathcal{L}_n(\theta) - \nabla \mathcal{L}(\theta)\|\leq c_3 r \sqrt{\frac{d + \log(1/\delta)}{n}} \label{eq:concentration_Gaussian_mixtures}
\end{align}
for any $r > 0$ as long as the sample size $n \geq c_{4} d \log(1/ \delta)$.

Given the results in equations~\eqref{eq:homogeneous_Gaussian_mixtures} and~\eqref{eq:concentration_Gaussian_mixtures}, an application of Theorem~\ref{theorem:statistical_rate_EGD} results in the following statistical and computational complexities of EGD updates when $\theta^{*} = 0$.
\begin{corollary}
\label{corollary_Gaussian_mixture}
Given the over-specified symmetric two-component location Gaussian mixtures~\eqref{eq:Gaussian_mixture}, namely, $\theta^{*} = 0$, and the initialization $\theta_{n}^{0} \in \mathbb{B}(\theta^{*}, \frac{\sigma}{2})$, there exist universal constants $C_{1}, C_{2}, C_{3}$ such that as long as $n \geq C_{1} d \log(1/ \delta)$, and $t \geq C_{2} \log(n/ (d + \log(1/\delta))$, with probability $1 - \delta$ we find that
\begin{align*}
    \min_{1 \leq k \leq t} \|\theta_{n}^{k} - \theta^{*}\| \leq C_{3} \parenth{\frac{d}{n}}^{1/4}.
\end{align*}
\end{corollary}
The result of Corollary~\ref{corollary_Gaussian_mixture} indicates that the EGD updates reach the final statistical radius $\mathcal{O}((d/n)^{1/4})$ after $\mathcal{O}(\log(n/d))$ iterations. Since each iteration of the EGD algorithm has a complexity of $\mathcal{O}(n d)$, this implies that the total computational complexity of the EGD algorithm for reaching the final statistical radius of estimating the location parameter in over-specified two-component location Gaussian mixtures is $\mathcal{O}(n d \log(n/d))$. It is much better than that of the EM algorithm, which is of the order $\mathcal{O}(n^{3/2} d^{1/2})$ in terms of the sample size $n$. 

\begin{figure}[!t]
    \centering
    \includegraphics[width=0.48\textwidth]{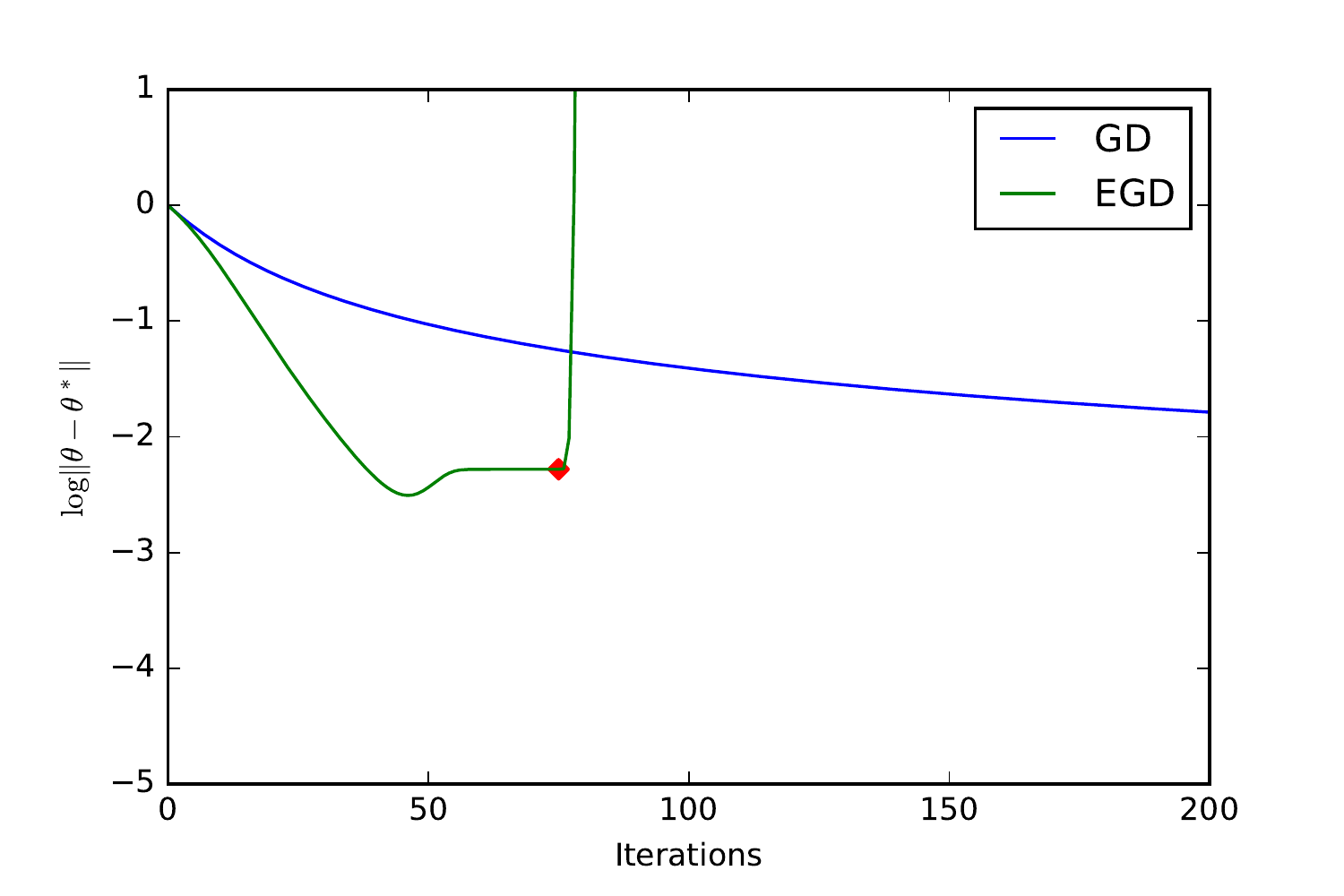}
    \includegraphics[width=0.48\textwidth]{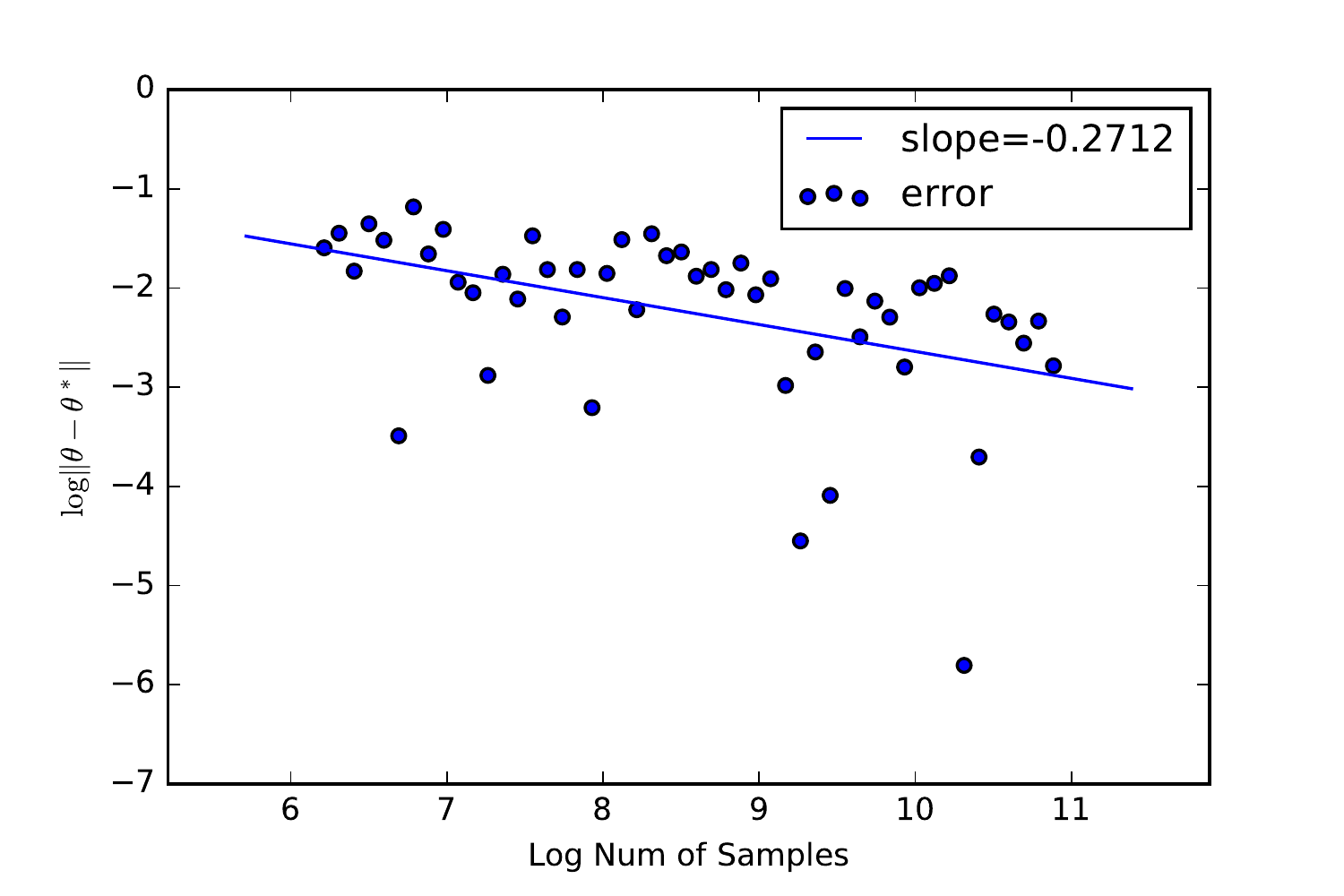}
    \caption{\textit{Over-specified (Low SNR) GMM: Computational complexity and statistical rates of the EM and EGD iterates for solving the parameter estimation under the over-specified (low SNR) regime of the symmetric two-component Gaussian mixture models~\eqref{eq:Gaussian_mixture} when $d = 4$}. \textbf{Left:} The EGD iterates first converge to the statistical radius linearly before diverging, while the EM iterates converge to the statistical radius sub-linearly. The red diamond shows the EGD iterate with the minimum validation loss when we use 90\% of the data to compute the gradient for the GD and EGD iterates, and use the remaining 10\% of the data to perform cross validation. \textbf{Right:} Both the EM and EGD can find a solution within the statistical radius $\mathcal{O}(n^{-1/4})$.}
    \label{fig:GMM_Low_SNR}
\end{figure}

\vspace{0.5 em}
\noindent
\textbf{Experimental results for the over-specified regime:} We now provide experimental results for the statistical and computational complexities of the EM and EGD algorithms under the over-specified settings of the Gaussian mixture model~\eqref{eq:Gaussian_mixture}. We use $d = 4$, $\beta = 0.9$, and $\eta = 0.001$. The experimental results are shown in Figure~\ref{fig:GMM_Low_SNR}. Similar to the experiments in Figure~\ref{fig:GLM_Low_SNR}, we also perform cross-validation for the EGD iterates in Figure~\ref{fig:GMM_Low_SNR}
and the early stopping is represented by the diamond point in that figure. As we observe from Figure~\ref{fig:GMM_Low_SNR}, the early stopping at the red diamond point aligns quite well with the final statistical radius under the over-specified regime of the symmetric two-component Gaussian mixtures.

\subsection{The EGD algorithm in the high SNR regime} 
We now move to the high SNR regime of the symmetric two-component Gaussian mixture model~\eqref{eq:Gaussian_mixture}, namely, $\|\theta^{*}\|/\sigma \geq C$ where $C$ is some universal constant. 

\vspace{0.5 em}
\noindent
\textbf{The EGD algorithm with fixed $\beta$ in the high SNR regime:}
From Corollary 1 of~\cite{Siva_2017}, the population loss function $\mathcal{L}$ is locally strongly convex and smooth in $\mathbb{B} \parenth{\theta^{*}, \frac{\|\theta^{*}\|}{4}}$, namely, the homogeneous assumption~\ref{assump:homogeneous} is satisfied with the constant $\alpha = 0$. It means that as long as $\theta \in \mathbb{B} \parenth{\theta^{*}, \frac{\|\theta^{*}\|}{4}}$ we have
\begin{align}
    \lambda_{\max}(\nabla^2 \mathcal{L}(\theta)) & \leq c_1, \nonumber \\
    \|\nabla \mathcal{L}(\theta)\| & \geq c_2(\mathcal{L}(\theta) - \mathcal{L}(\theta^*))^{1/2}. \label{eq:homogeneous_Gaussian_mixtures_strong_signal}
\end{align}
Furthermore, the stability assumption~\ref{assump:stab} is satisfied with $\gamma = 0$, namely, the following uniform concentration bound holds:
\begin{align}
    \sup_{\theta\in \mathbb{B}(\theta^*, r)} \|\nabla \mathcal{L}_n(\theta) - \nabla \mathcal{L}(\theta)\|\leq c_3 \sqrt{\frac{d + \log(1/\delta)}{n}} \label{eq:concentration_Gaussian_mixtures_strong_signal}
\end{align}
for any $r > 0$ as long as the sample size $n \geq c_{4} d \log(1/ \delta)$.

Given the results in equations~\eqref{eq:homogeneous_Gaussian_mixtures_strong_signal} and~\eqref{eq:concentration_Gaussian_mixtures_strong_signal}, a direct application of Theorem~\ref{theorem:statistics_EGD_strongly_convex} leads to the following result for the statistical guarantee of the sample EGD iterates.

\begin{figure}[!t]
    \centering
    \includegraphics[width=0.48\textwidth]{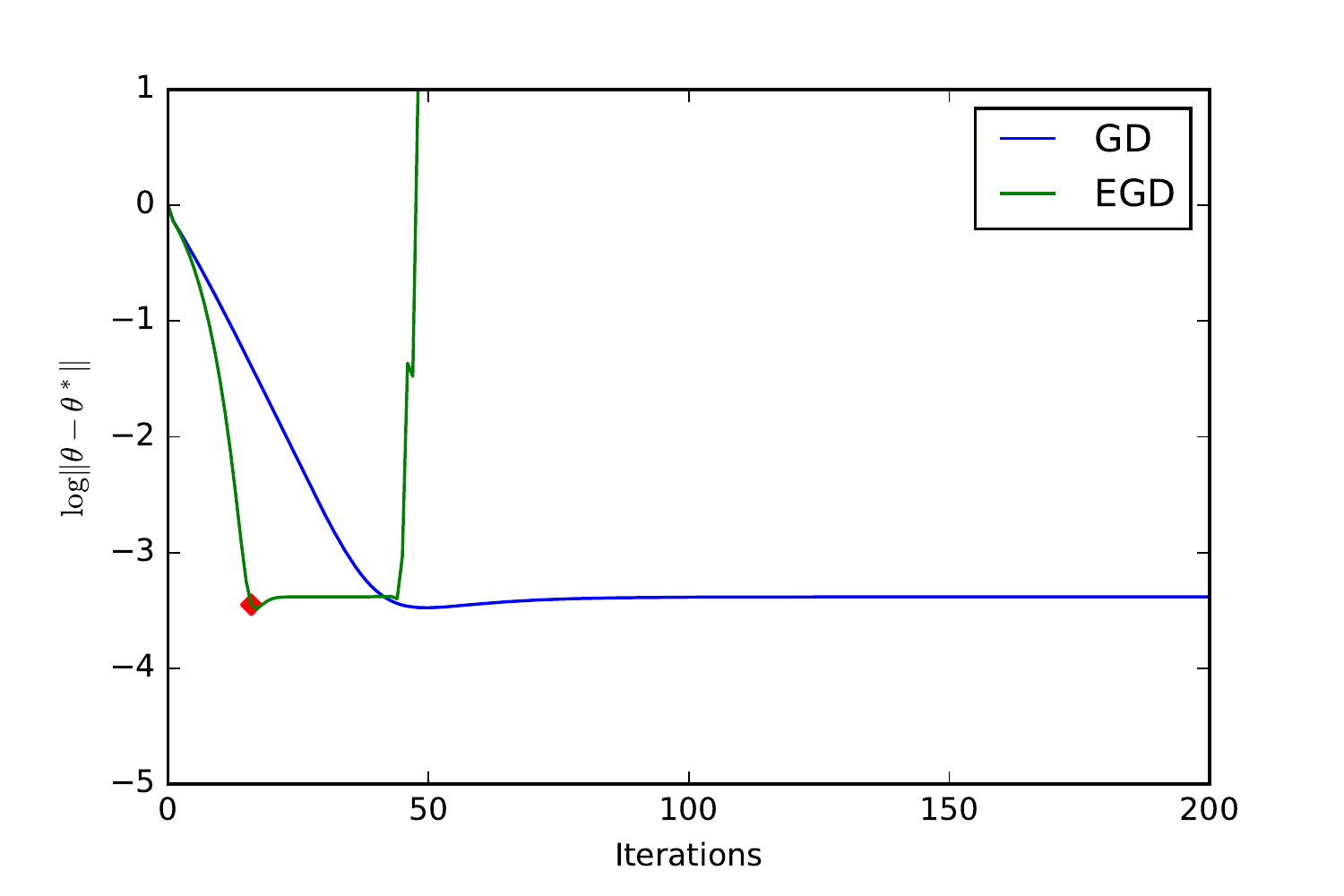}
    \includegraphics[width=0.48\textwidth]{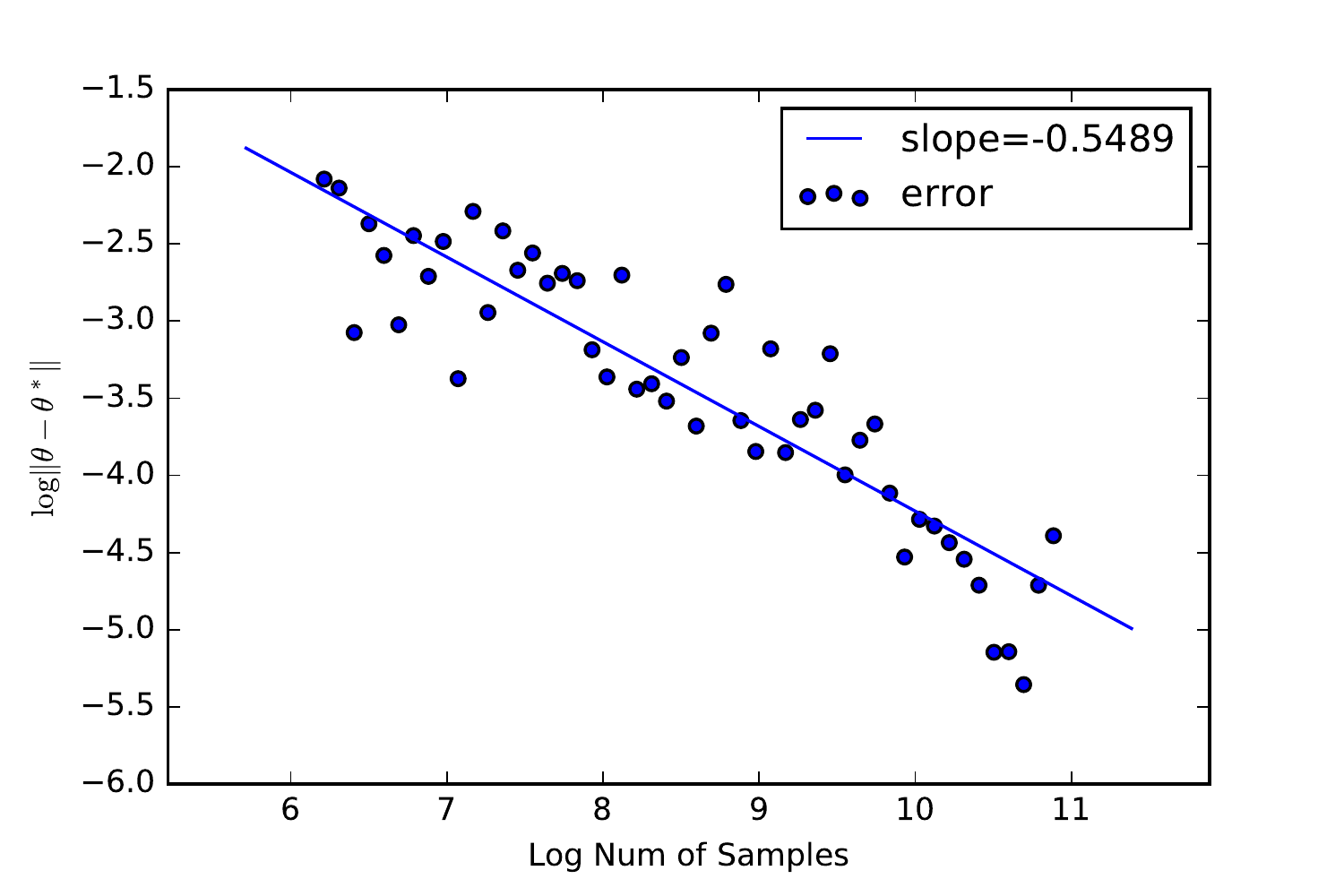}
    \caption{\textit{High SNR regime for GMM: Computational complexity and statistical rates of the EM and EGD iterates for solving the parameter estimation under the high SNR regime of the symmetric two-component Gaussian mixture model~\eqref{eq:Gaussian_mixture} when $d = 4$}. \textbf{Left:} The EM iterates converge to the statistical radius linearly, while the EGD iterates first converge to the statistical radius with a linear rate faster than EM iterates, then diverge. The red diamond shows the EGD iterate with the minimum validation loss when we use 90\% of the data to compute the gradient for the GD and EGD iterates, and use the remaining 10\% of the data to perform cross validation. \textbf{Right:} Both the EM and EGD can find a solution within the statistical radius $\mathcal{O}(n^{-1/2})$.}
    \label{fig:GMM_Strong_SNR}
\end{figure}

\begin{corollary}
\label{corollary_mixture_model_strong_signal}
Given the strong signal-to-noise regime of the symmetric two-component Gaussian mixture model~\eqref{eq:Gaussian_mixture}, i.e., $\|\theta^{*}\|/\sigma \geq C$ for some universal constant $C$, and the initialization $\theta_{n}^{0} \in \mathbb{B} \parenth{\theta^{*}, \frac{\|\theta^{*}\|}{4}}$, there exist universal constants $C_{1}$ and $C_{2}$ such that as long as $n \geq C_{1} d \log(1/ \delta)$, the step size $\eta < 1/ c_{1}$ and $\bar{T} = \log(1/(\eta c_{1}))/ \log(1/\beta)$ where $c_{1}$ is the universal constant in equation~\eqref{eq:homogeneous_Gaussian_mixtures_strong_signal}, with probability $1 - \delta$ we find that
\begin{align*}
    \min_{1 \leq t \leq \bar{T}} \|\theta_{n}^{t} - \theta^{*}\| \leq C_{2} \parenth{\frac{d}{n}}^{1/2} + \exp \parenth{- \frac{(1 - \eta c_{1})(\beta^{-1} - c_{1} \eta)}{2(\beta^{-2} - 1)}} \|\theta_{n}^{0} - \theta^{*}\|.
\end{align*}
\end{corollary}
For the optimization error $\exp \parenth{- \frac{(1 - \eta c_{1})(\beta^{-1} - c_{1} \eta)}{2(\beta^{-2} - 1)}} \|\theta_{n}^{0} - \theta^{*}\|$, this does not vanish to 0. Therefore, within the first $\bar{T}$ iterations the sample EGD iterates only converge to a neighborhood of $\theta^{*}$ that has radius $C_{2} \parenth{\frac{d}{n}}^{1/2} + \exp \parenth{- \frac{(1 - \eta c_{1})(\beta^{-1} - c_{1} \eta)}{2(\beta^{-2} - 1)}} \|\theta_{n}^{0} - \theta^{*}\|$. After $\bar{T}$ iterations, the sample EGD iterates diverge (Please refer to Figure~\ref{fig:GMM_Strong_SNR} for these behavior of the EGD algorithm in the high SNR regime of the symmetric two-component Gaussian mixture~\eqref{eq:Gaussian_mixture}). On the other hand, under this regime the EM iterates converge to the final statistical radius $\mathcal{O}(\sqrt{d/n})$ within the true parameter $\theta^{*}$ after $\mathcal{O}(\log(n/d))$ iterations. 

\vspace{0.5 em}
\noindent
\textbf{The EGD algorithm with varied $\beta$:} Similar to the generalized linear model, the appearance of the non-vanishing optimization error $\exp \parenth{- \frac{(1 - \eta c_{1})(\beta^{-1} - c_{1} \eta)}{2(\beta^{-2} - 1)}} \|\theta_{n}^{0} - \theta^{*}\|$ can be undesirable. When we choose $\beta^2 = 1 - \frac{C}{\log (n/d)}$ for some constant $C$, under the high SNR regime of the symmetric two-component Gaussian mixture model~\eqref{eq:Gaussian_mixture}, the EGD iterates are able to reach to the final statistical radius $\mathcal{O}((d/n)^{1/2})$ (up to some logarithmic factor) after $\mathcal{O}(\log(n/d))$ logarithmic number of iterations, which is comparable to the GD iterates. Under the low SNR regime of these models, Proposition~\ref{proposition:rate_EGD_varied_beta} in Appendix~\ref{sec:varied_beta} demonstrates that the EGD iterates still reach the final statistical radius after $\mathcal{O}(\log^2(n/d))$ number of iterations, which is slightly more expensive than the number of iterations $\mathcal{O}(\log(n/d))$ of the EGD iterates when we fix $\beta$ in Corollary~\ref{corollary_generalized_linear} and is still much cheaper than the number of iterations $\mathcal{O}((n/d)^{(p - 1)/p})$ of the GD iterates. We summarize these results in the following corollary.
\begin{corollary}
\label{corollary_gaussian_mixture_varied_beta}
Given the symmetric two-component Gaussian mixtures~\eqref{eq:Gaussian_mixture} and the initialization $\theta_{n}^{0} \in \mathbb{B}(\theta^{*}, \rho)$ for some radius $\rho > 0$, by choosing the scaling parameter $\beta$ such that 
\begin{align}
\beta^2 = 1 - \frac{(1 - \eta c_{1})^2}{2 \log (n/d)} \label{eq:varied_beta_value}
\end{align}
where $c_{1}$ is the universal constant in the homogeneous assumption~\ref{assump:homogeneous}, there exist universal constants $\{C_{i}\}_{i = 1}^{6}$ such that as long as the sample size $n \geq C_{1} d \log(d/\delta)$, with probability $1 - \delta$ the following holds:

(a) High SNR regime: When $\|\theta^{*}\|/ \sigma \geq C_{2}$, as long as $t \geq C_{3} \log(n/(d + \log(1/\delta))$ we find that $$\min_{1 \leq k \leq t} \|\theta_{n}^{t} - \theta^{*}\| \leq C_{4} \log \parenth{\frac{n}{d + \log(1/\delta)}}\parenth{\frac{d + \log(1/\delta)}{n}}^{1/2}.$$

(b) Low SNR regime: When $\theta^{*} = 0$ and $t \geq C_{5} \log^2(n/(d + \log(1/\delta)))$, with the step size $\eta$ such that $\frac{\eta c_{1} \|\theta_{n}^{0} - \theta^{*}\|^{2} \beta}{12} + \beta \geq 1$ and $\eta c_{1}  \|\theta_{n}^{0} - \theta^{*}\|^{2} \leq 1/12$ we obtain that $$\min_{1 \leq k \leq t} \|\theta_{n}^{t} - \theta^{*}\| \leq C_{6} \log \parenth{\frac{n}{d + \log(1/\delta)}} \parenth{\frac{d + \log(1/\delta)}{n}}^{1/4}.$$
\end{corollary}

\vspace{0.5 em}
\noindent
\textbf{Experiments:} We now provide experimental results for the statistical and computational complexities of the EM and EGD algorithms under the high SNR settings of the Gaussian mixture model~\eqref{eq:Gaussian_mixture}. For both the low and high SNR regimes, we use $d = 4$, $\beta = 0.9$, and $\eta = 0.001$. The experimental results are shown in Figure~\ref{fig:GMM_Strong_SNR}.

\section{Discussion: Inhomogeneous Settings}
\label{sec:inhomogeneous_settings}
In this section, we briefly discuss the behaviors of EGD algorithm for solving the population loss function $f$ that goes beyond the homogeneous assumption~\ref{assump:homogeneous}. 
\subsection{Diagonal Settings}
\label{sec:diagonal_setting}
We first consider the diagonal settings of the population loss function $f$, namely, $f(\theta) = \sum_{i = 1}^{d} \theta_{i}^{2\alpha_{i}}$ where $\alpha_{1}, \ldots, \alpha_{d} > 1$. When $\alpha_{1} = \alpha_{2} = \ldots = \alpha_{d}$, we go back to the homogeneous setting in Section~\ref{sec:optimization_rate_EGD}. When not all the powers $\alpha_{1}, \ldots, \alpha_{d}$ are the same, the homogeneous Assumption~\ref{assump:homogeneous} no longer holds. The following result demonstrates the linear convergence rate of both the EGD iterates and their objective values to the optimal parameter $\theta^{*}$ and optimal objective value $f(\theta^{*})$.
\begin{corollary}
\label{cor:diagonal_setting}
Assume that $f(\theta) = \sum_{i = 1}^{d} \theta_{i}^{2\alpha_{i}}$ where $\alpha_{1}, \ldots, \alpha_{d} > 1$. Then
\begin{align*}
    |\theta_i^t| \leq \bar{C}_i \beta^{\frac{t}{2\alpha_i - 2}},
\end{align*}
where $\bar{C}_i$ is some universal constant that $\bar{C}_i \leq \left(\theta_i^0\right)^{2\alpha_i}$.
\end{corollary}
The key observation is that, in the diagonal setting, each dimension is optimized individually. As a result, we can apply our analysis to every individual dimension, which leads to the Corollary~\ref{cor:diagonal_setting}.
\subsection{Beyond Diagonal Settings}
\label{sec:beyond_diagonal_setting}
One interesting observation from the diagonal setting is that, on the whole optimization trajectory, all of eigenvalues of the Hessian matrix scale as $\Theta(\beta^t)$. Hence, the EGD algorithm can be intuitively understood as a variant of the Newton method on the objective functions whose Hessian matrix has this desired property. However, for the scenarios beyond diagonal settings, such property is not necessarily held and also hard to check. To illustrate this, we consider the following simple example $f(\theta) = (\theta_1^2 + \theta_2^4)^2$, whose Hessian matrix can be written as
\begin{align*}
    \nabla_{\theta}^2 f(\theta) = \begin{bmatrix}
    12\theta_1^2 + 4\theta_2^4 & 16\theta_1 \theta_2^3\\
    16\theta_1 \theta_2^3 & 24\theta_1^2\theta_2^2 + 56 \theta_2^6
    \end{bmatrix}.
\end{align*}
It is straightforward to see that there does not exist a regime of $\theta_1$ and $\theta_2$, such that eigenvalues of $\nabla_{\theta}^2 f(\theta)$ are always at the same order. Hence, understanding the behavior of the EGD algorithm on this kinds objective can be non-trivial, and we leave it as the future work.
\section{Discussion: The EGD Algorithm under the Middle SNR Regime}
\label{sec:discussion_middle_SNR}
\begin{figure}[!t]
    \centering
    \includegraphics[width=0.48\textwidth]{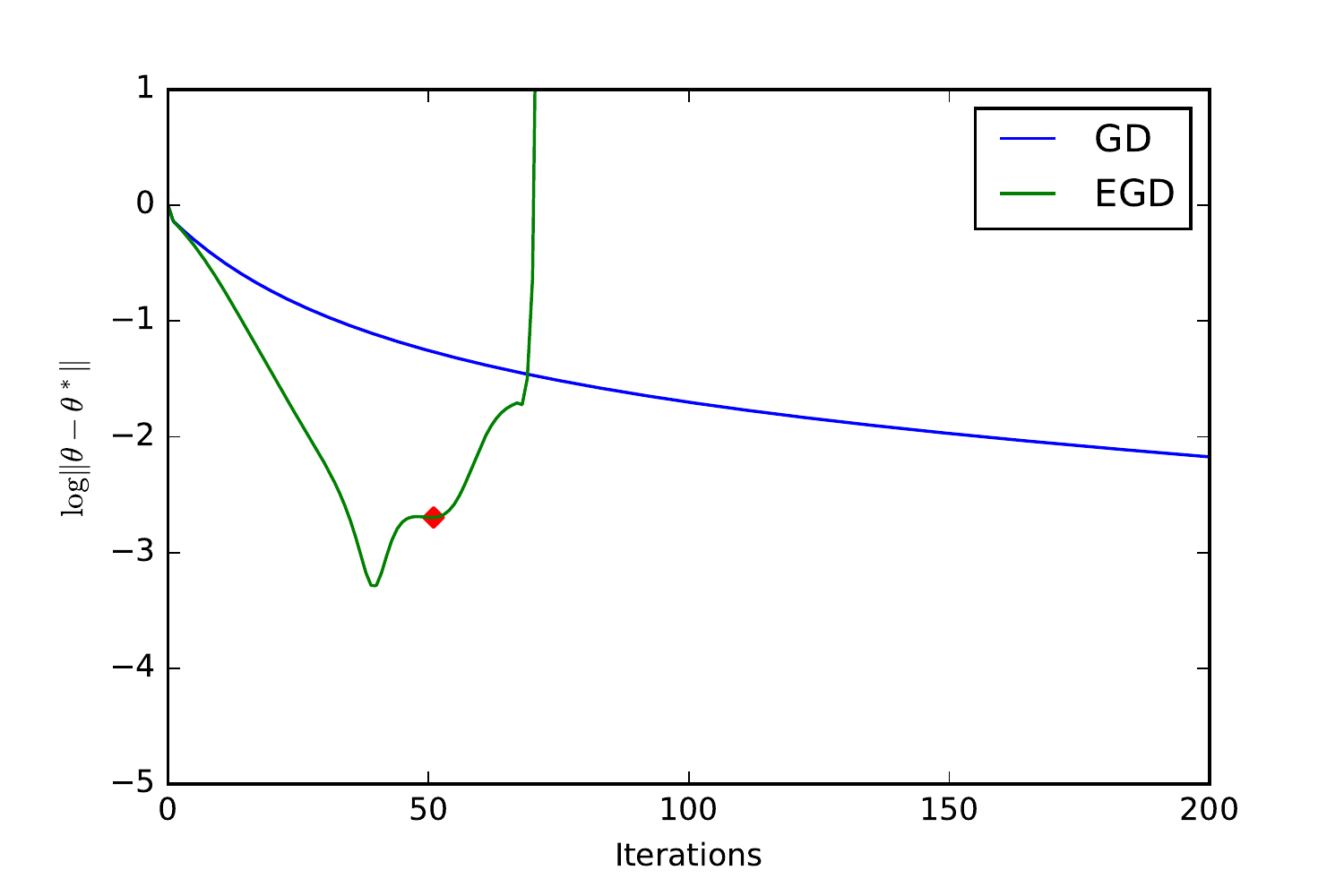}
    \includegraphics[width=0.48\textwidth]{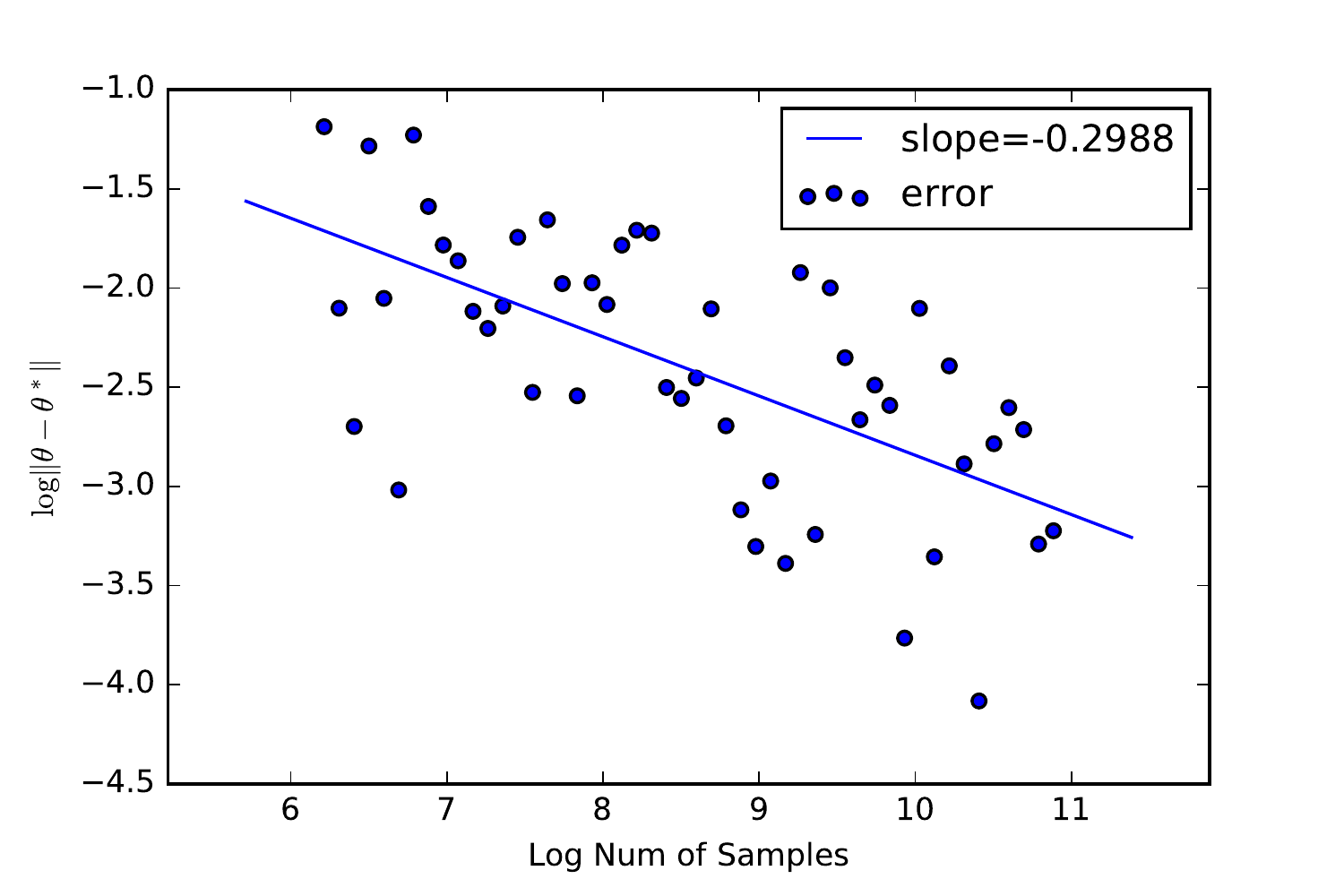}
    \caption{\textit{Middle SNR regime for GLM: Computational complexity and statistical rates for the GD and EGD iterates for solving the parameter estimation under the middle SNR regime of the generalized linear model~\eqref{eq:Gaussian_mixture} when $p = 2$, $d = 4$, and $\|\theta^{*}\| = (d/n)^{1/6}$}. \textbf{Left:} The EM iterates converge to the statistical radius sub-linearly, while the EGD iterates first converge to the statistical radius with a linear rate, then diverge. The red diamond shows the EGD iterate with the minimum validation loss when we use 90\% of the data to compute the gradient for the GD and EGD iterates, and use the remaining 10\% of the data to perform cross validation. \textbf{Right:} Both the EM and EGD can find a solution within the statistical radius $\mathcal{O}(\|\theta^{*}\|^{-1} (d/n)^{1/2})$.}
    \label{fig:GMM_middle_SNR}
\end{figure}
Thus far in the paper, we have not established a statistical guarantee for the EGD algorithm under the middle SNR regime of the generalized linear models~\eqref{eq:generalized_linear_model}. This is largely due to the challenge of establishing the homogeneous assumption~\ref{assump:homogeneous} for the population loss functions of these models in the middle SNR regime. To simplify the discussion, we specifically focus on the phase retrieval settings. Under the middle SNR regime, we have $C_{1} (d/n)^{1/4} \leq \|\theta^{*}\| \leq C_{2}$ where $C_{1}$ and $C_{2}$ are some universal constants. By adapting the techniques from the current work of Kwon et al.~\cite{Kwon_minimax}, we can verify that the GD iterates reach the final statistical radius $\mathcal{O} (\|\theta^{*}\|^{-1} \sqrt{d/n})$ after $\mathcal{O} \parenth{\|\theta^{*}\|^{-2} \log(n/d)}$ iterations. When $\|\theta^{*}\| = \mathcal{O}((d/n)^{-\tau})$ where $\tau < 1/4$, the total computational complexity of the GD algorithm for reaching the final statistical radius is $\mathcal{O}(n^{2\tau})$, which is polynomial in $n$ and is computationally expensive when the sample size $n$ is large. For the statistical behaviors of the EGD algorithm in the middle SNR regime, we conjecture that the EGD iterates still reach the final statistical radius after a logarithmic number of iterations.
\begin{conjecture}
\label{conjecture:middle_SNR_Gaussian}
In the middle SNR regime of generalized linear model~\eqref{eq:generalized_linear_model} when the power of the link function $p = 2$, namely, $C_{1} (d/n)^{1/4} \leq \|\theta^{*}\| \leq C_{2}$ where $C_{1}$ and $C_{2}$ are some universal constants, and the initialization $\theta_{n}^{0} \in \mathbb{B}(\theta^{*}, \rho)$ for some constant $\rho$, there exist universal constants $C_{3}, C_{4}, C_{5}$ such that as long as $n \geq C_{3} (d \log(d/ \delta))^4$, and $t \geq C_{4} \log(n/ (d + \log(1/\delta))$, with probability $1 - \delta$ we find that
\begin{align*}
    \min_{1 \leq k \leq t} \|\theta_{n}^{k} - \theta^{*}\| \leq C_{5} \frac{1}{\|\theta^{*}\|} \sqrt{\frac{d}{n}}.
\end{align*}
\end{conjecture}
We provide the experimental results in Figure~\ref{fig:GMM_middle_SNR} to support this conjecture. For the experimental setting, we choose $d = 4$, $\|\theta^{*}\| = (d/n)^{1/6}$, $\beta = 0.9$, and $\eta = 0.001$. As indicated in Figure~\ref{fig:GMM_middle_SNR}, the EGD iterates only need a  logarithmic number of iterations to reach the statistical radius $\mathcal{O} (\|\theta^{*}\|^{-1} \sqrt{d/n})$, which is in stark contrast to the GD iterates that need a polynomial number of iterations. We leave the theoretical verification of the above conjecture for the future work.
\end{document}

%% file: final_macros.tex
\newcommand{\brackets}[1]{\left[ #1 \right]}
\newcommand{\parenth}[1]{\left( #1 \right)}

\newcommand{\Exs}{\ensuremath{{\mathbb{E}}}}

\newtheoremstyle{named}{}{}{\itshape}{}{\bfseries}{.}{.5em}{\thmnote{#3's }#1}
\theoremstyle{named}

\theoremstyle{plain}

\newtheorem{theorem}{Theorem}
\newtheorem{proposition}{Proposition}

\newtheorem{corollary}{Corollary}

\newtheorem{conjecture}{Conjecture}

\newlength{\widebarargwidth}
\newlength{\widebarargheight}
\newlength{\widebarargdepth}

\makeatletter
\long\def\@makecaption#1#2{
        \vskip 0.8ex
        \setbox\@tempboxa\hbox{\small {\bf #1:} #2}
        \parindent 1.5em  
        \dimen0=\hsize
        \advance\dimen0 by -3em
        \ifdim \wd\@tempboxa >\dimen0
                \hbox to \hsize{
                        \parindent 0em
                        \hfil
                        \parbox{\dimen0}{\def\baselinestretch{0.96}\small
                                {\bf #1.} #2
                                }
                        \hfil}
        \else \hbox to \hsize{\hfil \box\@tempboxa \hfil}
        \fi
        }
\makeatother

\long\def\comment#1{}
\definecolor{battleshipgrey}{rgb}{0.52, 0.52, 0.51}
\definecolor{darkgray}{rgb}{0.66, 0.66, 0.66}
\definecolor{darkgreen}{rgb}{0.0, 0.2, 0.13}
\definecolor{darkspringgreen}{rgb}{0.09, 0.45, 0.27}
\definecolor{dukeblue}{rgb}{0.0, 0.0, 0.61}
\definecolor{olivedrab7}{rgb}{0.24, 0.2, 0.12}
\definecolor{darkblue}{rgb}{0.0, 0.0, 0.55}
\definecolor{darkscarlet}{rgb}{0.34, 0.01, 0.1}
\definecolor{candyapplered}{rgb}{1.0, 0.03, 0.0}
\definecolor{ao(english)}{rgb}{0.0, 0.5, 0.0}
\definecolor{applegreen}{rgb}{0.55, 0.71, 0.0}
